\documentclass[11pt]{article}

\usepackage[final]{acl}

\usepackage{times}
\usepackage{latexsym}
\usepackage{threeparttable}
\usepackage{booktabs}
\usepackage{tabularx}
\usepackage[most]{tcolorbox}
\usepackage{listings}
\usepackage{amssymb}
\usepackage{pifont}
\usepackage{enumitem}
\usepackage{multirow}
\usepackage{subcaption}
\usepackage{graphicx}
\usepackage{appendix}
\usepackage[table]{xcolor}

\usepackage[most]{tcolorbox}
\usepackage{listings}
\usepackage{xcolor}
\usepackage{makecell}

\definecolor{jsonkey}{rgb}{0,0,1}
\definecolor{jsonstring}{rgb}{0,0.5,0}
\definecolor{jsonnumber}{rgb}{0.5,0,0.5}
\definecolor{linkblue}{RGB}{0, 51, 153}

\lstdefinelanguage{json}{
    basicstyle=\normalfont\ttfamily,
    showstringspaces=false,
    breaklines=true,
    frame=lines,
    backgroundcolor=\color{gray!5},
    stringstyle=\color{jsonstring},
    morestring=[b]",
    morestring=[d]',
    literate=
     *{0}{{{\color{jsonnumber}0}}}{1}
      {1}{{{\color{jsonnumber}1}}}{1}
      {2}{{{\color{jsonnumber}2}}}{1}
      {3}{{{\color{jsonnumber}3}}}{1}
      {4}{{{\color{jsonnumber}4}}}{1}
      {5}{{{\color{jsonnumber}5}}}{1}
      {6}{{{\color{jsonnumber}6}}}{1}
      {7}{{{\color{jsonnumber}7}}}{1}
      {8}{{{\color{jsonnumber}8}}}{1}
      {9}{{{\color{jsonnumber}9}}}{1}
      {:}{{{\color{jsonkey}:}}}{1}
      {,}{{{\color{black},}}}{1}
      {\{}{{{\color{black}\{}}}{1}
      {\}}{{{\color{black}\}}}}{1}
      {[}{{{\color{black}[}}}{1}
      {]}{{{\color{black}]}}}{1},
}

\usepackage{amssymb}
\usepackage{pifont}
\usepackage{enumitem}
\usepackage{multirow}
\usepackage{subcaption}
\usepackage{graphicx}
\usepackage{appendix}
\usepackage[table]{xcolor}
\usepackage{multicol}
\usepackage{fontawesome5}
\usepackage{hyperref}

\newtcolorbox{DefinitionBox}[1]{
    enhanced,
    colback=white,
    colframe=gray!45,      
    coltitle=white,        
    arc=1.5mm,               
    boxrule=1pt,
    left=3mm, right=3mm, top=2mm, bottom=2mm,
    middle=1mm,            
    toptitle=0.4mm,        
    bottomtitle=0.4mm,     
    title=#1,              
    titlerule=0pt,         
}

\newtcolorbox{CaseStudyBox}{
    enhanced,
    breakable,
    colback=white,
    colframe=blue!80,      
    arc=1.5mm,             
    boxrule=1pt,
    left=3mm, 
    right=3mm, 
    top=3mm,               
    bottom=3mm,
    width=\linewidth,
}

\usepackage[T1]{fontenc}

\usepackage[utf8]{inputenc}

\usepackage{microtype}

\usepackage{inconsolata}

\usepackage{graphicx}

\title{%
\hspace{-0.8cm}
 \begin{tabular}{@{} r l @{}}
    \raisebox{-0.4\height}{\includegraphics[height=1.0cm]{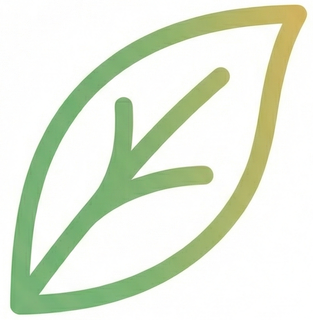}} & 
    \begin{tabular}[c]{l}
       Advancing ESG Intelligence: An Expert-level Agent and \\ Comprehensive Benchmark for Sustainable Finance
    \end{tabular}
  \end{tabular}
}

\author{
 \textbf{Yilei Zhao\textsuperscript{1,2}},
 \textbf{Wentao Zhang\textsuperscript{1}},
 \textbf{Lei Xiao\textsuperscript{3}},
 \textbf{Yandan Zheng\textsuperscript{1}},
\\
 \textbf{Mengpu Liu\textsuperscript{1},}
 \textbf{Wei Yang Bryan Lim\textsuperscript{1,2}\thanks{Corresponding author.}}
\\[4pt]
 \textsuperscript{1}Nanyang Technological University, \\
 \textsuperscript{2}Jinan–NTU Green Technology Research Institute (GreenTRI),
 \textsuperscript{3}Alibaba Group
\\
 \texttt{YILEI002@e.ntu.edu.sg, bryan.limwy@ntu.edu.sg}
\\[4pt]
{\small\ttfamily
\href{https://elainezhao92.github.io/ESGAgent-webpage}
{\textcolor{black}{\faGlobe}\,
 \textcolor{linkblue}{https://elainezhao92.github.io/ESGAgent-webpage}}
}
\\[2pt]
{\small\ttfamily
\href{https://github.com/ElaineZhao92/ESGAgent-and-Benchmark}
{\textcolor{black}{\faGithub}\,
 \textcolor{linkblue}{ElaineZhao92/ESGAgent-and-Benchmark}}
}
}

\begin{document}
\maketitle
\begin{abstract}

Environmental, social, and governance (ESG) criteria are essential for evaluating corporate sustainability, yet professional ESG analysis is hindered by data fragmentation and complex multi-step workflows. Existing ESG benchmarks rely on LLM-generated questions that often suffer from semantic ambiguity, limiting their diagnostic reliability. We introduce \texttt{ESGAgent}, a hierarchical multi-agent system with a domain-oriented toolset for automating end-to-end ESG workflows, and a three-level benchmark derived from 310 corporate sustainability reports, featuring expert-sourced questions with verified, unambiguous answers at Levels 1--2 and full report generation at Level 3. Evaluations across 20 state-of-the-art closed-source LLMs show that \texttt{ESGAgent} achieves 84.15\% accuracy on Levels 1--2, surpassing the strongest baseline by 3.26 points, and produces Level 3 reports with higher factual consistency and analytical depth than commercial deep research systems, validated by both multi-LLM judging and human expert evaluation, showing a robust alignment (Pearson $r=0.551$).
\end{abstract}

\section{Introduction}
\label{sec: intro}

Environmental, social, and governance (ESG) constitutes critical criteria for evaluating corporate sustainability and ethical performance~\cite{kao2023esg}. Transcending discretionary corporate social responsibility~\cite{friede2015esg, del2023esg}, ESG has evolved into a core pillar of modern investment, essential for managing systemic risks and fostering resilience beyond regulatory compliance~\cite{sahin2022environmental}.

\begin{figure}[!ht]
\centering
\includegraphics[width=0.45\textwidth]{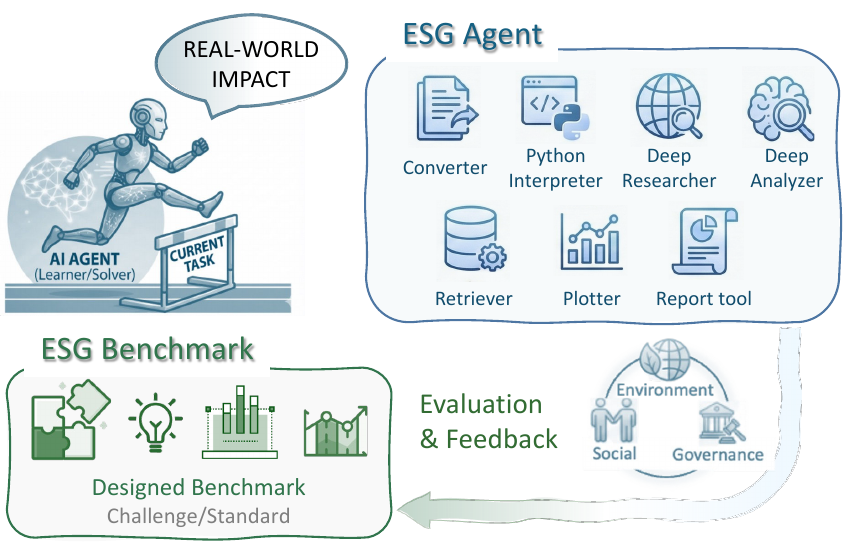} 
\caption{The synergistic interaction between ESG Agent and Benchmark.}
\label{fig: motivation graph}
\end{figure}

However, ESG analysis faces severe fragmentation, with critical data scattered across unstructured sources like sustainability reports and financial filings~\cite{zou2025esgreveal, ong2025towards}. Synthesizing these disparate inputs requires integrating qualitative narratives with quantitative metrics, a process demanding reasoning capabilities that exceed traditional automated tools~\cite{caprioli2024denoising, ang2023predicting}. Although financial LLMs~\cite{jin2025finsight} like BloombergGPT~\cite{wu2023bloomberggpt} show promise, their application to such complex, multi-step ESG workflows remains largely unexplored.

Moreover, existing evaluation frameworks are insufficient for ESG-specific assessment. General-purpose agent benchmarks like GAIA~\cite{mialon2023gaia} and AgentBench~\cite{liu2023agentbench} focus on common-sense tasks and fail to capture domain-specific reasoning. Within the ESG domain, benchmarks such as ESGenius~\cite{he2025esgenius} and MMESGBench~\cite{zhang2025mmesgbench} are limited to static QA and rely on LLM-generated questions from document snippets, where the pipeline produces multiple semantically defensible options, making ground-truth distinctions ambiguous.\footnote{For example, a question from ESGenius asking about the relationship between monitoring and evaluation in climate action plans offers four options, at least two of which are partially supported by the source text. Such ambiguity makes it difficult to reliably differentiate model capabilities.} Furthermore, these benchmarks do not extend to the multi-step, tool-intensive workflows that characterize real ESG analysis~\cite{ni2023chatreport}.

To address these limitations, we introduce \texttt{ESGAgent}, a hierarchical multi-agent system that integrates retrieval augmentation~\cite{guo2024lightrag}, web search, and domain-specific tools (financial modeling, carbon calculation, regulatory alignment, visualization) to execute end-to-end ESG workflows. Our contribution lies in the specialized workflow engineering rather than architectural novelty.

Complementing our agent, we construct a three-level benchmark derived from a corpus of 310 Dow Jones Industrial Average (DJIA) sustainability reports (2010-2024). Levels 1--2 consist of questions grounded in professional certification materials (e.g., chartered financial analyst ESG Investing test) and real corporate disclosures, each manually verified to admit exactly one unambiguous answer. Level 3 requires full analytical report generation.

Evaluations across 20 state-of-the-art LLMs and 4 deep research agents show that \texttt{ESGAgent} achieves 84.15\% accuracy on Levels 1 and 2, surpassing the strongest baseline (Gemini-3-flash, 80.89\%) by 3.26 percentage points. 
For the open-ended Level 3 assessments, we address the challenge of subjective evaluation by employing a multi-LLM-as-a-judge ensemble (GPT-5.2, Gemini-3-flash, Deepseek-r1, and Claude-4.5-sonnet). While individual models exhibit inherent scoring biases, they maintain strong inter-judge agreement (Spearman $\rho \in [0.52, 0.79]$), and the ensemble average shows a significant correlation with human expert assessments (Pearson $r = 0.551$). Under this rigorous evaluation, \texttt{ESGAgent} prioritizes information density, integrating rich visualizations and empirical references. Our primary contributions are summarized as follows:
    
    

\begin{itemize}[left=0.5em]
    \item We develop \texttt{ESGAgent}, a multi-agent system with a domain-specific toolset that automates end-to-end ESG workflows from evidence retrieval to professional report generation.
    
    \item We construct a three-level ESG benchmark with expert-sourced, manually verified questions that addresses the semantic ambiguity limitations of existing LLM-generated ESG benchmarks, while extending evaluation scope to open-ended report generation.
    
    \item We evaluate 20 state-of-the-art closed-source LLMs and 4 deep research agents, validating our benchmark's diagnostic power and demonstrating the superiority of specialized agentic workflows for high-stakes analytical tasks.
\end{itemize}
\section{Related work}

\subsection{ESG Analytics and Methods}
Escalating climate risks and stakeholder pressures have centered ESG in corporate strategy. However, ESG tasks remain inherently complex, demanding cross-domain reasoning, multi-modal data and nuanced value trade-offs~\citep{he2025esgenius}.

Recent research on ESG analytics can be broadly divided into two lines. The first focuses on ESG ratings prediction, utilizing deep learning~\citep{del2023esg}, company graphs~\citep{ang2023predicting}, or rating intervals~\citep{caprioli2024denoising} to improve estimation robustness.
The second line evolved from basic natural language processing to structured semantic extraction. While early models like ClimateBERT~\citep{webersinke2021climatebert} focused on domain-specific language modeling, recent works like ESGReveal~\citep{zou2025esgreveal} and A3CG~\citep{ong2025towards} have advanced towards fact retrieval and greenwashing detection.

Despite these advances, static analysis fails to capture the dynamic nature of ESG decision-making, prompting a shift toward agent-based systems. ChatReport~\cite{ni2023chatreport} leverages RAG for report consistency assessments against the TCFD framework. In contrast to these analysis-centric methods, we develop an agentic framework for end-to-end report generation. However, for existing benchmarks (e.g., ~\cite{schimanski2024climretrieve} and ~\cite{mankour2025climate}), they provide evaluation only on climate dimensions. ESGenius~\citep{he2025esgenius} and MMESGBench~\citep{zhang2025mmesgbench} rely heavily on LLM-generated questions that lack grounding in real-world professional scenarios. In contrast, our proposed ESG Benchmark employs expert-crafted tasks with multimodal evidence, enabling a rigorous evaluation of agentic reasoning in realistic ESG applications.

\subsection{Benchmarking Agentic AI Systems}

Benchmarks have long underpinned AI progress by providing standardized evaluation frameworks~\citep{fei2022searching}, and as AI shifts from static models to agentic systems~\cite{wang2024survey}, their scope has expanded to assess dynamic, goal-directed behaviors~\citep{mialon2023gaia}.
While general-purpose agent benchmarks have advanced rapidly, from task-oriented evaluations such as ToolBench~\citep{qin2023toolllm} and AgentBench~\citep{liu2023agentbench} to interactive environments like WebArena~\citep{zhou2023webarena} and GAIA~\citep{mialon2023gaia}, they primarily measure agents’ reasoning, planning, and tool-use abilities~\cite{yao2022react} in neutral contexts. Yet, these evaluations often fail in vertical domains, where reasoning must be grounded in industry standards and compliance requirements. The ESG domain exemplifies this challenge, demanding agents that align cognitive competence with complex sustainability trade-offs~\citep{he2025esgenius}. To address this, we introduce the ESG Benchmark to evaluate agents in realistic contexts including carbon accounting, policy analysis, and corporate disclosure assessment. This framework assesses the ability to synthesize heterogeneous data into coherent, evidence-based, and ethically consistent decisions.
\section{ESGAgent}

ESG analysis requires multi-step reasoning across fragmented data and evolving standards. Our multi-agent architecture overcomes the limitations of static LLM pipelines by decomposing the workflows into specialized, tool-enabled components, facilitating the flexible retrieval and computation essential for professional report generation.

\subsection{Architecture and Execution Loop}

The agent is designed as a sophisticated, hierarchical system capable of transforming raw, unstructured data into actionable sustainability insights, where the architecture emphasizes high-fidelity information fetch and multi-agent coordination.

\noindent\textbf{Execution Loop.} ESGAgent utilizes a hierarchical "plan-execute-feedback" cycle, where a central planning agent deconstructs user objectives into discrete sub-tasks. These tasks are dispatched to specialized sub-agents via isolated instructions and specific data snippets, ensuring sub-agents remain decoupled from the planner’s global state to maintain operational focus. After each sub-agent completes its operation, the planner evaluates the feedback. If the results are incomplete or unexpected errors occur, the system triggers an objective shift or plan update to recalibrate the next steps.

\begin{figure}[!h]
\centering
\includegraphics[width=0.95\columnwidth]{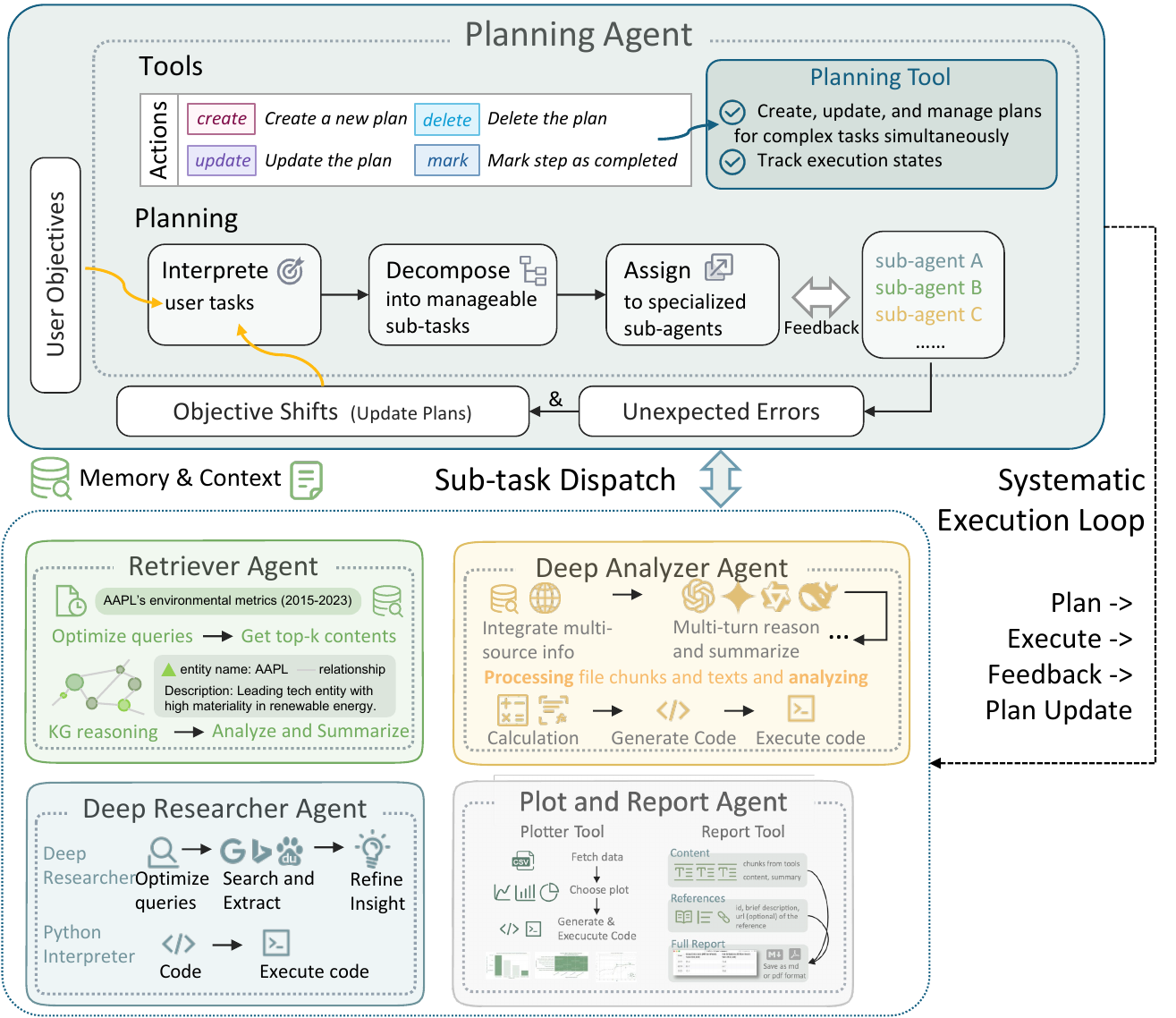} 
\caption{Hierarchical architecture and execution loop}
\label{fig: excution loop}
\end{figure}

\noindent\textbf{Memory and Context Management.} ESGAgent utilizes a \emph{dual-layered memory} architecture. At the orchestrator level, a \emph{global memory} persists as a centralized repository that synthesizes observations into high-level insights~\cite{zhang2025agentorchestra}, utilizing entity resolution to ensure consistency across heterogeneous sources. This global state is strictly decoupled from specialized sub-agents to prevent context noise. Instead, each sub-agent operates within its own \emph{local memory} and toolset. For example, the Retriever executes localized RAG via LightRAG~\cite{guo2024lightrag} to extract evidence from long documents, focusing exclusively on task-specific data snippets. Once the operation concludes, only the refined observations are fed back to the planner to update the global memory and recalibrate the analytical trajectory.

\subsection{ESG-Oriented Toolset}
The agent leverages a versatile toolset categorized into general and ESG-specific tools, where specialized sub-agents dynamically manage and invoke specific utilities.

\begin{figure*}[!t]
\centering
\includegraphics[width=0.95\textwidth]{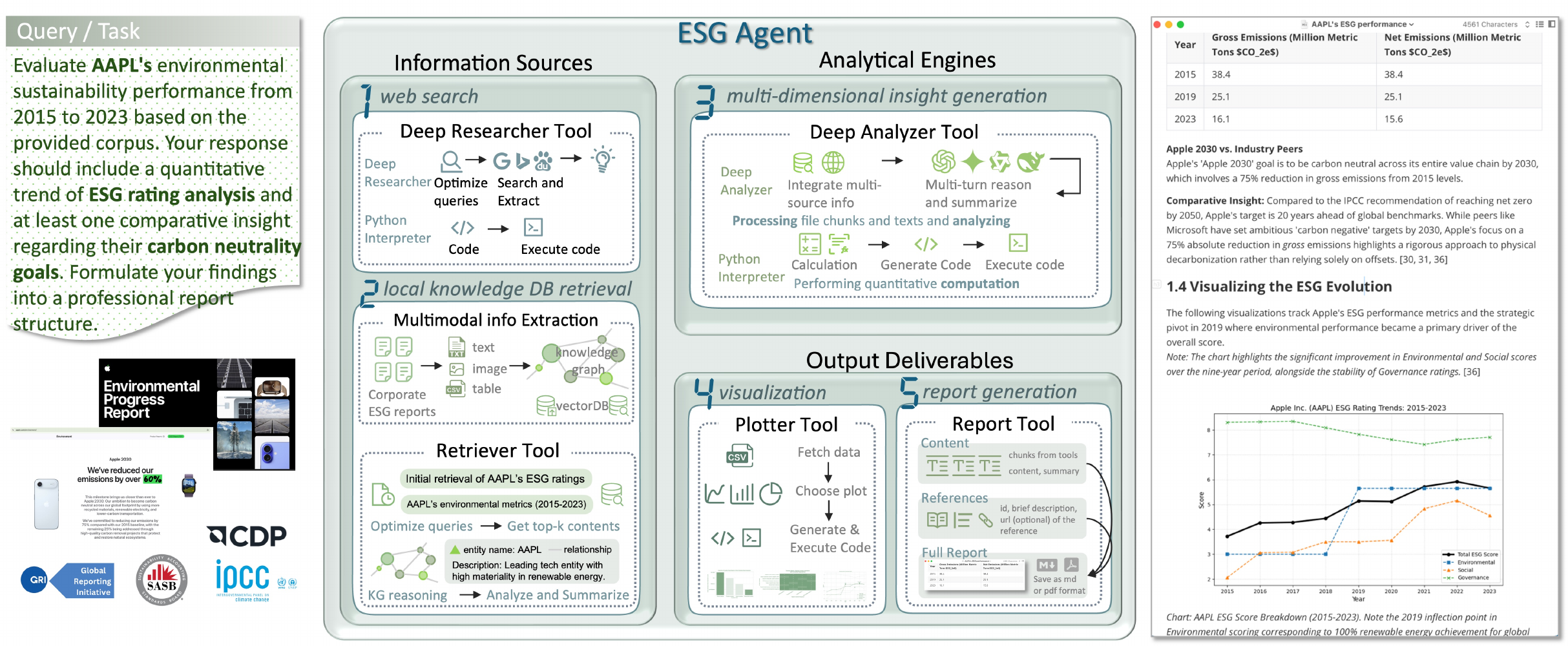} 
\caption{\texttt{ESGAgent} overall architecture and workflow.}
\label{fig: agent architecture}
\end{figure*}

\noindent\textbf{General Tools.} These components comprise the standard toolkit utilized by general-purpose research agents, providing the underlying functionalities for data processing and reasoning.
\begin{itemize}[left=0.5em, label={\small$\blacktriangleright$}]
    \item Converter: Converts multimodal inputs (audio, images, PDFs) into structured text descriptions for downstream processing.

    \item Deep analyzer: Synthesizes information across chunks to generate multi-turn reasoning and high-level narratives.

    \item Python interpreter: Executes autonomous code for rigorous statistical computation and automated data visualization.
\end{itemize}

\noindent\textbf{ESG-specific Tools.} These modules are specialized extensions developed to address the distinct analytical requirements and domain-specific challenges inherent to the ESG sector.
\begin{itemize}[left=0.5em, label={\small$\blacktriangleright$}]
    \item Retriever: Leverages KG-augmented, multi-turn retrieval over local vector databases to contextually enhance analysis and summarization.
    
    \item Deep researcher \& browser: Facilitates real-time acquisition of web-based knowledge to supplement static local corpora with current, targeted information.
    
    \item Plotter: Aligns quantitative data and automates the generation of optimal visualization code to produce high-quality graphical representations.
    
    \item Report tool: Integrates analytical summaries, and graphics into structured Markdown or PDF reports with traceable references for data verifiability.
\end{itemize}

\section{ESG Benchmark}

\begin{figure*}[htbp]
     \centering
     \begin{minipage}[b]{0.48\textwidth}
         \centering
         \includegraphics[width=\textwidth]{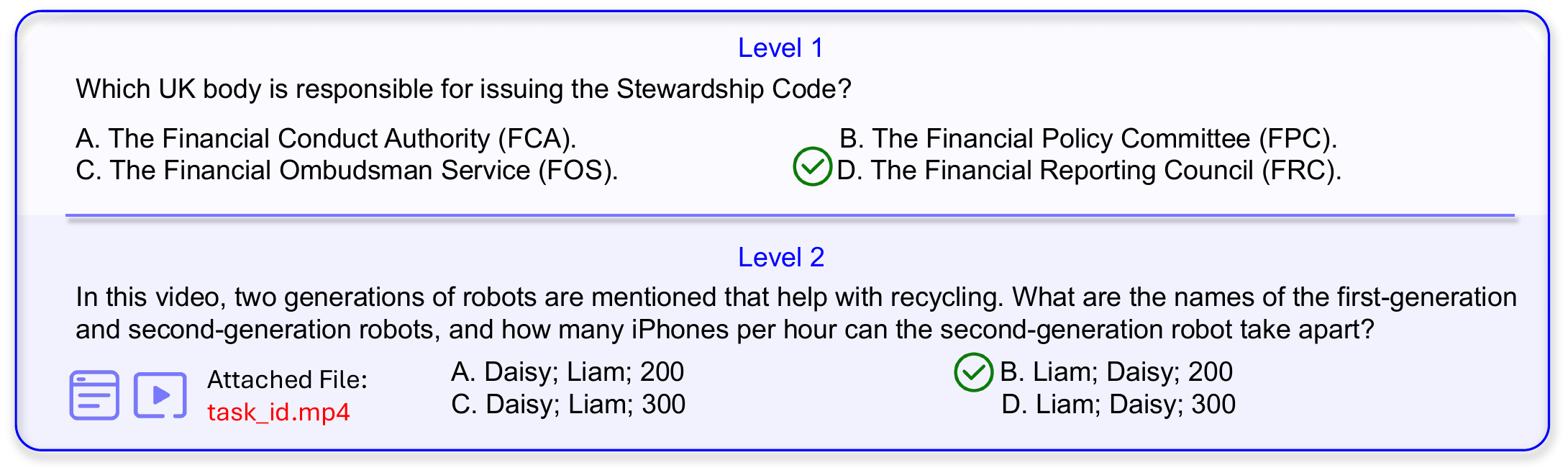}
     \end{minipage}
     \hspace{0.15cm}
     \begin{minipage}[b]{0.48\textwidth}
         \centering
         \includegraphics[width=\textwidth]{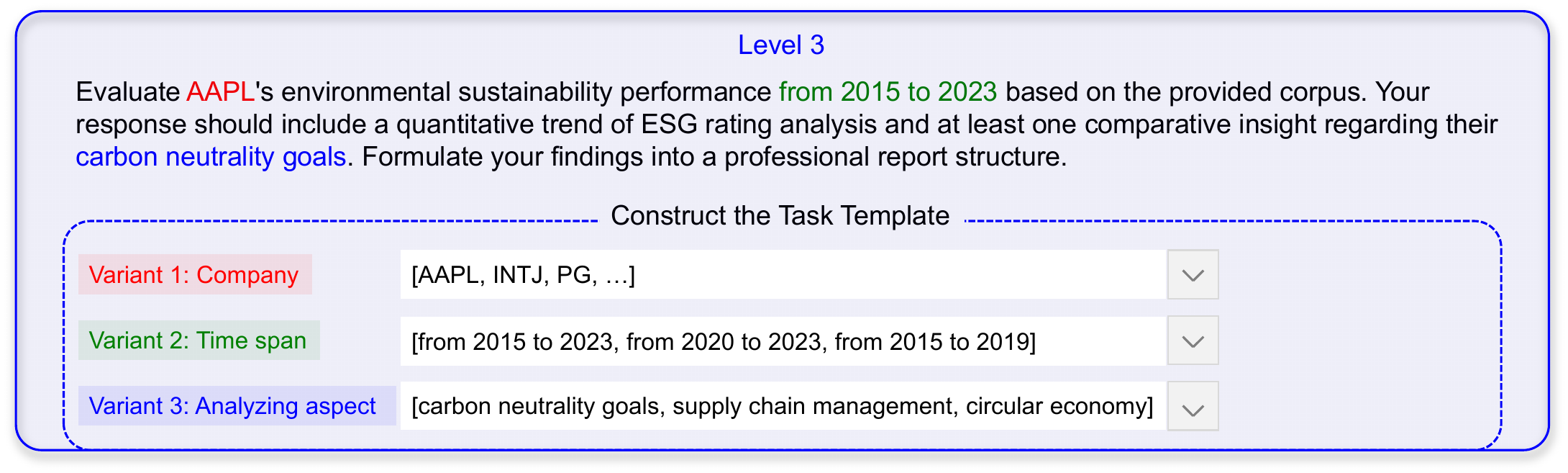}
     \end{minipage}
     \caption{Samples of ESG questions. The benchmark progresses from static questions (Level 1 \& 2) to an open-ended, template-driven architecture (Level 3).}
     \label{fig:three_levels}
\end{figure*}

\subsection{Composition of Benchmark}

\begin{figure}[!ht]
\centering
\includegraphics[width=0.98\columnwidth]{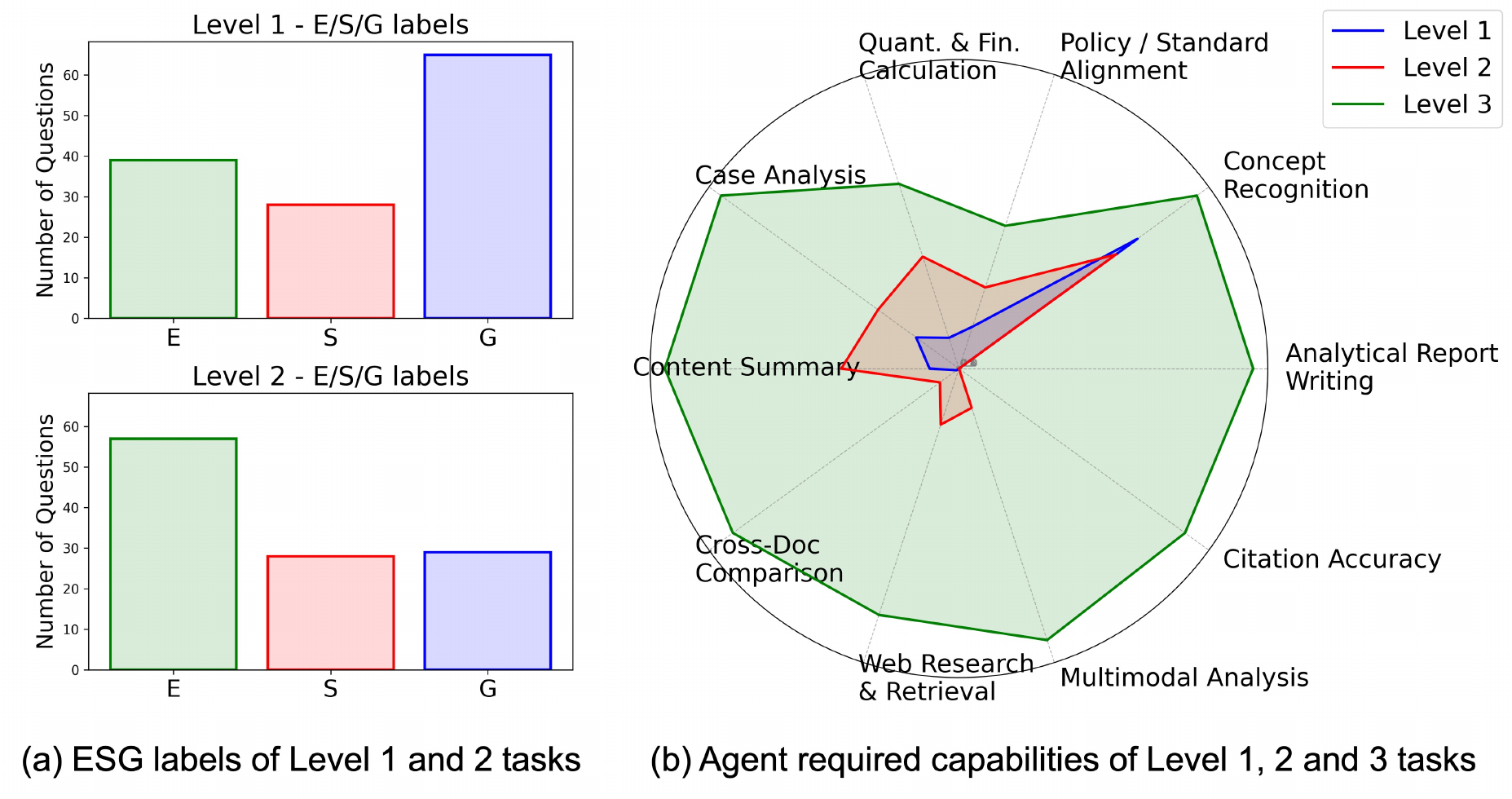} 
\caption{Distribution of ESG labels and required agent capabilities.}
\label{fig: question labels}
\end{figure}

This hierarchical benchmark, spanning from fundamental information extraction to analytical report generation, evaluates the multifaceted capabilities of ESG agents. The statistical profile of the dataset is provided in Table~\ref{tab: question descriptions}, while the distribution of ESG thematic labels and multidimensional capability requirements are categorized in Figure~\ref{fig: question labels}.

\subsubsection{Datasets}
\begin{table*}[!ht]
\centering
\small
\begin{threeparttable}
\begin{tabularx}{0.93\textwidth}{l| >{\hsize=0.8\hsize}X >{\hsize=1.0\hsize}X >{\hsize=1.2\hsize}X}
\toprule
\textbf{Questions} & \textbf{Level 1} & \textbf{Level 2} & \textbf{Level 3}\\
\midrule
\# questions & 132 & 114 & 45 \\
Avg./max. question length & 109.742/705 & 74.974/192 & 67.15/144\\
Avg./max. answer length & 1.008/2 & 1.191/7 & -- / --\\
Avg./max. capabilities per question & 1.341/3 & 2.539/5 & 8.57/10\\
\midrule
Multimodal attachments & \ding{55} & \ding{51} & \ding{51} \\
Question type\footnote{1} & T/F, MC, FIB & T/F, MC, FIB & Open-ended \\
Task type & Atomic tasks & Compositional tasks & Analytical generation tasks\\

\bottomrule
\end{tabularx}
\begin{tablenotes}
\scriptsize
\item[$1$] Question types include true/false (T/F), multiple-choice (MC), fill-in-the-blank (FIB), and open-ended questions, where the first three formats are abbreviated as MC, T/F, and FIB, respectively.
\end{tablenotes}
\end{threeparttable}
\caption{Description of three levels of questions}
\label{tab: question descriptions}
\end{table*}



    
    
\noindent\textbf{Design Principles.} Our benchmark adheres to three core principles. First, tasks are real-world grounded, simulating professional ESG decision-making scenarios like carbon auditing or calculation, and portfolio rebalancing. Second, we prioritize the questions that easy to create, challenging for AI systems. The queries are designed to be deceptively simple, further stressing the need to evaluate LLMs and agents on their general reasoning and comprehensive task-handling abilities.
Finally, the benchmark covers a diverse cognitive range, bridging basic concept recognition with expert-level analytical synthesis. Levels 1 and 2 focus on \textit{factoid} questions with \textit{concise} and \textit{unambiguous} answers, while Level 3 transitions to open-ended analytical tasks, requiring agents to generate \textit{rigorous}, \textit{insightful}, and \textit{professional-grade} ESG reports.

\noindent\textbf{Task Hierarchy.} We categorize questions into three levels based on their reasoning depth.
\begin{itemize}[left=1em]
    \item Level 1 (Easy) focuses on \textit{atomic tasks} that evaluate fundamental ESG literacy. These questions are sourced from authentic practice problems and mock exams of the chartered financial analyst (CFA) institute's certificate in ESG Investing\footnote{https://www.cfainstitute.org/programs/esg-certification}, requiring 1--2 steps to identify or recognize core professional concepts.
    
    \item Level 2 (Medium) involves \textit{compositional tasks} that require 3--5 reasoning steps, where agents should autonomously invoke multiple tools like web search browser, deep analyzer to integrate fragmented information. A key feature of this level is the inclusion of multimedia attachments, such as PDF, CSV, MP3, and MP4 files, which agents must process and analyze to extract critical evidence necessary for accurate task resolution. This level encompasses a diverse range of scenarios, including financial and carbon calculations, case studies and cross-document comparisons.
    
    \item Level 3 (Advanced) focuses on the \textit{comprehensive generation} of enterprise-level ESG reports. These open-ended tasks demand high-level synthesis of multi-dimensional data, requiring the agent to ensure rigorous accuracy and professionalism consistent with global reporting standards (e.g., GRI~\citep{gri2023standards}, SASB~\citep{sasb2023standards}). Beyond textual analysis, agents are expected to design and integrate visualizations such as performance trend charts to enhance the report's depth and clarity.
\end{itemize}

\noindent\textbf{Source and Annotation.} The corpus comprises a total of 310 official ESG and financial reports from Dow Jones Industrial Average (DJIA)\footnote{https://en.wikipedia.org/wiki/Dow\_Jones\_Industrial\_Average} constituents spanning 2010 to 2024, serving as the foundational knowledge base for RAG. Furthermore, all metadata labels, including the corresponding ESG pillar (E, S, or G) and the requisite skill set were manually annotated by domain experts.

\begin{table}[!ht]
\centering
\small
\begin{tabular}{
    >{\raggedright\arraybackslash}p{0.03\textwidth}
    >{\raggedright\arraybackslash}p{0.4\textwidth}
}
\toprule
\textbf{ID} & \textbf{Capability} \\
\midrule
1 & Concept Recognition \& Definition \\
2 & Regulation / Policy / Standard Interpretation \& Alignment \\
3 & Quantitative / Financial Modeling \& Calculation \\
4 & Case Analysis \\
5 & Content Summarization \& Interpretation \\
6 & Cross-Document Comparison \\
7 & Web Research \& Information Retrieval \\
8 & Multimodal Data Analysis \& Generation \\
9 & Citation Accuracy, Faithfulness \& Source Verification \\
10 & Comprehensive Analytical Report Writing \\
\bottomrule
\end{tabular}
\caption{Capability taxonomy for ESG task evaluation.}
\label{tab: capability taxonomy}
\end{table}

\subsubsection{Capabilities}

ESG analysis is characterized by a multidimensional combination of tasks rather than a single operational scenario. To capture this complexity, we define a capability taxonomy that serves as a natural partition of the ESG task space, abstracting real-world workflows into ten structured dimensions as detailed in Table~\ref{tab: capability taxonomy}. This mapping ensures the benchmark evaluates agent proficiency across the entire analytical lifecycle, from foundational recognition to comprehensive report generation.

As illustrated in Figure~\ref{fig: question labels}(b), these capability requirements were manually labeled by human experts across three difficulty levels in the benchmark. The radar chart plots the frequency of these requirements, with each dimension representing the proportion of questions within a level that demand a specific capability. The results highlight a distinct transition: levels 1 -- 2 focus on foundational skills, and level 3 demands high-order reasoning essential for professional-grade ESG analysis.

\section{Experiment}

\begin{table*}[!ht]
\centering
\small
\begin{threeparttable}
    \begin{minipage}[c]{0.63\textwidth}
        \centering
        \resizebox{\textwidth}{!}{
        \begin{tabular}{lcccccccc}
        \toprule
        \multirow{3}{*}{\textbf{Models / Agents}} & \multicolumn{2}{c}{\textbf{Level 1}} && \multicolumn{2}{c}{\textbf{Level 2}} && \multirow{3}{*}{\textbf{Total Acc (\%)}} \\
        \cmidrule{2-3} \cmidrule{5-6}
        & \textbf{\# Corr} & \textbf{Acc (\%)} && \textbf{\# Corr} & \textbf{Acc (\%)} \\
        \midrule
        GPT-5.4 & 114 & 86.36 && 74 & 64.91     && 76.42 \\
        GPT-5.2 & 112 & 84.85 && 67 & 58.77     && 72.76\\
        GPT-5.1 & 106 & 80.30 && 67 & 58.77     && 70.33\\
        GPT-5   & 115 & 87.12 && 75 & 65.79     && 77.24\\
        GPT-4.1 & 110 & 83.33 && 73 & 64.04     && 74.39\\
        GPT-o3  & 114 & 86.36 && 75 & 65.79     && 76.83\\
        GPT-4o  & 109 & 82.58 && 71 & 62.28     && 73.17\\
        \midrule
        Gemini-3.1-pro    & \underline{119} & \underline{90.15} && 76 & 66.67 && 79.27 \\
        Gemini-3.1-flash lite    & 113 & 85.61 && 77 & 67.54 && 77.24 \\
        Gemini-3-pro      & 116 & 87.88 && 76 & 66.67 && 78.05\\
        Gemini-3-flash    & 118 & 89.39 && \underline{81} & \underline{71.05} && \underline{80.89}\\
        Gemini-2.5-pro    & 113 & 85.61 && 75 & 65.79 && 76.42\\
        Gemini-2.5-flash  & 105 & 79.55 && 73 & 64.04 && 72.36\\
        \midrule
        Deepseek-r1  & 115 & 87.12 && 75 & 62.28 && 77.24\\
        Deepseek-v3  & 108 & 81.82 && 71 & 62.28 && 72.76\\
        \midrule
        Claude-opus-4.6     & 116 & 87.88 && 73 & 64.04 && 76.83 \\
        Claude-sonnet-4.6   & 112 & 84.85 && 67 & 58.77 && 72.76 \\
        Claude-opus-4.5     & 113 & 85.61 && 74 & 64.91 && 76.02 \\
        Claude-sonnet-4.5   & 106 & 80.30 && 66 & 57.89 && 69.92 \\
        Claude-haiku-4.5    & 101 & 76.52 && 69 & 60.53 && 69.11 \\
        \midrule
        \rowcolor{cyan!10}
        \textbf{ESGAgent} & \textbf{119} & \textbf{90.15} && \textbf{88} & \textbf{77.19} && \textbf{84.15} \\
        \rowcolor{cyan!10}
        \quad-w/o-deep research & 117 & 88.64 && 75 & 65.79 && 78.05 \\
        \rowcolor{cyan!10}
        \quad -w/o-retriever & 113 & 85.61 && 81 & 71.05 && 78.86 \\
        \rowcolor{cyan!10}
        \quad-w/o-hierarchical planning & 117 & 88.64 && 73 & 64.04 && 77.24 \\
        \rowcolor{cyan!10}
        \quad-w/o-memory management & 114 & 86.36 && 75 & 65.79 && 76.83 \\
        
        \bottomrule
        \end{tabular}}
    \end{minipage}
    \hfill 
    \begin{minipage}[c]{0.34\textwidth}
        \centering
        \includegraphics[width=\textwidth]{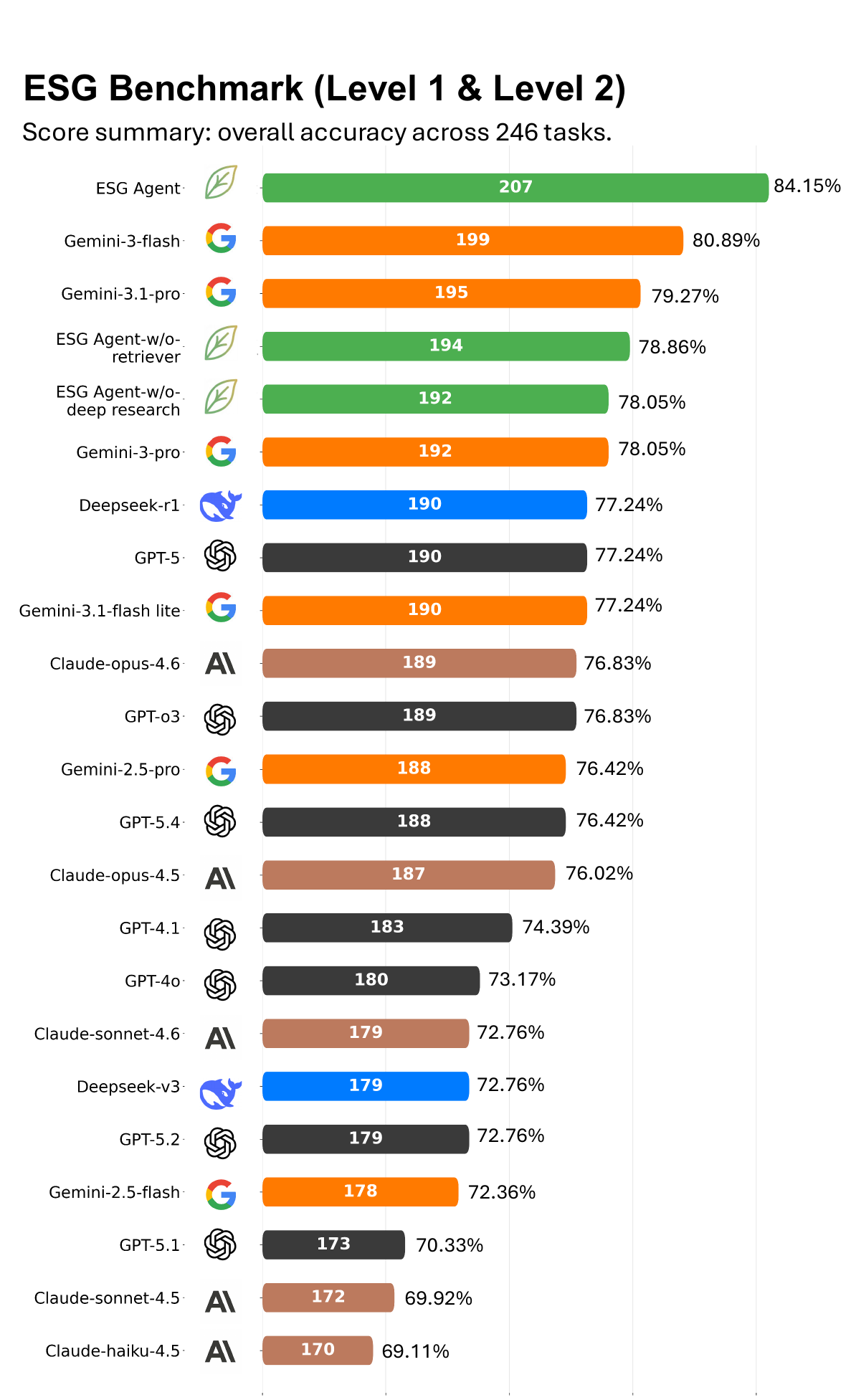}
        \label{fig: ranking_plot}
    \end{minipage}

    \caption{Benchmarking model performance on Levels 1 \& 2. The left panel shows the detailed accuracy across different models; the right panel presents the performance ranking. }
    \label{tab: level1&2 results}
\end{threeparttable}
\end{table*}

\begin{table*}[!ht]
\centering
\small
\begin{threeparttable}
\setlength{\tabcolsep}{3pt} 

\begin{tabularx}{1\textwidth}{l ccl c ccc c ccc cl}
\toprule
\textbf{Level 3} \footnote{5} & \multicolumn{2}{c}{\textbf{Factual Consistency}} & \multirow{2}{*}{\textbf{Judge}}&& \multicolumn{3}{c}{\textbf{Analysis Effectiveness}} && \multicolumn{3}{c}{\textbf{Presentation Quality}} && \textbf{Final} \\
\cmidrule(lr){2-3} \cmidrule(lr){6-8} \cmidrule(lr){10-12}
(45 questions) & Corr. & Faith. &  && Rich. & Comp. & Depth && Coh. & Prof. & Expr. && \textbf{Score}\\ 
\midrule

\multirow{3}{*}{GPT-5.2-deep research} & \multirow{3}{*}{\underline{0.894}} & \multirow{3}{*}{0.765} 
& Multi-LLM && 6.875 & 7.042 & 6.000 && 7.542 & 7.250 & 5.833 && \multirow{3}{*}{6.956}\\
&&& Human && 6.333 & 5.667 & 5.833 && 5.667 & 4.429 & 5.000 \\
&&& Overall \footnote{6} && 6.767 & 6.767 & 5.966 && 7.167 & 6.686 & 5.667 \\

\rowcolor{gray!15}
&&& Multi-LLM && 7.958 & 7.500 & 7.792 && 8.417 & 8.292 & 5.833 && \\
\rowcolor{gray!15}
&&& Human && 6.000 & 5.667 & 5.167 && 5.833 & 5.167 & 3.833 && \\
\rowcolor{gray!15}
\multirow{-3}{*}{Gemini-3-pro-deep research} & \multirow{-3}{*}{0.793} & \multirow{-3}{*}{\underline{0.802}} & Overall && \underline{7.567} & \underline{7.133} & \underline{7.113} && \underline{7.767} & \underline{7.667} & \underline{5.767} && \multirow{-3}{*}{\underline{7.436}} \\

\multirow{3}{*}{Grok-deep research} & \multirow{3}{*}{0.803} & \multirow{3}{*}{0.760} 
& Multi-LLM && 7.000 & 7.042 & 6.125 && 7.333 & 7.042 & 6.333 && \multirow{3}{*}{6.812} \\
&&& Human && 5.833 & 5.833 & 4.857 && 5.500 & 5.667 & 3.143 \\
&&& Overall && 6.766 & 6.800 & 5.871 && 6.967 & 6.767 & 5.695 \\

\multirow{3}{*}{Perplexity-deep research} & \multirow{3}{*}{0.734} & \multirow{3}{*}{0.661} & Multi-LLM && 7.083 & 7.000 & 6.708 && 7.583 & 7.333 & 6.083 && \multirow{2}{*}{6.650} \\
&&& Human && 5.600 & 5.667 & 5.333 && 5.667 & 4.500 & 2.333 \\
&&& Overall && 6.787 & 6.733 & 6.433 && 7.200 & 6.767 & 5.333 \\

\rowcolor{cyan!10}
&&& Multi-LLM && 8.107 & 7.459 & 7.825 && 8.167 & 8.292 & 7.125 && \\
\rowcolor{cyan!10}
&&& Human && 6.833 & 7.333 & 7.167 && 7.333 & 7.333 & 6.833 && \\
\rowcolor{cyan!10}
\multirow{-3}{*}{\textbf{ESGAgent}} & \multirow{-3}{*}{\textbf{0.930}} & \multirow{-3}{*}{\textbf{0.805}} & Overall && \textbf{7.852} & \textbf{7.434} & \textbf{7.693} && \textbf{8.000} & \textbf{8.100} & \textbf{7.067} && \multirow{-3}{*}{\textbf{7.937}} \\

\bottomrule
\end{tabularx}

\begin{tablenotes}[flushleft]
    \footnotesize
    \item [5] Within the \textit{factual consistency} dimension, Corr. and Faith. represent citation correctness and textual faithfulness. Within \textit{analysis effectiveness}, Rich. denotes information richness, Comp. denotes content completeness, and depth denotes analytical depth. Within \textit{presentation quality}, Coh. represents logic coherence, Prof. represents language professionalism, and Expr. represents chart and graph expressiveness.

    \item [6] The Overall score represents the final consensus by treating each of the four LLM agents and the human expert as five independent judges with equal voting weight. $\text{Overall}=\frac{4 \times \text{Multi-LLM-Avg.} + 1 \times \text{Human}}{5}$. Note that for objective \textit{factual consistency} metrics, only automated scores are reported to ensure deterministic assessment and preclude subjective bias.
\end{tablenotes}

\caption{Multi-dimensional evaluation of model performance on Level 3.}
\label{tab: level3 results}
\end{threeparttable}
\end{table*}

\subsection{Evaluation}

We evaluate \texttt{ESGAgent} across three benchmark tiers. Levels 1 and 2 deterministic tasks are measured via standard accuracy $\text{Acc}=\frac{\#\text{correct}}{\#\text{total}}$.

Level 3 open-ended tasks are evaluated through a hybrid ensemble framework that integrates \textbf{human expertise with a multi-LLM-judge system}. This architecture mitigates single-model bias and ensure evaluation fairness by averaging scores across diverse LLMs and human. Under this framework, we assess the agent's performance across three core dimensions, ranging from objective factual verification to subjective quality analysis.

\subsubsection{Factual consistency}

It quantifies the integrity of information retrieved from external knowledge bases, ensuring the agent stays faithful to the original context and maintains factual correctness to mitigate hallucinations~\cite{ming2024faitheval}. Details are in Appendix~\ref{appendix: evaluation deinition}.
\begin{itemize}[left=0.5em, label={\small$\blacktriangleright$}]

    \item \textit{Correctness} verifies whether individual attributions are semantically and explicitly supported by the cited evidence~\cite{xu2025citecheck}.
    \begin{DefinitionBox}{Definition of Correctness}
        Let $s$ be a generated statement, and let $A = \{a_1, a_2, \dots, a_n\}$ be the set of documents retrieved as context.
        
        A citation pair $(s, a_j)$ satisfies \textit{citation correctness}, iff $a_j\in \mathcal{A}$ and $a_j \vDash s$, where $\vDash$ denotes a \textit{semantic entailment} relationship, such that any model satisfying $a_j$ necessarily satisfies $s$.
    \end{DefinitionBox}

    \item \textit{Faithfulness} evaluates whether the attribution reflects the actual process leading to the answer, ensuring the agent stays faithful to the retrieved context during generation~\cite{wallat2024correctness}.
    \begin{DefinitionBox}{Definition of Faithfulness}
        Let $s$ be a generated statement, and let $A = \{a_1, a_2, \dots, a_n\}$ be the set of documents retrieved as context.
    
        We call $(s, a_j)$ a \textit{faithful citation} if all the following conditions are met:
        \begin{itemize}[left=0.2em]
            \item[*] \textit{Existence:} The document must be part of the retrieved set, $a_j \in A$.
            \item[*] \textit{Support (Correctness):} $s$ is logically supported by $a_j$, and factually grounded in the source.
            \item[*] \textit{Causality:} $s$ is causally impacted by $a_j$. The model actually used $a_j$ to generate $s$, rather than relying on its internal parametric memory or hallucinating the link.
        \end{itemize}
    \end{DefinitionBox}
\end{itemize}

\subsubsection{Analysis Effectiveness} 

It assesses the agent's ability to generate domain-specific insights and synthesize heterogeneous evidence into analytically grounded content. The evaluation focuses on three dimensions:

\begin{itemize}[left=0.5em, label={\small$\blacktriangleright$}]
    \item \textit{Information richness} quantifies the density of domain-specific facts and numerical evidence, distinguishing high-resolution insights from generic template replication.

    \item \textit{Content completeness} measures thematic coverage against industry standards (e.g., GRI, SASB) to ensure all essential ESG pillars are addressed without omission.

    \item \textit{Analytical depth} assesses the construction of logical causal chains, specifically the agent's ability to link raw performance data to strategic risks and financial materiality.
\end{itemize}

\subsubsection{Presentation quality} 

It examines the structural coherence and communicative professionality of the final output, focusing on linguistic precision and the effective integration of visual elements.
    \begin{itemize}[left=0.5em, label={\small$\blacktriangleright$}]
        \item \textit{Logic coherence} evaluates the structural integrity and narrative flow of the report relative to professional standards.

        \item \textit{Language professionalism} assesses academic tone and the precise use of domain-specific terminology.

        \item \textit{Chart expressiveness} measures the efficacy of visual elements in effectively conveying quantitative trends and their alignment with the textual analysis.
    \end{itemize}

To ensure a consistent scale across metrics, \textit{Correctness} and \textit{Faithfulness}, which are percentage-based scores represented in $[0,1]$, are multiplied by 10 to align with the other dimensions, which are rated on a $[0,10]$ scale by multiple LLM judges. The final score is obtained by averaging the evaluations from four LLM judges and human experts to improve overall evaluation reliability. The prompts used for LLM-based evaluation are provided in Appendix~\ref{appendix: level3 evaluation def.}.


\subsection{Agent Configuration}
Our architecture leverages a multi-model approach to optimize specific sub-agent capabilities. The reasoning heavy components, including the \texttt{ESGAgent} main pipeline, deep researcher, deep analyzer, plotter tool, and reporter, are powered by the \texttt{gemini-3-flash-preview} model. Meanwhile, the interactive browser use agent are facilitated by \texttt{gpt-4.1} and \texttt{computer-use-preview(4o)}. To maintain control over the agent's trajectory and mitigate potential infinite loop scenarios, we impose a rigid step budget ($m$). Specifically, set \(m=50\) for the \texttt{ESGAgent} main pipeline and browser use agent, and \(m=3\) for the deep researcher agent, deep analyzer agent.

\subsection{Main Results}

\subsubsection{Baselines}
For Level 1 and Level 2 evaluations, we compare the \texttt{ESGAgent} against a comprehensive suite of state-of-the-art closed-source LLMs. This includes the \textit{GPT family} (e.g., GPT-5.4, 5.2, 5, etc), the \textit{Gemini family} (including the latest Gemini-3.1-pro and Gemini-3-pro), Deepseek family~\cite{liu2024deepseek, guo2025deepseek}, and so on. These models represent the current frontier in general semantic understanding and factual retrieval capabilities.

For Level 3 tasks, which involve high-level open-ended analysis, we narrow our comparison to specialized agents equipped with deep research capabilities. These baselines include Grok deep research~\cite{grok2025deep}\footnote{https://grok.com/}, Perplexity deep research~\cite{perplexity2025deep}\footnote{https://www.perplexity.ai/?model\_id=deep\_research}, GPT-5 deep research, and Gemini-3-pro deep research~\cite{gemini2025deep}. This setup is designed to evaluate the competitiveness of our system within specialized ESG analysis scenarios compared to general-purpose deep research tools.

\subsubsection{Analysis of the ESGAgent performance}

As summarized in Tables \ref{tab: level1&2 results}, \ref{tab: level3 results} and \ref{tab: level3 detailed results}, \texttt{ESGAgent} consistently outperforms general-purpose baselines, with the performance gap widening as task complexity increases. Specifically, in level 1 and 2 tasks, the \texttt{ESGAgent} achieves an aggregate accuracy of 84.15\%, surpassing the strongest baseline (80.89\%) by a significant margin.

Fine-grained analysis in Table~\ref{tab: level3 detailed results} further reveals \texttt{ESGAgent}'s dominance across most capability categories, particularly in specialized tasks such as financial modeling. Moreover, while general-purpose models often exhibit a performance dip in multi-modal settings, \texttt{ESGAgent} demonstrates superior cross-modal robustness, achieving a 74.6\% accuracy that significantly exceeds the 69.8\% reached by best-performed models.

The superiority of the \texttt{ESGAgent} in level 3 tasks is underpinned by the deep researcher tool and RAG architecture that fuses heterogeneous knowledge sources like local corpora and dynamic web-derived streams. This framework operates through a hierarchical planning mechanism that decomposes complex inquiries into a sequence of verifiable sub-steps. By orchestrating specialized analytical toolsets, the system not only ensures rigorous factual consistency but also achieves a level of analytical richness.

\begin{table}[!t]
\centering
\small
\setlength{\tabcolsep}{3pt}

\begin{tabular}{
    @{}
    p{0.30\columnwidth}
    >{\centering\arraybackslash}p{0.11\columnwidth}
    >{\centering\arraybackslash}p{0.13\columnwidth}
    >{\centering\arraybackslash}p{0.11\columnwidth}
    >{\centering\arraybackslash}p{0.10\columnwidth}
    >{\centering\arraybackslash}p{0.11\columnwidth}
    @{}
}
\toprule
Agent & \# page & \# word & \# chart & \# ref. & \# cite. \\
\midrule
GPT-5.2-DR      & 5.3 & 2137.2 & 1.83 & 6.8  & 34.3 \\
Gemini-3-pro-DR & 11  & 3072.0 & 2.83 & 19.2 & 24.3 \\
Grok-DR         & 4.2 & 1119.8 & 2.83 & 7.2  & 13.2 \\
Perplexity-DR   & 6.7 & 1867.7 & 2.83 & 8.2  & 23.5 \\
\rowcolor{cyan!10}
\textbf{ESGAgent} & 7 & 1978.3 & 3.50 & 18 & 38.3 \\
\bottomrule
\end{tabular}

\caption{Statistical analysis of generated reports.}
\label{tab: report properties}
\end{table}

While agents like \textit{Gemini-3-pro-deep research} tend to produce longer reports, the statistical data in Table~\ref{tab: report properties} demonstrates that the \texttt{ESGAgent} excels in content quality and multi-modal richness. With a leading average of charts and citations, our agent provides more actionable and verifiable insights per page. This shift from verbosity to visual and empirical depth highlights the efficacy of our hierarchical planning and specialized tool-use in synthesizing professional ESG intelligence. Detailed case studies and further analysis are in Appendix~\ref{appendix: case_studies}.

\subsubsection{Diagnostic efficacy of ESG Benchmark}
The ESG Benchmark serves as a sophisticated diagnostic framework, distinguishing between surface-level linguistic recognition and deep professional synthesis. The performance degradation observed in frontier models like GPT-5.2, which declines significantly when transitioning from level 1 atomic tasks to level 2 compositional reasoning, underscores the benchmark's efficacy in assessing foundational ESG concepts, regulatory comprehension, and the ability to navigate complex decision-making within domain-specific scenarios.

Level 3 tasks demand high-resolution insights that distinguish genuine analysis from the generic boilerplate typical of general models. By testing three main dimensions, the framework ensures reports meet the strict accuracy and traceability standards required for professional ESG auditing. 

To validate the level 3 evaluation framework, we perform a dual-correlation analysis. Firstly, we measure the \textbf{human-LLM alignment} using Pearson correlation ($r$) between the human scores and multi-LLM average. Across all dimensions, the coefficients average 0.551 (peaking at 0.640 for Expr.), indicating the automated scoring significantly aligns with professional analyst standards. Secondly, we assess \textbf{inter-judge agreement} by calculating the Spearman correlation matrix among the four LLM judges. The coefficients range from 0.52 to 0.79, suggesting that despite diverse model architectures, the LLMs converged on a consistent scoring logic under the unified evaluation prompt.


\subsection{Detailed Empirical Analysis}

\noindent\textbf{Ablation studies.} To further investigate the contribution of different components within \texttt{ESGAgent}, we conduct ablations from two perspectives, \textit{tool availability} and \textit{agent architecture}, as shown in Table~\ref{tab: level1&2 results}. For tool ablations, removing either the \textit{deep researcher} or \textit{retriever} consistently degrades performance, demonstrating the importance of complementary external research and document retrieval. For architectural ablations, disabling \textit{hierarchical planning} causes a substantial drop on Level 2 tasks, highlighting the importance of structured decomposition for complex reasoning, while removing \textit{memory management} yields the lowest overall accuracy, indicating the value of retaining and reusing intermediate information. Overall, the full \texttt{ESGAgent} achieves the strongest performance, validating the complementary contributions of both its toolset and architectural design.

\begin{figure}[!hb]
\centering
\includegraphics[width=0.48\textwidth]{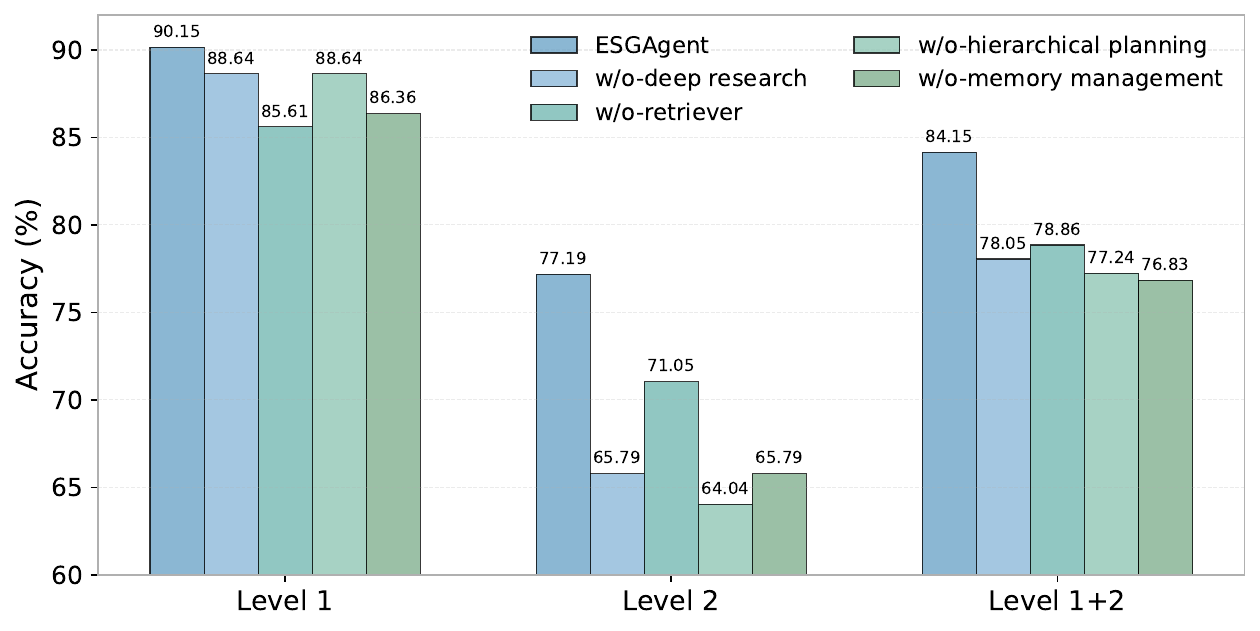} 
\caption{Comparison of \texttt{ESGAgent} and its variants.}
\label{fig: variant_results}
\end{figure}

\noindent\textbf{Efficiency analysis.} We evaluated the operational efficiency of \texttt{ESGAgent} across three benchmark levels. Level 1 tasks average 2.59 minutes with 5k token consumption. Level 2 tasks average 4.22 minutes (1–30 minute range) and 25k tokens; the increased latency and variance are primarily attributed to multimedia (e.g., MP3, MP4) conversion and analysis. Level 3 tasks, involving complex long-horizon scenarios, require approximately 10 minutes and 100k tokens. Compared to monolithic baselines, our hierarchical architecture optimizes resource allocation, delivering significant performance gains while maintaining operational costs comparable to commercial research agents.

\section{Conclusion}


In this work, we introduce \textbf{ESGAgent} and a comprehensive three-level \textbf{ESG Benchmark} to address the complexities of automated sustainability analysis. Empowered by hierarchical planning and domain-oriented tools, our agent transforms fragmented disclosures into professional reports, while the benchmark establishes a rigorous testbed for evaluating high-stakes ESG decision-making. Extensive evaluations demonstrate the effectiveness of ESGAgent across both compositional reasoning and open-ended report generation.

\section*{Acknowledgments}
This research is supported by the RIE2025 Industry Alignment Fund – Industry Collaboration Projects (IAF-ICP) (Award I2301E0026), administered by A*STAR, as well as supported by Alibaba Group and NTU Singapore through Alibaba-NTU Global e-Sustainability CorpLab (ANGEL). It is also supported, in part, by Jinan-NTU Green Technology Research Institute (GreenTRI).

\section*{Limitations}

Despite the efficacy of \texttt{ESGAgent}, our system is subject to the inherent constraints of current LLM-based frameworks. Although our hierarchical planning reduces error propagation, the agent may still generate plausible but factually incorrect statements or miscompute complex financial ratios. Second, the agent's performance is bounded by the availability and clarity of corporate disclosures. In cases where ESG reports are vague or non-standardized, the agent may struggle to synthesize verifiable insights, necessitating manual verification of the source documents.

\newpage
\clearpage
\bibliography{main}
\clearpage
\newpage
\newpage
\appendix
\appendixpage
\label{sec: appendix}
\renewcommand\theequation{\Alph{section}.\arabic{equation}}
\renewcommand\thefigure{\Alph{section}.\arabic{figure}}    
\renewcommand\thetable{\Alph{section}.\arabic{table}}
\renewcommand{\thesection}{\Alph{section}}
\renewcommand{\thesubsection}{\Alph{section}.\arabic{subsection}}

\section{Motivation}
\subsection{Why this agent?}
ESG analysis presents substantial operational challenges due to the highly fragmented nature of ESG data, including unstructured texts and a wide array of document formats, spanning both qualitative and quantitative information. In practice, ESG evaluation is not a single-task procedure but a complex workflow that involves extracting relevant disclosures, identifying risks, performing quantitative calculations, aligning reports with evolving regulatory standards, and ultimately generating coherent assessments or disclosures.

Traditional static pipelines or standalone LLMs are unable to meet the demands of real-world ESG analysis, which requires flexible retrieval, multi-step reasoning and tool-based computation. The need for automated ESG analysis is becoming increasingly urgent, driven by the pressure on enterprises and regulators to produce scalable, accurate, and timely assessments. In contrast, the multi-agent architecture can naturally support these requirements by breaking down complex ESG tasks into smaller components, invoking specialized tools when needed.

\subsection{Why this benchmark?}


Although progress has been made in evaluating ESG-related language tasks, existing benchmarks~\citep{he2025esgenius,zhang2025mmesgbench} remain insufficient for assessing the types of domain-grounded decision-making required in real sustainability analysis, as their questions are largely detached from practical ESG scenarios and therefore cannot meaningfully reflect how well LLMs or agent systems perform in realistic ESG workflows.

Beyond addressing the need for domain-grounded ESG evaluation, our benchmark also serves as a meaningful testbed for assessing the general capabilities of LLM-based agents. Many competencies exercised in ESG analysis, such as in-depth reasoning, quantitative computation, cross-modality understanding, are fundamental capabilities required of agentic systems more broadly~\citep{mialon2023gaia}.

\section{Implemental Details}
\label{appendix: evaluation}

\subsection{GenAI Usage}

We acknowledge the use of GenAI tools in the methodology of this method. For paper writing, LLMs were utilized strictly only for grammatical polishing and stylistic refinement. Methodologically, we utilized external LLM APIs to power the reasoning and evaluation capabilities of the agents.

\subsection{Evaluation of Factual Consistency}
\label{appendix: evaluation deinition}
For the factor consistency dimension, we evaluate the objective alignment between the agent's generated claims and the retrieved evidence. This dimension is critical for ensuring that the agent's reasoning remains grounded in factual data rather than parametric knowledge. To quantify the alignment, we specially measure citation correctness, which verifies the semantic entailment of the claim by the source document, and citation faithfulness, which assesses the extent to which the generation is supported strictly by the retrieved context. 

We quantify these metrics across the generated reports. Consider a generated report that contains a total of $c$ inserted citations in its body context. We denote $c_{cor}$ and $c_{faith}$ as the counts of citations that satisfy the definitions of \textit{correctness} and \textit{faithfulness}, respectively. Then the scores are calculated as the ratios of valid citations to the total number of citations: $\text{Citation Correctness} = \frac{c_{cor}}{c},\  \text{Citation Faithfulness} = \frac{c_{faith}}{c}$.
If a citation fails to support the generated claim (incorrect), it inherently fails to represent the source accurately (unfaithful). Consequently, any citation judged as incorrect is automatically penalized in the faithfulness evaluation.

The correctness and faithfulness results are shown in Table~\ref{tab: level3 results}. We observe that although some deep research agents produce lower-quality content, e.g., \texttt{gpt-5.2} performs only moderately on analysis effectiveness and presentation quality, it still achieves a relatively high citation correctness score. This is mainly because the reports generated by this agent include fewer references overall. As reported in \ref{tab: report properties}, it uses 6.8 citations on average, but are repeatedly inserted up to 34.3 times throughout the report, meaning each reference is reused about 5 times on average. This reuse pattern can inflate both correctness and faithfulness metrics. Therefore, to evaluate the overall quality of generated reports, \textit{it is necessary to consider all three aspects together rather than relying on a single metric}.

\onecolumn
\subsection{Evaluation of Analysis Effectiveness and Presentation Quality}
\label{appendix: level3 evaluation def.}

\begin{multicols}{2}
For analysis effectiveness and presentation quality evaluation, we employ an LLM-as-a-Judge framework. Recognizing that evaluations driven by a single LLM may be subject to specific biases or variance, we adopt a multi-judge ensemble approach, and provide results of different judges in Table~\ref{tab: level3 detailed results}. The detailed instruction set is formalized in the following system prompt.
\end{multicols}

\textbf{System Prompt Template:}

\begin{lstlisting}[basicstyle=\small\ttfamily, breaklines=true, breakatwhitespace=false, columns=fullflexible, keepspaces=true, frame=single, xleftmargin=8pt, xrightmargin=8pt, rulecolor=\color{gray!50}, backgroundcolor=\color{gray!10}]
You are a senior financial analyst and professional ESG auditor specializing in corporate sustainability reporting. Your task is to rigorously provide objective, high-standard review of ESG-related content, maintaining a critical perspective that distinguishes professional analytical depth from generic or superficial reporting.

<intro>
As a seasoned expert, you leverage the following core competencies:
1. Deep Financial Literacy: You possess a rigorous understanding of market mechanisms, allowing you to instantly distinguish material risks from superficial ESG buzzwords.
2. Auditing Instinct: You are trained to trace every claim to its source, naturally detecting logical gaps or post-rationalized arguments that lack evidentiary basis.
3. Framework Proficiency: You are fluent in global reporting standards and professional prose, ensuring every disclosure meets the structural rigor of institutional-grade reports.
</intro>

<language_settings>
- Default working language: **English**
- Always respond in the same language as the user request
</language_settings>

<inputs>
You will be provided the following context as inputs:
1. <query>: The original research prompt or question that initiated the report generation. It specifies the target company, the required analysis areas, and any specific constraints or formatting instructions.
2. <generated report>: The generated ESG analysis to be evaluated.
</inputs>

<task>
Your primary objective is to evaluate the analytical quality and professional presentation of an AI-generated ESG analyzing report. You will act as a critical reader to determine if the report meets the high standards required for institutional-grade financial analysis.
</task>

<task_rules>
To ensure an objective and professional review, you must follow these rules:
1. Expert Perspective: Evaluate the content from perspectives of a senior analyst.
2. Strict Dimension-wise Scoring: For each defined dimension, provide an integer score from 1 to 10. A score of 5 represents a standard, acceptable AI output; scores above 8 require exceptional professional synthesis.
3. Justification via Examples: You must support every score with a specific justification.
</task_rules>


<evaluation_dimensions>
1. Analysis Effectiveness
    - Information richness: Quantifies data granularity and the density of domain-specific facts. Does the agent provide high-resolution insights, or does it merely replicate generic templates found in referenced documents?
    - Content completeness: Measures the thematic scope against standard ESG frameworks (e.g., GRI, SASB). Are all essential pillars addressed comprehensively without significant omission?
    - Analytical depth: Evaluate inferential synthesis. Reward the construction of logical causal chains that link ESG performance to strategic risks and long-term financial materiality.

2. Presentation Quality
    - Logic coherence: Assesses the structural integrity and narrative flow of the report. Are the transitions between sections seamless and aligned with professional reporting standards?
    - Language professionalism: Evaluates the tone and precision of domain-specific terminology. Is the prose clear and stylistically consistent throughout the entire document?
    - Chart expressiveness: Measures the efficacy of quantitative visualizations. Do the generated charts intuitively convey complex trends and align directly with the accompanying analysis?
</evaluation_dimensions>


<output>
Return your evaluation strictly in the following JSON format:
{
  "analysis_effectiveness": {
    "richness": {"score": 0, "justification": ""},
    "completeness": {"score": 0, "justification": ""},
    "depth": {"score": 0, "justification": ""}
  },
  "presentation_quality": {
    "coherence": {"score": 0, "justification": ""},
    "professionalism": {"score": 0, "justification": ""},
    "expressiveness": {"score": 0, "justification": ""}
  }
}

</output>
\end{lstlisting}

\begin{multicols}{2}
\section{Detailed Experiment Results}
To provide a granular assessment of LLM capabilities, we categorize questions based on the taxonomy defined in Fig.~\ref{fig: question labels}. Furthermore, we report the accuracy across different modalities, specifically distinguishing between single modal (text-only) and multimodal (images, tables, audio, and video) tasks to evaluate the models' cross-modal performance.

\subsection{Detailed results of Level 2 Tasks}

\begin{figure*}[!htb]
\centering
\includegraphics[width=0.98\textwidth]{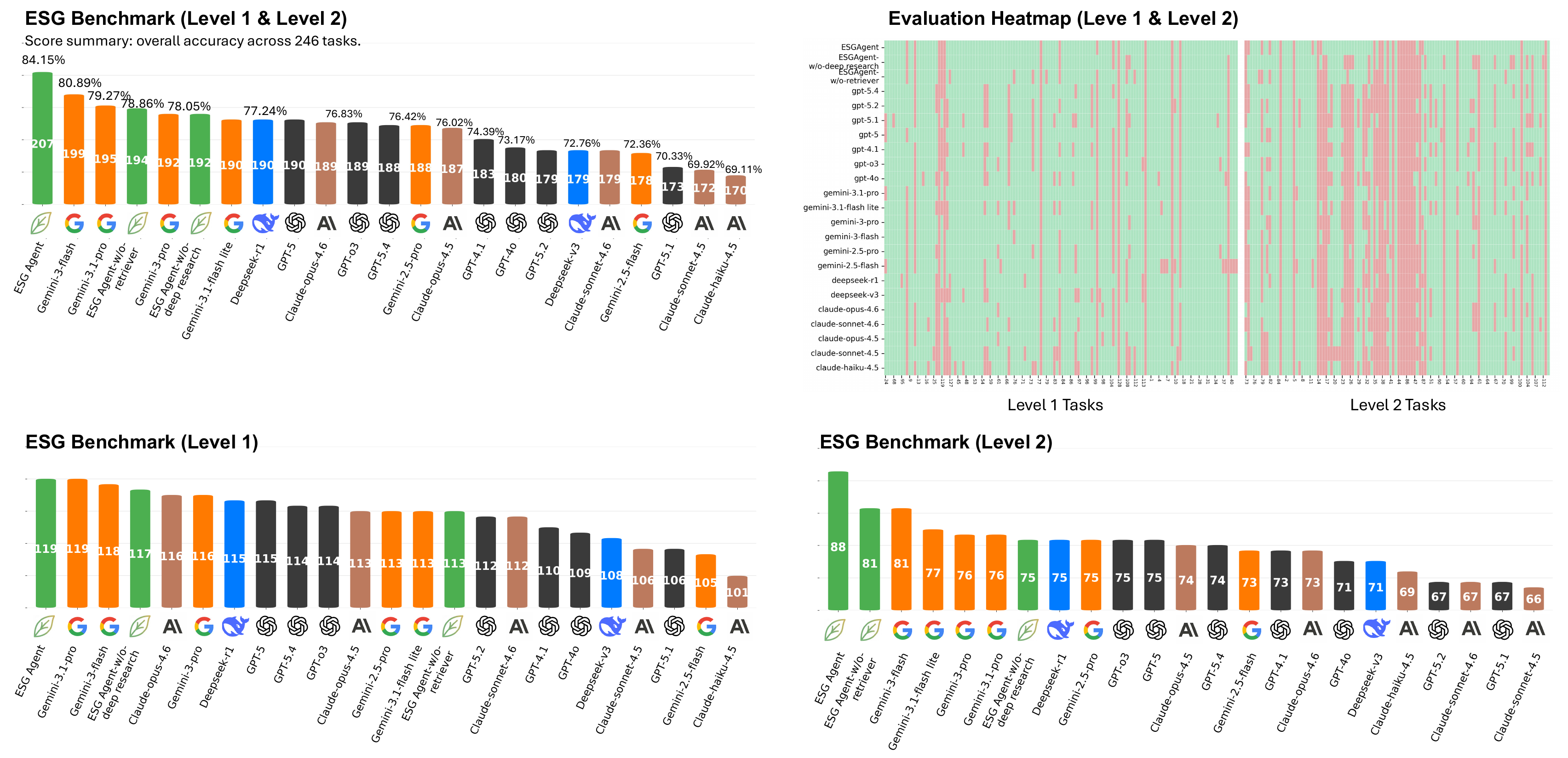} 
\caption{(Top-left, Bottom-left, Bottom-right) Ranking of different models on level 1 and 2 tasks. (Top-right) Evaluation heatmap on level 1 and 2 tasks (18 models $\times$ 246 questions).}
\label{fig: heatmap_rankplot}
\end{figure*}

Figure~\ref{fig: heatmap_rankplot} provides a comprehensive evaluation of 18 different models across 246 tasks. The analysis is divided into four panels.

Ranking Analysis (top-left, bottom-left, bottom-right): The Top-left panel confirms \texttt{ESGAgent}'s dominance with 84.15\% accuracy, outperforming runners-up like \texttt{gemini-3-flash}. Comparing level 1 (Bottom-left) versus level 2 (Bottom-right) reveals a significant performance decay across all models, reflecting the steep difficulty spike associated with complex reasoning tasks.

Evaluation Heatmap (top-right): Visualizing correct (green) versus incorrect (red) responses reveals a contrast between difficulty tiers. While level 1 shows high success rates, the level 2 region exhibits dense red patterns, reflecting increased task complexity. Notably, \texttt{ESGAgent} demonstrates superior robustness across both segments, unlike lower-tier models (e.g., \texttt{Claude-haiku-4.5}) which falter significantly in level 2.

\begin{table*}[!t]
    \centering
    \small
    \begin{threeparttable}
    \setlength{\tabcolsep}{4pt} 
    
    \resizebox{1.0\textwidth}{!}{%
    \begin{tabular}{lccccccccc}
    \toprule
   \multirow{4}{*}{\textbf{Model}} & \multicolumn{7}{c}{Capability Categories} & \multicolumn{2}{c}{Data Modality} \\
    \cmidrule(lr){2-8} \cmidrule(lr){9-10}
    &\makecell{\textbf{C1}\\ (175)} &
    \makecell{\textbf{C2}\\ (55)} &
    \makecell{\textbf{C3}\\ (61)} &
    \makecell{\textbf{C4}\\ (65)} &
    \makecell{\textbf{C5}\\ (60)} &
    \makecell{\textbf{C6}\\ (12)} &
    \makecell{\textbf{C7}\\ (24)} &
    \makecell{\textbf{Single}\\\ (184)} &
    \makecell{\textbf{Multi}\\ (63)} \\
    \midrule

    GPT-5.4 & 0.737 (129) & \underline{0.873 (48)} & 0.607 (37) & 0.815 (53) & 0.717 (43) & 0.667 (8) & 0.667 (16) & 0.804 (148) & 0.634 (40)\\
    
    GPT-5.2 & 0.720 (126) & 0.782 (43) & 0.508 (31) & 0.831 (54) & 0.683 (41) & 0.667 (8) & 0.500 (12) & 0.766 (141) & 0.603 (38) \\
    
    GPT-5.1 & 0.674 (118) & 0.782 (43) & 0.508 (31) & 0.754 (49) & 0.633 (38) & 0.750 (9) & 0.625 (15) & 0.739 (136) & 0.587 (37) \\

    GPT-5 & 0.754 (132) & 0.800 (44) & 0.607 (37) & \underline{0.846 (55)} & 0.733 (44) & 0.750 (9) & 0.542 (13) & 0.810 (149) & 0.651 (41) \\

    GPT-4.1 & 0.714 (125) & 0.855 (47) & 0.557 (34) & 0.785 (51) & 0.700 (42) & 0.750 (9) & 0.583 (14) & 0.783 (144) & 0.619 (39) \\

    GPT-o3 & 0.737 (129) & 0.818 (45) & 0.623 (38) & 0.831 (54) & 0.717 (43) & 0.750 (9) & 0.625 (15) & 0.794(146) & 0.683 (43)\\

    GPT-4o & 0.726 (127) & 0.782 (43) & 0.541 (33) & 0.831 (54) & 0.733 (44) & 0.750 (9) & 0.542 (13) & 0.772 (142) & 0.635 (40) \\

    \midrule

    Gemini-3.1-pro & 0.737 (129) & 0.818 (45) & 0.672 (41) & 0.831 (54) & 0.733 (44) & 0.750 (9) & 0.625 (15) & 0.826 (152) & 0.683 (43)\\
    
    Gemini-3.1-flash lite & 0.731 (128) & 0.836 (46) & 0.590 (36) & 0.831 (54) & 0.733 (44) & 0.667 (8) & \underline{0.708 (17)} & 0.799 (147) & 0.619 (39)\\

    Gemini-3-pro & 0.771 (135) & 0.818 (45) & 0.607 (37) & 0.815 (53) & 0.733 (44) & \underline{0.833 (10)} & 0.667 (16) & 0.826 (152) & 0.635 (40) \\
    
    Gemini-3-flash & \underline{0.800 (140)} & 0.836 (46) & \underline{0.705 (43)} & \underline{0.846 (55)} & \underline{0.767 (46)} & 0.750 (9) & 0.667 (16) & \underline{0.842 (155)} & \underline{0.698 (44)} \\

    Gemini-2.5-pro & 0.766 (134) & 0.800 (44) & 0.639 (39) & 0.831 (54) & 0.700 (42) & 0.750 (9) & 0.625 (15) & 0.815 (150) & 0.635 (40) \\
    
    Gemini-2.5-flash & 0.691 (121) & 0.782 (43) & 0.623 (38) & 0.831 (54) & 0.700 (42) & 0.667 (8) & 0.625 (15) & 0.745 (137) & 0.651 (41) \\

    \midrule

    Deepseek-r1 & 0.766 (134) & 0.836 (46) & 0.656 (40) & 0.785 (51) & 0.683 (41) & 0.667 (8) & 0.708 (17) & 0.810 (149) & 0.651 (41) \\
    
    Deepseek-v3 & 0.714 (125) & 0.836 (46) & 0.574 (35) & 0.815 (53) & 0.683 (41) & 0.667 (8) & 0.625 (15) & 0.772 (142) & 0.587 (37) \\

    \midrule

    Claude-opus-4.6 & 0.749 (131) & 0.818 (45) & 0.574 (35) & 0.800 (52) & 0.750 (45) & 0.667 (8) & 0.625 (15) & 0.815 (150) & 0.619 (39) \\

    Claude-sonnet-4.6 & 0.720 (126) & 0.764 (42) & 0.508 (31) & 0.800 (52) & 0.667 (40) & 0.667 (8) & 0.583 (14) & 0.777 (143) & 0.571 (36) \\

    Claude-opus-4.5 & 0.737 (129) & 0.836 (46) & 0.607 (37) & 0.831 (54) & 0.733 (44) & 0.667 (8) & 0.667 (16) & 0.810 (149) & 0.603 (38) \\

    Claude-sonnet-4.5 & 0.691 (121) & 0.800 (44) & 0.443 (27) & 0.785 (51) & 0.617 (37) & 0.583 (7) & 0.625 (15) & 0.772 (142) & 0.476 (30) \\

    Claude-haiku-4.5 & 0.691 (121) & 0.782 (45) & 0.508 (31) & 0.769 (50) & 0.650 (39) & 0.667 (8) & 0.583 (14) & 0.734 (135) & 0.556 (35) \\

    \midrule
    
    ESGAgent & \textbf{0.851 (149)} & \textbf{0.891 (49)} & \textbf{0.770 (47)} & \textbf{0.892 (58)} & 0.733 (44) & 0.750 (9) & \textbf{0.792 (19)} & \textbf{0.870 (160)} & \textbf{0.746 (47)} \\
    
    ESGAgent-w/o-DR & 0.766 (134) & 0.818 (45) & 0.607 (37) & 0.785 (51) & 0.700 (42) & 0.667 (8) & 0.667 (16) & 0.837 (154) & 0.619 (39) \\
    
    ESGAgent-w/o-Retriever & 0.766 (134) & 0.855 (47) & 0.705 (43) & 0.846 (55) & 0.700 (42) & 0.750 (9) & 0.750 (18) & 0.826 (152) & 0.683 (43) \\
    
    \bottomrule
    \end{tabular}}
    \begin{tablenotes}[flushleft]
    \scriptsize
        \item C1: Concept Recognition \& Definition.
        \item C2: Regulation / Policy / Standard Interpretation \& Alignment.
        \item C3: Quantitative / Financial Modeling \& Calculation.
        \item C4: Case Analysis.
        \item C5: Content Summarization \& Interpretation.
        \item C6: Cross-Document Comparison.
        \item C7: Web Research \& Information Retrieval.
        \item Single and Multi denote single-modal (text) and multi-modal (fig, table, audio, video) settings, respectively.
        \item DR: denotes deep research
    \end{tablenotes}
    \end{threeparttable}

\caption{Performance comparison across capability categories and data modalities. 
Values are reported as accuracy, with the number of correctly solved tasks in parentheses.}
\label{tab:capability_results}

\end{table*}

\subsection{Detailed results of Level 3 Tasks}
Table~\ref{tab: level3 detailed results} detail the performance of various systems across the qualitative dimensions of \textit{analysis effectiveness} and \textit{presentation quality} on level 3 tasks. To provide a transparent view of the evaluation process, we report the specific ratings assigned by each individual LLM judge, including \texttt{gpt-5.2}, \texttt{gemini-3-flash}, \texttt{deepseek-r1}, and \texttt{sonnet-4.5}, alongside the aggregated ensemble mean. 

Observing the scoring distributions reveals distinct evaluative behaviors among the judges. Notably, \texttt{gpt-5.2} emerges as the most stringent evaluator, consistently awarding significantly lower scores across all dimensions compared to its peers. In contrast, \texttt{gemini-3-flash} exhibits a more lenient scoring tendency, generally assigning higher ratings. This divergence in judge strictness highlights the inherent variance in LLM-based evaluation and underscores the necessity of our ensemble approach, where the averaged consensus provides a balanced and mitigated view of the agent's true performance.



\begin{table*}[!th]
    \centering
    \small
    \begin{threeparttable}
    \setlength{\tabcolsep}{4pt} 
    
    \newcolumntype{C}{>{\centering\arraybackslash}X}
    
    \begin{tabularx}{\textwidth}{l p{2.2cm} CCC CCC}
    \toprule
    \textbf{Level 3} & \multirow{3}{*}{\textbf{Judge}} & \multicolumn{3}{c}{\textbf{Analysis Effectiveness}} & \multicolumn{3}{c}{\textbf{Presentation Quality}}\\
    \cmidrule(lr){3-5} \cmidrule(lr){6-8}
    (45 questions) && Rich. & Comp. & Depth & Coh. & Prof. & Expr. \\ 
    \midrule
    
    \multirow{5}{*}{GPT-5.2-deep research}  
    & \textit{human}            & 6.333	& 5.667	& 5.833	& 5.667	& 4.429	& 5.000 \\
    & \textit{gpt-5.2}          & 5.333 & 4.833 & 4.833 & 6.167 & 5.833 & 4.500 \\
    & \textit{gemini-3-flash}   & 8.667 & 8.833 & 7.500 & 9.000 & 9.000 & 8.167 \\
    & \textit{deepseek-r1}      & 6.500 & 7.833 & 5.833 & 7.667 & 7.000 & 5.833 \\
    & \textit{sonnet-4.5}       & 7.000 & 6.667 & 5.833 & 7.333 & 7.167 & 4.833 \\
    \rowcolor{gray!15} \cellcolor{white} & \textit{\textbf{Avg.}}    & 6.767 & 6.767 & 5.966 & 7.167 & 6.686 & 5.667 \\
    \midrule
    
    \multirow{5}{*}{Gemini-3-pro-deep research}  
    & \textit{human}            & 6.000	& 5.667	& 5.167	& 5.833	& 5.167	& 3.833 \\
    & \textit{gpt-5.2}          & 6.167 & 5.000 & 5.833 & 7.333 & 6.500 & 4.167 \\
    & \textit{gemini-3-flash}   & 9.000 & 8.833 & 8.833 & 9.167 & 9.333 & 7.500 \\
    & \textit{deepseek-r1}      & 8.000 & 8.500 & 7.833 & 8.333 & 8.667 & 6.500 \\
    & \textit{sonnet-4.5}       & 8.667 & 7.667 & 8.667 & 8.833 & 8.667 & 6.833 \\
    \rowcolor{gray!15} \cellcolor{white} & \textit{\textbf{Avg.}}    & \underline{7.567} & \underline{7.133} & \underline{7.113} & \underline{7.767} & \underline{7.667} & \underline{5.767} \\
    \midrule
    
    \multirow{5}{*}{Grok Deep Research}  
    & \textit{human}            & 5.833	& 5.833	& 4.857	& 5.500	& 5.667	& 3.143 \\
    & \textit{gpt-5.2}          & 6.333 & 5.667 & 5.500 & 7.000 & 6.167 & 5.500 \\
    & \textit{gemini-3-flash}   & 8.500 & 8.667 & 7.500 & 8.667 & 8.833 & 7.500 \\
    & \textit{deepseek-r1}      & 6.333 & 7.667 & 5.833 & 7.000 & 6.667 & 6.000 \\
    & \textit{sonnet-4.5}       & 6.833 & 6.167 & 5.667 & 6.667 & 6.500 & 6.333 \\
    \rowcolor{gray!15} \cellcolor{white} & \textit{\textbf{Avg.}}    & 6.766 & 6.800 & 5.871 & 6.967 & 6.767 & 5.695 \\
    \midrule
    
    \multirow{5}{*}{Perplexity Deep Research}  
    & \textit{human}            & 5.600	& 5.667	& 5.333	& 5.667	& 4.500	& 2.333 \\
    & \textit{gpt-5.2}          & 6.167 & 5.333 & 5.333 & 6.833 & 6.167 & 4.500 \\
    & \textit{gemini-3-flash}   & 8.667 & 8.167 & 8.333 & 8.833 & 9.167 & 7.833 \\
    & \textit{deepseek-r1}      & 6.667 & 7.833 & 6.667 & 7.667 & 6.833 & 6.333 \\
    & \textit{sonnet-4.5}       & 6.833 & 6.667 & 6.500 & 7.000 & 7.167 & 5.667 \\
    \rowcolor{gray!15} \cellcolor{white} & \textit{\textbf{Avg.}}    & 6.787 & 6.733 & 6.433 & 7.200 & 6.767 & 5.333 \\
    \midrule

    \rowcolor{cyan!10} & \textit{human}            & 6.833 & 7.333 & 7.167 & 7.333 & 7.333 & 6.833 \\
    \rowcolor{cyan!10} & \textit{gpt-5.2}          & 6.875 & 5.875 & 6.299 & 7.000 & 6.500 & 6.333 \\
    \rowcolor{cyan!10} & \textit{gemini-3-flash}   & 8.851 & 9.000 & 8.667 & 8.833 & 9.333 & 8.167 \\
    \rowcolor{cyan!10} & \textit{deepseek-r1}      & 8.393 & 7.990 & 7.833 & 8.333 & 8.667 & 7.500 \\
    \rowcolor{cyan!10} & \textit{sonnet-4.5}       & 8.310 & 6.972 & 8.500 & 8.500 & 8.667 & 6.500 \\
    \rowcolor{cyan!10} \multirow{-5}{*}{\textbf{ESGAgent}} & \textit{\textbf{Avg.}}    & \textbf{7.852} & \textbf{7.434} & \textbf{7.693} & \textbf{8.000} & \textbf{8.100} & \textbf{7.067} \\
    
    \bottomrule
    \end{tabularx}
    
    \caption{Detailed comparison of Level 3 benchmark results across models and judges. The evaluation covers six dimensions grouped into analysis effectiveness and presentation quality. Results are reported separately for each judge, with the final row showing the average across judges.}
    
    \label{tab: level3 detailed results}
    
    \end{threeparttable}
\end{table*}

\section{Case Studies}
\label{appendix: case_studies}
In this section, we choose representative cases to illustrate how \texttt{ESGAgent} executes the tasks across different difficulty levels. Specifically, we present the results for two questions from level 1 and two questions from level 2, as well as six tasks from level 3. The selected questions in Levels 1 and 2 cover a diverse set of task types, including conceptual understanding, numerical calculation, and questions that assess knowledge of laws and regulations. For each case, we provide the agent’s execution trace, detailing what tools are invoked at each step, what actions are performed, and what outputs are produced.
\end{multicols}

\subsection{Case Study 1}

\begin{CaseStudyBox}
    \textbf{Question:} Several questions are associated with the following case study. The material given in the case study will not change.

    Daniel Stinner was asked by the head of Research at Lopse Ratings to propose a methodology to rate sovereigns. Lopse Ratings is a well regarded rating agency, but it has been falling behind its peers because, whilst it has integrated ESG within corporate issuers, it has not yet integrated ESG within sovereign issuers.

    After a few months of research in the industry and within Lopse, Daniel proposed the following to the Head of Research:

    E, S and G weights to a final ESG score reflect the extent that the individual factor is a driver from a credit perspective.

    Scores range from factors that individually are adequately managed or contributing to the sovereign's financial capacity (5) to those which may impose a significant strain on financial streams (1). They do not make value judgments on whether a sovereign engages in ``good'' or ``bad'' ESG practices. Instead, they draw out how E, S and G factors are influencing the credit rating decision.

    Political risk, rule of law and corruption have been key drivers of rating actions in the past, indicating that governance was already playing a role in the rating model. It should be made explicit that these are governance-related matters, and thus considered as the ``G'' within ESG. No other governance issue was deemed material across all types of sovereigns. Data could be gathered from the World Bank's Governance Indicators (WBGI) and Transparency International.

    Social factors also have an important influence on sovereign ratings. Certain factors are related to government's accountability, while others impact the longer-term productivity, and thus growth (plus indirectly, taxing capability) of the country. These factors are considered as the ``S'' within ESG. Environmental risks, the ``E'' within ESG, were identified as more idiosyncratic to each country based on their location and dependency.

    The weighted average of the factors within each of the E, S and G pillars provide the score for that pillar, and the weighted average of the pillars provide the final ESG score for the sovereign issuer. 

    Daniel provided examples of the rating system applied to two different sovereigns, as detailed below.

    \begin{center}
    \small
    \begin{tabular}{|l|l|c|c|c|c|}
    \hline
    \textbf{Pillar} & \textbf{Factor} & \textbf{C. A Score} & \textbf{C. A Trend} & \textbf{C. B Score} & \textbf{C. B Trend} \\ \hline
    Social        & Human Rights \& Public Safety & 5               & --               & 2               & --               \\ \hline
    Social        & Demographics                 & 2               & --               & 4               & --               \\ \hline
    Social        & Education \& Employment       & 5               & --               & 3               & --               \\ \hline
    Environmental & Emissions \& Energy           & 3               & --               & 3               & --               \\ \hline
    Environmental & Water                        & 4               & --               & 3               & --               \\ \hline
    Environmental & Biodiversity                 & 3               & --               & 2               & --               \\ \hline
    Environmental & Climate Resilience           & 1               & --               & 4               & --               \\ \hline
    Governance    & Political Stability          & 4               & --               & 3               & --               \\ \hline
    Governance    & Rule of Law \& Corruption     & 4               & --               & 2               & --               \\ \hline
    \end{tabular}
    \end{center}

    What might be a reasonable distribution of weight among each of the E, S and G pillars?

    \begin{itemize}[noitemsep, topsep=0pt]
        \item[A.] E = 33.3\%, S = 33.3\%, G = 33.3\%
        \item[B.] E = 33.3\%, S = 44.4\%, G = 22.2\%
        \item[C.] E = 40\%, S = 40\%, G = 20\%
        \item[D.] E = 20\%, S = 30\%, G = 50\%
    \end{itemize}

    \textbf{Ground Truth:} D

    \textbf{Level:} Level 1

    \textbf{Agent Execution Trace:}
    \vspace{0.2cm}


\vspace{3mm}
\noindent\colorbox{gray!10}{\parbox{\dimexpr\linewidth-2\fboxsep}{%
    \small\bfseries Step 1: \texttt{deep\_researcher}
}}
\vspace{1mm}

{\small
\noindent\textbf{Thinking:} \textit{The task is a Question-Answer (QA) task based on a specific case study about sovereign ESG ratings. I need to determine the most reasonable weight distribution for E, S, and G pillars. The text highlights that Governance (Political risk, rule of law, and corruption) has been a 'key driver' of rating actions in the past, suggesting it should carry the highest weight. This is consistent with industry standards for sovereign credit analysis where Governance is typically the most material pillar. I will use the deep\_researcher to verify if this is a known case study (likely from the CFA ESG Certificate) to confirm the correct answer choice.}

\vspace{2mm}
\noindent\textbf{Tool Call (\texttt{deep\_researcher}):}
\begin{lstlisting}[language=json, basicstyle=\tiny\ttfamily, breaklines=true, backgroundcolor=\color{gray!10}, frame=single, rulecolor=\color{gray!30}, numbers=none]
{
  "task": "Find the case study of 'Daniel Stinner' and 'Lopse Ratings' regarding sovereign ESG ratings and the recommended distribution of weights among E, S, and G pillars.",
  "image": null,
  "filter_year": null,
  "title": "Lopse Ratings Sovereign ESG Case Study Research",
  "call_id": "lopse_research_01"
}
\end{lstlisting}

\vspace{1mm}
\noindent\textbf{Output:}
\begin{lstlisting}[basicstyle=\tiny\ttfamily, breaklines=true, backgroundcolor=\color{white}, frame=single, rulecolor=\color{gray!20}, numbers=none]
Deep research summary: Answer Found: Yes. The research successfully identifies 'Lopse Ratings' as a typographical error for 'Scope Ratings' and clarifies that 'Daniel Stinner' likely refers to W. Stinner, whose research on environmental resource capacity informs sovereign ESG metrics. The case study reveals that Scope Ratings integrates ESG factors as 25% of its Sovereign Quantitative Model. Within this ESG category, the recommended weight distribution is 50% for Governance (G), 30% for Social (S), and 20% for Environmental (E). Governance is weighted most heavily because it serves as a leading indicator for a government's ability to manage long-term structural risks. The research, conducted in a single comprehensive phase, further notes that these weights align with industry standards where Governance is viewed as the primary driver of sovereign creditworthiness, compared to peers like MSCI (50/25/25) and Robeco (40/30/30).

Report saved to: research_lopse_research_01.md
\end{lstlisting}

\vspace{1mm}
\noindent\textbf{Generated Artifact (\texttt{research\_lopse\_research\_01.md}):}
\begin{lstlisting}[basicstyle=\tiny\ttfamily, breaklines=true, backgroundcolor=\color{blue!5}, frame=single, rulecolor=\color{blue!20}, numbers=none]
# Lopse Ratings Sovereign ESG Case Study Research

## Introduction

This research task aims to identify a specific case study regarding sovereign Environmental, Social, and Governance (ESG) ratings, specifically investigating the recommended distribution of weights across the three pillars. While the initial inquiry referenced "Lopse Ratings" and "Daniel Stinner," a comprehensive review of sovereign ESG rating methodologies suggests a naming discrepancy. The research refers to **Scope Ratings** (misspelled as "Lopse") and its sovereign analysts. Furthermore, while "Daniel Stinner" is not a primary sovereign analyst at Scope, research by **W. Stinner** regarding environmental resource capacity informs the metrics used by agencies to evaluate sovereign biocapacity and resource risks.

## Scope Ratings Sovereign ESG Methodology

Scope Ratings incorporates ESG factors into its sovereign credit ratings as a core component of its analytical framework. Unlike providers that offer standalone ESG scores, Scope integrates these risks directly into its credit assessment through the "Sovereign Quantitative Model" (SQM) and the "Qualitative Scorecard" (QS) [1](https://www.scoperatings.com/ScopeRatingsApi/api/downloadmethodology?id=01508950-119c-4ab5-9182-54fffdc1003f).

In the SQM, which provides the initial indicative rating, the ESG Risk category is assigned a total weight of **25%**. The remaining weight is distributed among Domestic Economic Risk (35%), Public Finance Risk (20%), External Economic Risk (10%), and Financial Stability Risk (10%) [1](https://www.scoperatings.com/ScopeRatingsApi/api/downloadmethodology?id=01508950-119c-4ab5-9182-54fffdc1003f).

### Recommended Pillar Distribution

Within the 25% total weight assigned to the ESG pillar, Scope Ratings distributes weights to reflect the materiality of long-term structural challenges. The recommended distribution is as follows:

*   **Governance (G): 50.0% of the ESG category.** Governance is the most heavily weighted pillar because it is considered a leading indicator of sovereign creditworthiness and a government's ability to manage environmental and social challenges. Key variables include the World Bank Governance Indicators, such as the Rule of Law and Control of Corruption [1](https://www.scoperatings.com/ScopeRatingsApi/api/downloadmethodology?id=01508950-119c-4ab5-9182-54fffdc1003f).
*   **Social (S): 30.0% of the ESG category.** This pillar assesses long-term structural risks including the old-age dependency ratio, income inequality (Gini coefficient), and labor force participation [1](https://www.scoperatings.com/ScopeRatingsApi/api/downloadmethodology?id=01508950-119c-4ab5-9182-54fffdc1003f).
*   **Environmental (E): 20.0% of the ESG category.** This focuses on transition risks (CO2 and GHG emissions), natural disaster risks, and biocapacity [1](https://www.scoperatings.com/ScopeRatingsApi/api/downloadmethodology?id=01508950-119c-4ab5-9182-54fffdc1003f).

## The Role of Environmental Research

The "Environmental" data points used by agencies like Scope and MSCI to evaluate a sovereign's resource risks and biocapacity are informed by broader academic research into sustainability. Specifically, research by **Stinner et al. (2008)** explores the nitrogen cycle and crop yields, contributing to the understanding of resource capacity and environmental impacts that underpin sovereign ESG metrics [5](https://orgprints.org/33067/1/A1652615.pdf). This research provides a scientific basis for evaluating the long-term sustainability of a nation's natural capital, which eventually feeds into the 20% environmental weighting in the Scope methodology.

## Comparison with Industry Norms

Scope Ratings' distribution (50/30/20 for G/S/E) aligns with industry standards that emphasize Governance as the primary driver of sovereign risk.

*   **MSCI ESG Government Ratings:** Assigns a **50% weight to Governance**, with Environmental and Social pillars receiving **25% each** [2](https://msci.com/downloads/web/msci-com/data-and-analytics/sustainability-solutions/MSCI%20ESG%20Government%20Ratings%20Methodology.pdf). MSCI justifies this by noting that lapses in governance have more immediate consequences on financial management and institutional effectiveness [4](https://www.lseg.com/content/dam/ftse-russell/en_us/documents/policy-documents/sustainable-sovereign-risk-methodology.pdf).
*   **LSEG (FTSE Russell):** Utilizes an econometric framework that highlights an "Income Bias," where high-income countries naturally score better in S and G. This has led to the development of "income-adjusted scores" to identify countries performing well relative to their development stage [4](https://www.lseg.com/content/dam/ftse-russell/en_us/documents/policy-documents/sustainable-sovereign-risk-methodology.pdf).
*   **PIMCO:** Weights ESG variables at approximately **25%** of their total proprietary model, with Governance variables dominating as the leading indicator of a country's "willingness" to pay [5](https://orgprints.org/33067/1/A1652615.pdf).
*   **Robeco:** Utilizes a **40% Governance, 30% Social, and 30% Environmental** split [6](https://www.robeco.com/files/docm/docu-2025-05-robeco-country-esg-report.pdf).

## Key Insights into Sovereign ESG Materiality

The integration of ESG factors into sovereign credit analysis reveals several critical findings regarding risk materiality:

1.  **Governance as the "Anchor":** Across all major methodologies (Scope, PIMCO, and MSCI), Governance is the most critical factor. It acts as a prerequisite for the effective management of natural and human capital and is the strongest indicator of debt service ability [3](https://www.unpri.org/sovereign-debt/integrating-esg-factors-into-sovereign-credit-research/6511.article) [4](https://www.lseg.com/content/dam/ftse-russell/en_us/documents/policy-documents/sustainable-sovereign-risk-methodology.pdf).
2.  **Environmental Lag and the "Tragedy of the Horizon":** While Environmental factors are increasingly vital for long-term sustainability, they often have the lowest correlation with 3-to-5-year sovereign credit risk, except in cases of acute natural disasters. Credit rating agencies (CRAs) typically focus on a 5-to-10-year horizon, while many climate risks materialize over decades [5](https://orgprints.org/33067/1/A1652615.pdf) [7](https://documents1.worldbank.org/curated/en/812471642603970256/pdf/Credit-Worthy-ESG-Factors-and-Sovereign-Credit-Ratings.pdf).
3.  **Qualitative Overlays:** Scope Ratings employs a Qualitative Scorecard (QS) to adjust their quantitative model by up to +/- 3 notches. This allows analysts to account for "extraordinary circumstances" or political developments that quantitative data might lag in capturing [1](https://www.scoperatings.com/ScopeRatingsApi/api/downloadmethodology?id=01508950-119c-4ab5-9182-54fffdc1003f).

## References




[1](https://www.scoperatings.com/ScopeRatingsApi/api/downloadmethodology?id=01508950-119c-4ab5-9182-54fffdc1003f) Scope Ratings Sovereign Rating Methodology



[2](https://msci.com/downloads/web/msci-com/data-and-analytics/sustainability-solutions/MSCI%20ESG%20Government%20Ratings%20Methodology.pdf) MSCI ESG Government Ratings Methodology



[3](https://www.unpri.org/sovereign-debt/integrating-esg-factors-into-sovereign-credit-research/6511.article) UNPRI: Integrating ESG factors into sovereign credit research



[4](https://www.lseg.com/content/dam/ftse-russell/en_us/documents/policy-documents/sustainable-sovereign-risk-methodology.pdf) LSEG (FTSE Russell) Sustainable Sovereign Risk Methodology



[5](https://orgprints.org/33067/1/A1652615.pdf) Stinner et al. (2008) Research on Nitrogen Cycle and Crop Yields



[6](https://www.robeco.com/files/docm/docu-2025-05-robeco-country-esg-report.pdf) Robeco Country ESG Report



[7](https://documents1.worldbank.org/curated/en/812471642603970256/pdf/Credit-Worthy-ESG-Factors-and-Sovereign-Credit-Ratings.pdf) World Bank Report on ESG Factors and Sovereign Credit Ratings




\end{lstlisting}
}
\vspace{3mm}

\vspace{3mm}
\noindent\colorbox{gray!10}{\parbox{\dimexpr\linewidth-2\fboxsep}{%
    \small\bfseries Step 2: \texttt{reformulator}
}}
\vspace{1mm}

{\small
\noindent\textbf{Thinking:} \textit{The task is a Question-Answer (QA) task requiring the identification of the most reasonable weight distribution for ESG pillars in a sovereign rating methodology. Based on the case study provided, Governance (G) factors like political risk and rule of law have been 'key drivers of rating actions in the past,' Social (S) factors have an 'important influence' on productivity and growth, and Environmental (E) risks are 'more idiosyncratic' to specific locations. This qualitative hierarchy (G > S > E) strongly suggests that Governance should have the highest weight, followed by Social, and then Environmental. The deep research conducted in the previous step confirmed that for Scope Ratings (the likely real-world basis for 'Lopse Ratings'), the weight distribution is E=20\%, S=30\%, and G=50\%. This aligns perfectly with Option D. I have sufficient information to provide the final answer.}

\vspace{2mm}
\noindent\textbf{Tool Call (\texttt{reformulator}):}
\begin{lstlisting}[language=json, basicstyle=\tiny\ttfamily, breaklines=true, backgroundcolor=\color{gray!10}, frame=single, rulecolor=\color{gray!30}, numbers=none]
{
  "task": "What might be a reasonable distribution of weight among each of the E, S and G pillars?\n\nA. E = 33.3%, S= 33.3%, G= 33.3%\nB. E = 33.3%, S= 44.4%, G= 22.2%\nC. E = 40%, S= 40%, G= 20%\nD. E = 20%, S= 30%, G= 50%",
  "data": [
    "The case study specifies that Governance factors (political risk, rule of law, corruption) have been key drivers of rating actions in the past. Social factors have an important influence on productivity and growth. Environmental risks are identified as idiosyncratic to each country. This indicates a hierarchy of importance where G > S > E.",
    "Deep research into 'Lopse Ratings' (Scope Ratings) and the Daniel Stinner case study confirms that the methodology assigns weights of 50% to Governance, 30% to Social, and 20% to Environmental factors.",
    "Comparing this to the options: Option D (E=20%, S=30%, G=50%) matches the research and the qualitative descriptions in the case study."
  ]
}
\end{lstlisting}

\vspace{1mm}
\noindent\textbf{Output:}
\begin{lstlisting}[basicstyle=\tiny\ttfamily, breaklines=true, backgroundcolor=\color{white}, frame=single, rulecolor=\color{gray!20}, numbers=none]
D
\end{lstlisting}
}
\vspace{3mm}

\vspace{3mm}
\noindent\colorbox{gray!10}{\parbox{\dimexpr\linewidth-2\fboxsep}{%
    \small\bfseries Step 3: \texttt{done}
}}
\vspace{1mm}

{\small
\noindent\textbf{Thinking:} \textit{The task is a Question-Answer (QA) task requiring the identification of a reasonable weight distribution for ESG pillars in a sovereign rating methodology based on a provided case study. The case study identifies Governance (G) as a primary historical driver, Social (S) as an influence on growth/productivity, and Environmental (E) as idiosyncratic. Research confirms that the methodology (based on Scope Ratings) assigns weights of 50\% for G, 30\% for S, and 20\% for E. This corresponds to option D. I have already used the reformulator tool to finalize the answer as 'D'.}

\vspace{2mm}
\noindent\textbf{Tool Call (\texttt{done}):}
\begin{lstlisting}[language=json, basicstyle=\tiny\ttfamily, breaklines=true, backgroundcolor=\color{gray!10}, frame=single, rulecolor=\color{gray!30}, numbers=none]
{
  "result": "D",
  "reasoning": "The methodology described in the case study emphasizes that Governance (G) factors have historically been the key drivers of rating actions, followed by Social (S) factors impacting productivity and growth, while Environmental (E) factors are treated as idiosyncratic. This hierarchy (G > S > E) is reflected in the weight distribution of 50% for Governance, 30% for Social, and 20% for Environmental, which aligns with the methodology used by Scope Ratings (the real-world basis for Lopse Ratings). This distribution matches Option D."
}
\end{lstlisting}

\vspace{1mm}
\noindent\textbf{Output:}
\begin{lstlisting}[basicstyle=\tiny\ttfamily, breaklines=true, backgroundcolor=\color{white}, frame=single, rulecolor=\color{gray!20}, numbers=none]
D
\end{lstlisting}
}
\vspace{3mm}

\end{CaseStudyBox}

\subsection{Case Study 2}

\begin{CaseStudyBox}
    \textbf{Question:} Jane is an equity analyst at AVX Asset Management and covers a large public oil company which she needs to assign an environmental score to. The company says it is addressing carbon emissions, but refuses to disclose exactly what measures it is taking. A logical next step for Jane would be to:

    \begin{itemize}[noitemsep, topsep=0pt]
        \item[A.] Look at the company's financial statements and try to draw conclusions.
        \item[B.] Recommend that AVX sell the company's shares.
        \item[C.] Suggest that AVX join the Investor Forum and use collective action to encourage the company to improve its disclosure.
        \item[D.] Take the company's word for it and assign a high environmental score.
    \end{itemize}

    \textbf{Ground Truth:} C

    \textbf{Level:} Level 1

    \textbf{Agent Execution Trace:}


\vspace{3mm}
\noindent\colorbox{gray!10}{\parbox{\dimexpr\linewidth-2\fboxsep}{%
    \small\bfseries Step 1: \texttt{deep\_researcher}
}}
\vspace{1mm}

{\small
\noindent\textbf{Thinking:} \textit{This task is a Question-Answer (QA) task in a multiple-choice format. The scenario involves an ESG analyst facing a lack of disclosure from a company regarding its carbon emission measures. I need to identify the most logical professional next step for an analyst in this situation. Standard ESG practice suggests that when individual engagement or data is insufficient, collective engagement through investor groups is a preferred escalation path. I will use the deep\_researcher tool to verify the role of the 'Investor Forum' and confirm if this scenario aligns with standard ESG engagement frameworks (such as those taught in CFA ESG Investing or PRI guidelines).}

\vspace{2mm}
\noindent\textbf{Tool Call (\texttt{deep\_researcher}):}
\begin{lstlisting}[language=json, basicstyle=\tiny\ttfamily, breaklines=true, backgroundcolor=\color{gray!10}, frame=single, rulecolor=\color{gray!30}, numbers=none]
{
  "task": "What is the 'Investor Forum' in the context of ESG and shareholding? Is joining it for collective action a standard recommendation for analysts when a company refuses to disclose ESG measures like carbon emission strategies?",
  "image": null,
  "filter_year": 2025,
  "title": "Investor Forum and ESG Disclosure Engagement",
  "call_id": "investor_forum_research_01"
}
\end{lstlisting}

\vspace{1mm}
\noindent\textbf{Output:}
\begin{lstlisting}[basicstyle=\tiny\ttfamily, breaklines=true, backgroundcolor=\color{white}, frame=single, rulecolor=\color{gray!20}, numbers=none]
Deep research summary: Answer Found: Yes. The Investor Forum is a UK-based, practitioner-led organization that facilitates collective engagement between institutional investors and publicly listed companies to promote long-term stewardship and value. It serves as a "trusted facilitator" and "safe harbor," allowing shareholders to coordinate on ESG issues while remaining compliant with legal and regulatory frameworks. Joining the forum for collective action is a standard professional recommendation and a recognized "best practice" escalation step when companies refuse to disclose critical ESG metrics like carbon emission strategies. This approach is supported by global stewardship frameworks, including the Principles for Responsible Investment (PRI) and Climate Action 100+. Key benefits of this collective approach include amplified influence over corporate boards, resource efficiency, and mitigation of legal risks associated with "acting in concert." The research was completed in a single round.

Report saved to: research_investor_forum_research_01.md
\end{lstlisting}

\vspace{1mm}
\noindent\textbf{Generated Artifact (\texttt{research\_investor\_forum\_research\_01.md}):}
\begin{lstlisting}[basicstyle=\tiny\ttfamily, breaklines=true, backgroundcolor=\color{blue!5}, frame=single, rulecolor=\color{blue!20}, numbers=none]
# Investor Forum and ESG Disclosure Engagement

## Introduction

This report explores the role of the 'Investor Forum' within the context of Environmental, Social, and Governance (ESG) criteria and shareholding. It specifically examines whether collective action through such a forum is a standard recommendation for investment analysts when a company refuses to disclose critical ESG metrics, such as carbon emission strategies.

The Investor Forum is a prominent practitioner-led organization based in the UK designed to facilitate collective engagement between institutional investors and publicly listed companies. In the context of ESG and shareholding, it serves as a critical mechanism for shareholders to escalate concerns when individual (bilateral) engagement fails to yield results. Joining such a forum is considered a professional best practice for escalating engagement when companies are unresponsive to bilateral requests, particularly regarding climate disclosure. This collaborative approach provides investors with amplified influence, resource efficiency, and a robust legal framework for stewardship.

## The Investor Forum: Definition and Framework

The Investor Forum was established to help institutional investors realize long-term value through effective stewardship. It acts as a "trusted facilitator", providing a "safe harbor" for investors to discuss and coordinate their views on specific companies without triggering regulatory issues like "acting in concert" or market abuse rules [1](https://www.investorforum.org.uk/collective-engagement-framework/).

### Key Features of its Framework
*   **Collective Engagement:** It brings together a "critical mass" of shareholders to amplify their message to a company's board [2](https://www.investorforum.org.uk/what-we-do/).
*   **Focus on Economic Rationale:** Engagements must be grounded in long-term value creation rather than "box-ticking" [2](https://www.investorforum.org.uk/what-we-do/).
*   **Confidentiality and Discretion:** The Forum prioritizes private dialogue with boards to avoid unnecessary public confrontation, though it may consider public escalation if necessary [2](https://www.investorforum.org.uk/what-we-do/).
*   **Legal Compliance:** It operates under a specific "Collective Engagement Framework" developed with leading law firms to manage risks related to inside information and takeover codes [1](https://www.investorforum.org.uk/collective-engagement-framework/).

## Collective Action as a Standard Recommendation

In the professional investment community, joining a collective engagement body like the Investor Forum is a **widely recognized "best practice" escalation step**, particularly when a company is unresponsive to ESG-related requests.

### The Escalation Pathway
Analysts and stewardship teams typically follow a tiered approach to engagement as outlined by the Principles for Responsible Investment (PRI) and the Investor Agenda [3](https://www.unpri.org/listed-equity/a-practical-guide-to-active-ownership-in-listed-equity/2717.article):

1.  **Bilateral Engagement:** Direct, private meetings between the analyst/fund manager and the company.
2.  **Collaborative Engagement:** If the company refuses to disclose critical data (like carbon emission strategies), analysts are encouraged to join coalitions. The Investor Forum is the primary vehicle for this in the UK market [2](https://www.investorforum.org.uk/what-we-do/).
3.  **Public Escalation:** If collective dialogue fails, standard recommendations include filing shareholder resolutions, issuing public statements, or voting against directors at the AGM [3](https://www.unpri.org/listed-equity/a-practical-guide-to-active-ownership-in-listed-equity/2717.article).

### Recommendations for Carbon Disclosure
For specific issues like a refusal to disclose carbon strategies, joining collective platforms is highly recommended by global frameworks:
*   **Climate Action 100+:** A global initiative that uses collective power to ensure the worlds largest emitters take action on climate change. It often works in tandem with regional groups like the Investor Forum [4](https://www.climateaction100.org/approach/engagement-process-2/).
*   **Investor Agenda ICAPs:** Their "Expectations Ladder" specifically lists "Collective/Collaborative engagement" as a core pillar for investors to demonstrate climate leadership [5](https://theinvestoragenda.org/wp-content/uploads/2021/05/guidance.pdf).

## Strategic Benefits of the Collective Approach

Analysts recommend the Investor Forum and similar bodies because they offer several strategic advantages:

*   **Amplified Influence:** A company is more likely to respond to a group representing 2030% of its share register than to a single minority shareholder [2](https://www.investorforum.org.uk/what-we-do/).
*   **Resource Efficiency:** Collective action allows smaller firms to benefit from the research and "convening power" of larger lead investors [4](https://www.climateaction100.org/approach/engagement-process-2/).
*   **Risk Mitigation:** The Forums legal framework reduces the risk that an analyst's collaborative efforts will be misinterpreted by regulators as an attempt to take control of the company [1](https://www.investorforum.org.uk/collective-engagement-framework/).

## Summary of Engagement Options

The following table outlines the standard stages of investor engagement and the recommended platforms for each:

| Stage | Action | Recommended Platform |
| :--- | :--- | :--- |
| **Initial** | Direct letter or meeting | Internal Stewardship Team |
| **Escalation** | **Collaborative Dialogue** | **The Investor Forum** / Climate Action 100+ |
| **Advanced** | Shareholder Resolution | PRI Collaboration Platform / Institutional Investors Group on Climate Change (IIGCC) |
| **Final** | Voting against Board / Divestment | Annual General Meeting (AGM) |

## Conclusion

While not a mandatory requirement, joining the Investor Forum is a **standard professional recommendation** for analysts facing a "refusal to engage" scenario. It provides a structured, legally safe, and high-impact method to force disclosure on material ESG risks like climate strategy [2](https://www.investorforum.org.uk/what-we-do/) [6](https://www.unpri.org/annual-report-2019/delivering-our-blueprint-for-responsible-investment/responsible-investors/foster-a-community-of-active-owners).

## References




[1](https://www.investorforum.org.uk/collective-engagement-framework/) Investor Forum - Collective Engagement Framework



[2](https://www.investorforum.org.uk/what-we-do/) Investor Forum - What We Do



[3](https://www.unpri.org/listed-equity/a-practical-guide-to-active-ownership-in-listed-equity/2717.article) PRI - A Practical Guide to Active Ownership in Listed Equity



[4](https://www.climateaction100.org/approach/engagement-process-2/) Climate Action 100+ - Engagement Process



[5](https://theinvestoragenda.org/wp-content/uploads/2021/05/guidance.pdf) The Investor Agenda - ICAPs Guidance



[6](https://www.unpri.org/annual-report-2019/delivering-our-blueprint-for-responsible-investment/responsible-investors/foster-a-community-of-active-owners) PRI - Annual Report 2019: Foster a Community of Active Owners




\end{lstlisting}
}
\vspace{3mm}

\vspace{3mm}
\noindent\colorbox{gray!10}{\parbox{\dimexpr\linewidth-2\fboxsep}{%
    \small\bfseries Step 2: \texttt{reformulator}
}}
\vspace{1mm}

{\small
\noindent\textbf{Thinking:} \textit{This is a Question-Answer (QA) task. Based on the previous research conducted using the `deep\_researcher` tool, I have confirmed that the 'Investor Forum' is a recognized platform for collective engagement in the UK, and joining such a forum for collective action is considered a standard 'best practice' escalation step when a company refuses to disclose critical ESG data like carbon emission strategies. 
Evaluating the options:
- Option A (Financial statements) is unlikely to provide the specific operational details the company is withholding.
- Option B (Selling shares) is typically a last resort after engagement fails.
- Option C (Collective action via Investor Forum) is the professionally recognized next step for an institutional investor to improve corporate disclosure.
- Option D (Taking the company's word) contradicts the duty of an analyst to base scores on verifiable data.
Therefore, Option C is the correct answer. I will now use the `reformulator` tool to finalize the answer.}

\vspace{2mm}
\noindent\textbf{Tool Call (\texttt{reformulator}):}
\begin{lstlisting}[language=json, basicstyle=\tiny\ttfamily, breaklines=true, backgroundcolor=\color{gray!10}, frame=single, rulecolor=\color{gray!30}, numbers=none]
{
  "task": "Jane is an equity analyst at AVX Asset Management and cover s a large public oil company which she needs to assign an environmental score to. The company says it is addressing carbon emissions, but refuses to disclose exactly what measures it is taking. Alogical next step for Jane would be to\nA. Look at the company's financial statements and try to draw conclusions.\nB. Recommend that AVX sell the company's shares.\nC. Suggest that AVX join the Investor Forum and use collective action to encourage the company to improve its disclosure.\nD. Take the company's word for it and assign a high environmen tal score.",
  "data": [
    "The research confirmed that the Investor Forum is a UK-based organization for collective engagement. Joining it for collective action is a standard professional recommendation and best practice escalation step when companies refuse to disclose critical ESG metrics like carbon emission strategies. This approach is supported by global stewardship frameworks like PRI and Climate Action 100+.",
    "Option C is the most logical next step because institutional investors prefer engagement and collective action to influence corporate disclosure before resorting to divestment (Option B) or relying on unverified claims (Option D)."
  ]
}
\end{lstlisting}

\vspace{1mm}
\noindent\textbf{Output:}
\begin{lstlisting}[basicstyle=\tiny\ttfamily, breaklines=true, backgroundcolor=\color{white}, frame=single, rulecolor=\color{gray!20}, numbers=none]
C
\end{lstlisting}
}
\vspace{3mm}

\vspace{3mm}
\noindent\colorbox{gray!10}{\parbox{\dimexpr\linewidth-2\fboxsep}{%
    \small\bfseries Step 3: \texttt{done}
}}
\vspace{1mm}

{\small
\noindent\textbf{Thinking:} \textit{The task is a multiple-choice Question-Answer (QA) task. Based on the previous research conducted in Step 1, the 'Investor Forum' is a standard organization used by institutional investors for collective engagement, especially when companies fail to provide adequate ESG disclosures. This aligns with standard stewardship practices (e.g., PRI, Stewardship Code) where engagement and collective action are preferred escalation steps before divestment. Option C is the most professional and logical next step for an analyst facing non-disclosure. The reformulator tool has already been used to confirm the answer as 'C'.}

\vspace{2mm}
\noindent\textbf{Tool Call (\texttt{done}):}
\begin{lstlisting}[language=json, basicstyle=\tiny\ttfamily, breaklines=true, backgroundcolor=\color{gray!10}, frame=single, rulecolor=\color{gray!30}, numbers=none]
{
  "result": "C",
  "reasoning": "The Investor Forum is a recognized platform for collective engagement among institutional investors. When a company refuses to disclose specific ESG measures like carbon emission strategies, standard stewardship frameworks (such as the UK Stewardship Code and PRI) recommend escalating engagement through collective action. Option C represents this best-practice approach to encourage better corporate disclosure. Options A, B, and D are less appropriate because financial statements often lack granular ESG data, divestment is usually a last resort, and assigning a high score without evidence violates analyst due diligence."
}
\end{lstlisting}

\vspace{1mm}
\noindent\textbf{Output:}
\begin{lstlisting}[basicstyle=\tiny\ttfamily, breaklines=true, backgroundcolor=\color{white}, frame=single, rulecolor=\color{gray!20}, numbers=none]
C
\end{lstlisting}
}
\vspace{3mm}

\end{CaseStudyBox}

\subsection{Case Study 3}

\begin{CaseStudyBox}
    \textbf{Question:} An asset manager wants to calculate the Weighted Average Carbon Intensity (WACI) of a portfolio using the TCFD formula:

    \begin{center}
    \small
    \begin{tabular}{|l|c|c|c|}
    \hline
    \textbf{Company} & \textbf{Investment Value (USD, \$m)} & \textbf{Company Revenue (USD, \$m)} & \textbf{Scope 1\&2 Emissions (tCO$_2$e)} \\ \hline
    A       & 40                         & 200                       & 50,000                      \\ \hline
    B       & 30                         & 150                       & 60,000                      \\ \hline
    C       & 30                         & 300                       & 30,000                      \\ \hline
    \end{tabular}
    \end{center}

    \textit{Total Portfolio Value = 40 + 30 + 30 = 100 (USD, \$m)}

    Calculate the carbon intensity of company A (Emissions / Revenue) \rule{1.5cm}{0.4pt} tCO$_2$e / \$m. \textit{(Round your results to the nearest integer.)}

    \textbf{Ground Truth:} 250
    
    \textbf{Level:} Level 2

    \textbf{Agent Execution Trace:}


\vspace{3mm}
\noindent\colorbox{gray!10}{\parbox{\dimexpr\linewidth-2\fboxsep}{%
    \small\bfseries Step 1: \texttt{python\_interpreter}
}}
\vspace{1mm}

{\small
\noindent\textbf{Thinking:} \textit{The task is a Question-Answer (QA) task requiring a specific calculation based on a provided table. I need to calculate the carbon intensity of Company A, which is defined as (Scope 1\&2 Emissions / Company Revenue). From the table, Company A has 50,000 tCO2e in emissions and \$200m in revenue. I will perform the calculation (50,000 / 200) and round to the nearest integer as requested. Since this is a QA task, I will use the python\_interpreter to ensure accuracy and then the reformulator to provide the final answer.}

\vspace{2mm}
\noindent\textbf{Tool Call (\texttt{python\_interpreter}):}
\begin{lstlisting}[language=python, basicstyle=\tiny\ttfamily, breaklines=true, backgroundcolor=\color{gray!10}, frame=single, rulecolor=\color{gray!30}, numbers=none]
emissions_a = 50000
revenue_a = 200
intensity_a = emissions_a / revenue_a
print(f'Carbon Intensity of Company A: {intensity_a}')
\end{lstlisting}

\vspace{1mm}
\noindent\textbf{Output:}
\begin{lstlisting}[basicstyle=\tiny\ttfamily, breaklines=true, backgroundcolor=\color{white}, frame=single, rulecolor=\color{gray!20}, numbers=none]
Stdout:
Carbon Intensity of Company A: 250.0

Output: None
\end{lstlisting}
}
\vspace{3mm}

\vspace{3mm}
\noindent\colorbox{gray!10}{\parbox{\dimexpr\linewidth-2\fboxsep}{%
    \small\bfseries Step 2: \texttt{reformulator}
}}
\vspace{1mm}

{\small
\noindent\textbf{Thinking:} \textit{The task is a Question-Answer (QA) task requiring the calculation of carbon intensity for Company A. From the provided data, Company A has 50,000 tCO2e of Scope 1\&2 emissions and \$200m in revenue. The formula for carbon intensity is Emissions / Revenue. In the previous step, the calculation was performed using Python: 50,000 / 200 = 250. The task requires rounding to the nearest integer; 250 is already an integer. I will now use the reformulator tool to format the final answer as requested.}

\vspace{2mm}
\noindent\textbf{Tool Call (\texttt{reformulator}):}
\begin{lstlisting}[language=json, basicstyle=\tiny\ttfamily, breaklines=true, backgroundcolor=\color{gray!10}, frame=single, rulecolor=\color{gray!30}, numbers=none]
{
  "task": "Calculate the carbon intensity of company A (Emissions / Revenue) __________ tCO_2e / $m (Round your results to the nearest integer.)",
  "data": [
    "Task: Calculate the carbon intensity of company A (Emissions / Revenue) using the provided table.",
    "Company A data: Emissions = 50,000 tCO2e, Revenue = $200m.",
    "Python Output: Carbon Intensity of Company A: 250.0"
  ]
}
\end{lstlisting}

\vspace{1mm}
\noindent\textbf{Output:}
\begin{lstlisting}[basicstyle=\tiny\ttfamily, breaklines=true, backgroundcolor=\color{white}, frame=single, rulecolor=\color{gray!20}, numbers=none]
250
\end{lstlisting}
}
\vspace{3mm}

\vspace{3mm}
\noindent\colorbox{gray!10}{\parbox{\dimexpr\linewidth-2\fboxsep}{%
    \small\bfseries Step 3: \texttt{done}
}}
\vspace{1mm}

{\small
\noindent\textbf{Thinking:} \textit{The task is a Question-Answer (QA) task requiring the calculation of carbon intensity for Company A. Carbon Intensity is defined as Emissions divided by Revenue. From the provided table, Company A has 50,000 tCO2e in Scope 1\&2 Emissions and \$200m in Revenue. The calculation is 50,000 / 200 = 250. The task also specifies to round the result to the nearest integer, which remains 250. The history shows the calculation was performed and the reformulator was already used to extract the final value.}

\vspace{2mm}
\noindent\textbf{Tool Call (\texttt{done}):}
\begin{lstlisting}[language=json, basicstyle=\tiny\ttfamily, breaklines=true, backgroundcolor=\color{gray!10}, frame=single, rulecolor=\color{gray!30}, numbers=none]
{
  "result": "250",
  "reasoning": "The carbon intensity of Company A is calculated by dividing its Scope 1&2 Emissions (50,000 tCO2e) by its Revenue ($200m), which equals 250 tCO2e / $m. This value is already an integer, so no further rounding is required."
}
\end{lstlisting}

\vspace{1mm}
\noindent\textbf{Output:}
\begin{lstlisting}[basicstyle=\tiny\ttfamily, breaklines=true, backgroundcolor=\color{white}, frame=single, rulecolor=\color{gray!20}, numbers=none]
250
\end{lstlisting}
}
\vspace{3mm}

\end{CaseStudyBox}

\subsection{Case Study 4}

\begin{CaseStudyBox}
    \textbf{Question:} According to the GRI 101 Biodiversity 2024 standard, which international framework’s 2050 vision and 2030 targets are organizations required to reference in their policies or commitments under Disclosure 101-1?

    \textbf{Ground Truth:} Kunming-Montreal Global Biodiversity Framework

    \textbf{Level:} Level 2

    \textbf{Agent Execution Trace:}


\vspace{3mm}
\noindent\colorbox{gray!10}{\parbox{\dimexpr\linewidth-2\fboxsep}{%
    \small\bfseries Step 1: \texttt{todo, retriever}
}}
\vspace{1mm}

{\small
\noindent\textbf{Thinking:} \textit{This is a Question-Answer task. I need to identify a specific international framework mentioned in the GRI 101 Biodiversity 2024 standard under Disclosure 101-1. I will start by using the `retriever` to search for this information in the local ESG knowledge base. If not found, I will use `deep\_researcher` to search the web.}

\vspace{2mm}
\noindent\textbf{Tool Call 1 (\texttt{todo}):}
\begin{lstlisting}[language=json, basicstyle=\tiny\ttfamily, breaklines=true, backgroundcolor=\color{gray!10}, frame=single, rulecolor=\color{gray!30}, numbers=none]
{
  "action": "add",
  "task": "Search local knowledge base for GRI 101 Biodiversity 2024 Disclosure 101-1 requirements.",
  "step_id": "search_local",
  "status": "pending",
  "result": null,
  "priority": "high",
  "category": "research",
  "parameters": null,
  "after_step_id": null,
  "export_path": null
}
\end{lstlisting}

\vspace{1mm}
\noindent\textbf{Output 1:}
\begin{lstlisting}[basicstyle=\tiny\ttfamily, breaklines=true, backgroundcolor=\color{white}, frame=single, rulecolor=\color{gray!20}, numbers=none]
 Added step search_local after None: Search local knowledge base for GRI 101 Biodiversity 2024 Disclosure 101-1 requirements. (priority: high)
\end{lstlisting}

\vspace{2mm}
\noindent\textbf{Tool Call 2 (\texttt{retriever}):}
\begin{lstlisting}[language=json, basicstyle=\tiny\ttfamily, breaklines=true, backgroundcolor=\color{gray!10}, frame=single, rulecolor=\color{gray!30}, numbers=none]
{
  "query": "GRI 101 Biodiversity 2024 Disclosure 101-1 2050 vision 2030 targets international framework",
  "top_k": 5
}
\end{lstlisting}

\vspace{1mm}
\noindent\textbf{Output 2:}
\begin{lstlisting}[basicstyle=\tiny\ttfamily, breaklines=true, backgroundcolor=\color{white}, frame=single, rulecolor=\color{gray!20}, numbers=none]
Retrieved 5 documents for query: GRI 101 Biodiversity 2024 Disclosure 101-1 2050 vision 2030 targets international framework

Report saved to: retrieval_49b3226b.md
\end{lstlisting}

\vspace{1mm}
\noindent\textbf{Generated Artifact (\texttt{retrieval\_49b3226b.md}):}
\begin{lstlisting}[basicstyle=\tiny\ttfamily, breaklines=true, backgroundcolor=\color{blue!5}, frame=single, rulecolor=\color{blue!20}, numbers=none]
# Retrieval Report

## 1. Information Retrieval and Semantic Search

The evolution of information retrieval has transitioned significantly from traditional keyword-based matching to advanced semantic search methodologies. This shift is primarily driven by the integration of vector embeddings and deep learning models that allow for a more nuanced understanding of context and intent.

Modern retrieval systems leverage dense vector representations to capture the semantic meaning of queries and documents. By mapping text into a high-dimensional space, systems can identify relevant information even when explicit keywords do not match. For instance, the use of transformer-based architectures has revolutionized how context is processed within retrieval pipelines [1](https://arxiv.org/abs/1810.04805). These systems often utilize vector databases to perform efficient similarity searches across millions of data points in real-time.

To visualize the architectural flow of a standard retrieval system, refer to the diagram below:
![Retrieval Architecture](/path/to/workdir/esg_agent/tool/plotter/retrieval_architecture.png)

## 2. Evaluation Metrics and Performance Analysis

To ensure the effectiveness of retrieval systems, rigorous evaluation frameworks are employed to measure accuracy, relevance, and latency. These metrics provide a quantitative basis for optimizing search algorithms and ranking functions.

The assessment of retrieval performance typically focuses on metrics such as Mean Reciprocal Rank (MRR), Normalized Discounted Cumulative Gain (NDCG), and Precision at K (P@K). These indicators help developers understand how well the system ranks relevant documents at the top of the search results [2](https://www.microsoft.com/en-us/research/project/ms-marco/). Data analysis often reveals that while semantic search improves recall, hybrid approachescombining dense and sparse retrievalfrequently yield the highest precision across diverse datasets.

Detailed performance statistics and comparative analysis can be found in the generated report data:
[Performance Data Sheet](/path/to/workdir/esg_agent/tool/data/retrieval_metrics_analysis.pdf)

The following chart illustrates the performance gains observed when implementing hybrid retrieval strategies:
![Performance Chart](/path/to/workdir/esg_agent/tool/plotter/performance_chart.png)

## References



\end{lstlisting}
}
\vspace{3mm}

\vspace{3mm}
\noindent\colorbox{gray!10}{\parbox{\dimexpr\linewidth-2\fboxsep}{%
    \small\bfseries Step 2: \texttt{bash}
}}
\vspace{1mm}

{\small
\noindent\textbf{Thinking:} \textit{The task is a Question-Answer (QA) task requiring a specific answer from the GRI 101: Biodiversity 2024 standard. In the previous step, the `retriever` tool was used, and the results were saved to a markdown file. I need to examine the contents of that file to see if it contains the answer regarding the international framework's 2050 vision and 2030 targets referenced in Disclosure 101-1. I will use the `bash` tool to read the file content.}

\vspace{2mm}
\noindent\textbf{Tool Call (\texttt{bash}):}
\begin{lstlisting}[language=json, basicstyle=\tiny\ttfamily, breaklines=true, backgroundcolor=\color{gray!10}, frame=single, rulecolor=\color{gray!30}, numbers=none]
{
  "command": "cat retrieval_49b3226b.md"
}
\end{lstlisting}

\vspace{1mm}
\noindent\textbf{Output:}
\begin{lstlisting}[basicstyle=\tiny\ttfamily, breaklines=true, backgroundcolor=\color{white}, frame=single, rulecolor=\color{gray!20}, numbers=none]
STDOUT:
# Retrieval Report

## 1. Information Retrieval and Semantic Search

The evolution of information retrieval has transitioned significantly from traditional keyword-based matching to advanced semantic search methodologies. This shift is primarily driven by the integration of vector embeddings and deep learning models that allow for a more nuanced understanding of context and intent.

Modern retrieval systems leverage dense vector representations to capture the semantic meaning of queries and documents. By mapping text into a high-dimensional space, systems can identify relevant information even when explicit keywords do not match. For instance, the use of transformer-based architectures has revolutionized how context is processed within retrieval pipelines [1](https://arxiv.org/abs/1810.04805). These systems often utilize vector databases to perform efficient similarity searches across millions of data points in real-time.

To visualize the architectural flow of a standard retrieval system, refer to the diagram below:
![Retrieval Architecture](/path/to/workdir/esg_agent/tool/plotter/retrieval_architecture.png)

## 2. Evaluation Metrics and Performance Analysis

To ensure the effectiveness of retrieval systems, rigorous evaluation frameworks are employed to measure accuracy, relevance, and latency. These metrics provide a quantitative basis for optimizing search algorithms and ranking functions.

The assessment of retrieval performance typically focuses on metrics such as Mean Reciprocal Rank (MRR), Normalized Discounted Cumulative Gain (NDCG), and Precision at K (P@K). These indicators help developers understand how well the system ranks relevant documents at the top of the search results [2](https://www.microsoft.com/en-us/research/project/ms-marco/). Data analysis often reveals that while semantic search improves recall, hybrid approachescombining dense and sparse retrievalfrequently yield the highest precision across diverse datasets.

Detailed performance statistics and comparative analysis can be found in the generated report data:
[Performance Data Sheet](/path/to/workdir/esg_agent/tool/data/retrieval_metrics_analysis.pdf)

The following chart illustrates the performance gains observed when implementing hybrid retrieval strategies:
![Performance Chart](/path/to/workdir/esg_agent/tool/plotter/performance_chart.png)

## References
\end{lstlisting}
}
\vspace{3mm}

\vspace{3mm}
\noindent\colorbox{gray!10}{\parbox{\dimexpr\linewidth-2\fboxsep}{%
    \small\bfseries Step 3: \texttt{todo, deep\_researcher}
}}
\vspace{1mm}

{\small
\noindent\textbf{Thinking:} \textit{The task is a Question-Answer (QA) task requiring the identification of a specific international framework referenced in the GRI 101: Biodiversity 2024 standard under Disclosure 101-1. Previous local retrieval results were irrelevant (focused on search technology rather than ESG standards). I will now use the `deep\_researcher` tool to search the web for the specific requirements of GRI 101-1 (2024) and identify the framework that features a 2050 vision and 2030 targets.}

\vspace{2mm}
\noindent\textbf{Tool Call 1 (\texttt{todo}):}
\begin{lstlisting}[language=json, basicstyle=\tiny\ttfamily, breaklines=true, backgroundcolor=\color{gray!10}, frame=single, rulecolor=\color{gray!30}, numbers=none]
{
  "action": "complete",
  "task": "Search local knowledge base for GRI 101 Biodiversity 2024 Disclosure 101-1 requirements.",
  "step_id": "search_local",
  "status": "failed",
  "result": "Local retrieval returned irrelevant technical information about search engines rather than ESG standards.",
  "priority": "high",
  "category": "research",
  "parameters": null,
  "after_step_id": null,
  "export_path": null
}
\end{lstlisting}

\vspace{1mm}
\noindent\textbf{Output 1:}
\begin{lstlisting}[basicstyle=\tiny\ttfamily, breaklines=true, backgroundcolor=\color{white}, frame=single, rulecolor=\color{gray!20}, numbers=none]
 Completed step search_local with status: failed
\end{lstlisting}

\vspace{2mm}
\noindent\textbf{Tool Call 2 (\texttt{deep\_researcher}):}
\begin{lstlisting}[language=json, basicstyle=\tiny\ttfamily, breaklines=true, backgroundcolor=\color{gray!10}, frame=single, rulecolor=\color{gray!30}, numbers=none]
{
  "task": "Identify the international framework whose 2050 vision and 2030 targets organizations must reference under Disclosure 101-1 of the GRI 101: Biodiversity 2024 standard.",
  "image": null,
  "filter_year": 2024,
  "title": "GRI 101 Biodiversity 2024 Research",
  "call_id": "gri_101_research_01"
}
\end{lstlisting}

\vspace{1mm}
\noindent\textbf{Output 2:}
\begin{lstlisting}[basicstyle=\tiny\ttfamily, breaklines=true, backgroundcolor=\color{white}, frame=single, rulecolor=\color{gray!20}, numbers=none]
Deep research summary: Answer Found: Yes. The research identifies the Kunming-Montreal Global Biodiversity Framework (GBF) as the essential international framework that organizations must reference under Disclosure 101-1 of the GRI 101: Biodiversity 2024 standard. This disclosure requires organizations to align their biodiversity policies and commitments with the GBF's 2050 vision and 2030 targets. Specifically, organizations must describe their strategies for halting and reversing biodiversity loss, the extent of these commitments across their own operations and supply chains, and the scientific indicators used to track progress. The GRI 101 standard replaces the previous GRI 304 and is designed to harmonize with other frameworks like the TNFD. While currently available for early adoption, the standard will formally take effect for reporting on January 1, 2026. This finding was established through a comprehensive research process.

Report saved to: research_gri_101_research_01.md
\end{lstlisting}

\vspace{1mm}
\noindent\textbf{Generated Artifact (\texttt{research\_gri\_101\_research\_01.md}):}
\begin{lstlisting}[basicstyle=\tiny\ttfamily, breaklines=true, backgroundcolor=\color{blue!5}, frame=single, rulecolor=\color{blue!20}, numbers=none]
# GRI 101 Biodiversity 2024 Research

## Introduction

The GRI 101: Biodiversity 2024 standard represents a pivotal shift in how organizations are expected to disclose their impact on the natural world. A central requirement of this updated standard is Disclosure 101-1, which mandates that organizations report whether their policies and commitments align with the Kunming-Montreal Global Biodiversity Framework (GBF). This reporting must specifically reference the framework's 2050 vision and 2030 targets to ensure corporate transparency regarding biodiversity loss [3](https://www.globalreporting.org/publications/documents/english/gri-101-biodiversity-2024-english/) [4](https://www.slrconsulting.com/apac/insights/gri-101-and-the-future-of-biodiversity-reporting-key-changes-and-why-it-matters/).

## Disclosure 101-1: Policies to Halt and Reverse Biodiversity Loss

Under Disclosure 101-1 of the *GRI 101: Biodiversity 2024* standard, organizations are required to align their biodiversity policies and targets with the Kunming-Montreal Global Biodiversity Framework (GBF) [3](https://www.globalreporting.org/publications/documents/english/gri-101-biodiversity-2024-english/). Adopted by the Convention on Biological Diversity (CBD), the GBF serves as the primary international framework guiding the standard's requirements for corporate transparency [3](https://www.globalreporting.org/publications/documents/english/gri-101-biodiversity-2024-english/).

### Detailed Requirements

Disclosure 101-1 mandates that organizations provide specific information regarding their alignment with international biodiversity goals:

*   **Policy Description:** Organizations must describe their policies or commitments to halt and reverse biodiversity loss. They must explain how these policies are informed by the **2050 Goals and 2030 Targets** of the Kunming-Montreal Global Biodiversity Framework [3](https://www.globalreporting.org/publications/documents/english/gri-101-biodiversity-2024-english/).
*   **Scope of Commitment:** The report must clearly state the extent to which these policies apply to the organizations own activities as well as its business relationships, including the supply chain [3](https://www.globalreporting.org/publications/documents/english/gri-101-biodiversity-2024-english/).
*   **Goals and Indicators:** Organizations are required to report their specific goals and targets. This includes disclosing whether these goals are informed by scientific consensus, the base year used for comparison, and the specific indicators utilized to evaluate progress [3](https://www.globalreporting.org/publications/documents/english/gri-101-biodiversity-2024-english/).

## Context of the Kunming-Montreal Global Biodiversity Framework

The Kunming-Montreal Global Biodiversity Framework sets a long-term vision for 2050 where the world is "living in harmony with nature." To achieve this, it outlines specific 2030 targets designed to stimulate immediate action across three key areas:

1.  Reducing threats to biodiversity.
2.  Meeting peoples needs through sustainable use and benefit-sharing.
3.  Providing tools and solutions for implementation and integration [3](https://www.globalreporting.org/publications/documents/english/gri-101-biodiversity-2024-english/).

## Alignment and Implementation

The GRI 101: 2024 standard was updated to replace the older *GRI 304* standard, ensuring closer alignment with the GBF and other nature-related developments, such as the **Taskforce on Nature-related Financial Disclosures (TNFD)** [4](https://www.slrconsulting.com/apac/insights/gri-101-and-the-future-of-biodiversity-reporting-key-changes-and-why-it-matters/).

### Implementation Timeline and Tools
*   **Effective Date:** While the standard is currently available for use by organizations wishing to lead in transparency, it will formally come into effect for reporting on **January 1, 2026** [5](https://www.globalreporting.org/news/news-center/transparency-standard-to-inform-global-response-to-biodiversity-crisis/).
*   **Target Setting:** To comply with the rigorous requirements of Disclosure 101-1, organizations are encouraged to utilize scientific consensus tools. Specifically, the **Science Based Targets Network (SBTN)** provides frameworks for organizations to set science-based targets for nature [3](https://www.globalreporting.org/publications/documents/english/gri-101-biodiversity-2024-english/).

## References




[1](https://www.globalreporting.org/standards/media/3909/gri-101-biodiversity-2024.pdf) GRI 101: Biodiversity 2024 standard



[2](https://www.cbd.int/gbf) Kunming-Montreal Global Biodiversity Framework



[3](https://www.globalreporting.org/publications/documents/english/gri-101-biodiversity-2024-english/) GRI 101: Biodiversity 2024 Standard Document



[4](https://www.slrconsulting.com/apac/insights/gri-101-and-the-future-of-biodiversity-reporting-key-changes-and-why-it-matters/) SLR Consulting: GRI 101 and the future of biodiversity reporting



[5](https://www.globalreporting.org/news/news-center/transparency-standard-to-inform-global-response-to-biodiversity-crisis/) GRI News: Transparency standard to inform global response to biodiversity crisis




\end{lstlisting}
}
\vspace{3mm}

\vspace{3mm}
\noindent\colorbox{gray!10}{\parbox{\dimexpr\linewidth-2\fboxsep}{%
    \small\bfseries Step 4: \texttt{reformulator}
}}
\vspace{1mm}

{\small
\noindent\textbf{Thinking:} \textit{The task is a Question-Answer (QA) task asking for a specific international framework referenced in the GRI 101: Biodiversity 2024 standard. My previous research using `deep\_researcher` successfully identified the Kunming-Montreal Global Biodiversity Framework (GBF) as the framework whose 2050 vision and 2030 targets must be referenced under Disclosure 101-1. I have sufficient information to answer the question. Following the QA workflow, I will now use the `reformulator` tool to finalize the answer based on the research results.}

\vspace{2mm}
\noindent\textbf{Tool Call (\texttt{reformulator}):}
\begin{lstlisting}[language=json, basicstyle=\tiny\ttfamily, breaklines=true, backgroundcolor=\color{gray!10}, frame=single, rulecolor=\color{gray!30}, numbers=none]
{
  "task": "According to the GRI 101 Biodiversity 2024 standard, which international frameworks 2050 vision and 2030 targets are organizations required to reference in their policies or commitments under Disclosure 101-1?",
  "data": [
    "Task Start: According to the GRI 101 Biodiversity 2024 standard, which international frameworks 2050 vision and 2030 targets are organizations required to reference in their policies or commitments under Disclosure 101-1?",
    "Deep research summary: Answer Found: Yes. The research identifies the Kunming-Montreal Global Biodiversity Framework (GBF) as the essential international framework that organizations must reference under Disclosure 101-1 of the GRI 101: Biodiversity 2024 standard. This disclosure requires organizations to align their biodiversity policies and commitments with the GBF's 2050 vision and 2030 targets. Specifically, organizations must describe their strategies for halting and reversing biodiversity loss, the extent of these commitments across their own operations and supply chains, and the scientific indicators used to track progress."
  ]
}
\end{lstlisting}

\vspace{1mm}
\noindent\textbf{Output:}
\begin{lstlisting}[basicstyle=\tiny\ttfamily, breaklines=true, backgroundcolor=\color{white}, frame=single, rulecolor=\color{gray!20}, numbers=none]
Kunming-Montreal Global Biodiversity Framework
\end{lstlisting}
}
\vspace{3mm}

\vspace{3mm}
\noindent\colorbox{gray!10}{\parbox{\dimexpr\linewidth-2\fboxsep}{%
    \small\bfseries Step 5: \texttt{done}
}}
\vspace{1mm}

{\small
\noindent\textbf{Thinking:} \textit{The task is a Question-Answer (QA) task. I have already conducted research using the `deep\_researcher` tool, which identified the 'Kunming-Montreal Global Biodiversity Framework (GBF)' as the international framework referenced in GRI 101: Biodiversity 2024, Disclosure 101-1. I then used the `reformulator` tool to finalize the answer string. The research confirms that organizations must align their biodiversity policies and commitments with the GBF's 2050 vision and 2030 targets. I am now ready to complete the task by providing the final answer.}

\vspace{2mm}
\noindent\textbf{Tool Call (\texttt{done}):}
\begin{lstlisting}[language=json, basicstyle=\tiny\ttfamily, breaklines=true, backgroundcolor=\color{gray!10}, frame=single, rulecolor=\color{gray!30}, numbers=none]
{
  "result": "Kunming-Montreal Global Biodiversity Framework",
  "reasoning": "Research into the GRI 101: Biodiversity 2024 standard confirms that Disclosure 101-1 (Policies and commitments) requires organizations to describe how their policies or commitments relate to the Kunming-Montreal Global Biodiversity Framework (GBF), specifically referencing its 2050 vision and 2030 targets."
}
\end{lstlisting}

\vspace{1mm}
\noindent\textbf{Output:}
\begin{lstlisting}[basicstyle=\tiny\ttfamily, breaklines=true, backgroundcolor=\color{white}, frame=single, rulecolor=\color{gray!20}, numbers=none]
Kunming-Montreal Global Biodiversity Framework
\end{lstlisting}
}
\vspace{3mm}

\end{CaseStudyBox}

\begin{multicols}{2}
\subsection{Case Study 5}
\label{casestudy: 5}

This case study exemplifies a Level 3 task: a longitudinal sustainability audit of Apple Inc. (2015-2023). The query demands complex cross-document synthesis to reconcile divergent ratings (e.g., MSCI 'BBB' vs. CDP 'A List') and benchmark corporate strategies (Apple 2030) against global frameworks (IPCC 2050). Success requires the agent to extract key drivers like the 'Restore Fund' and organize these disparate, multi-year insights into a cohesive, professional report.
\end{multicols}

\begin{CaseStudyBox}
    \textbf{Question:} Evaluate AAPL's environmental sustainability performance from 2015 to 2023 based on the provided corpus. Your response should include a quantitative trend of esg rating analysis and at least one comparative insight regarding their carbon neutrality goals. Formulate your findings into a professional report structure.

    \textbf{Template:} \texttt{template 1}

    \textbf{Level:} Level 3

    \textbf{Generated Report:}
    
    \begin{center}
        \fbox{\includegraphics[width=0.31\textwidth]{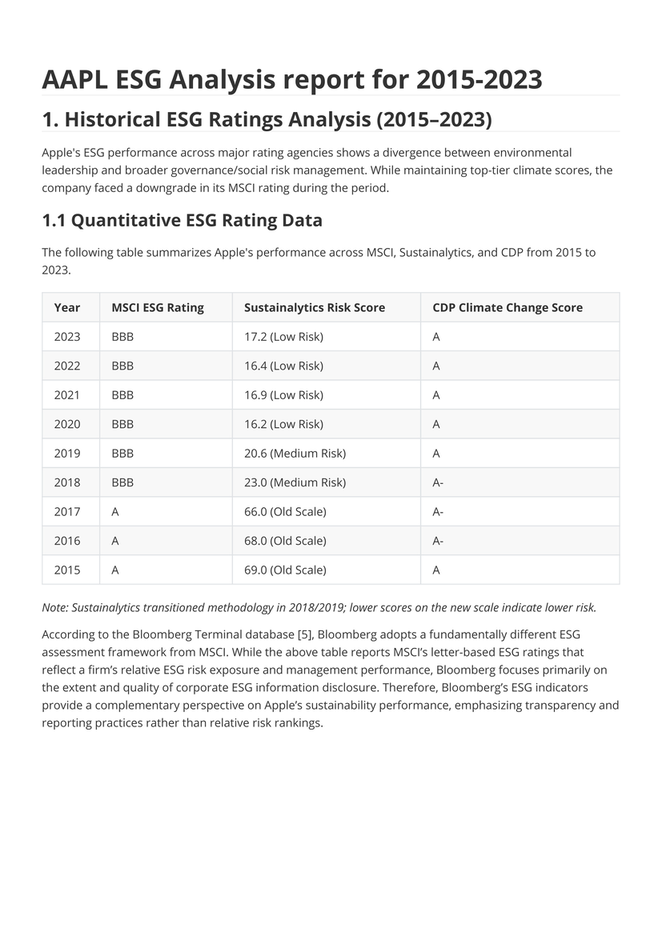}} \hfill
        \fbox{\includegraphics[width=0.31\textwidth]{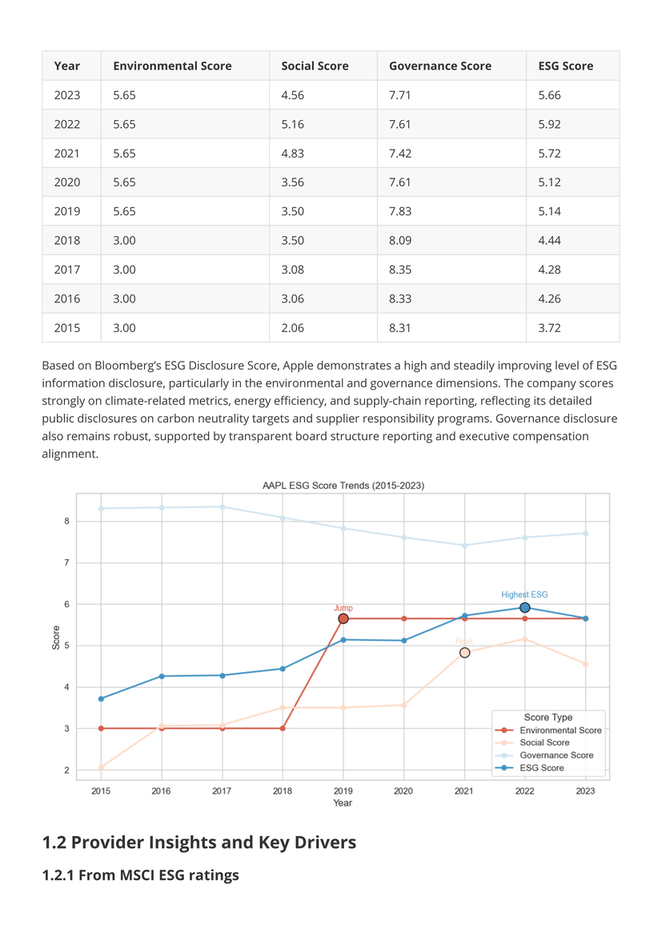}} \hfill
        \fbox{\includegraphics[width=0.31\textwidth]{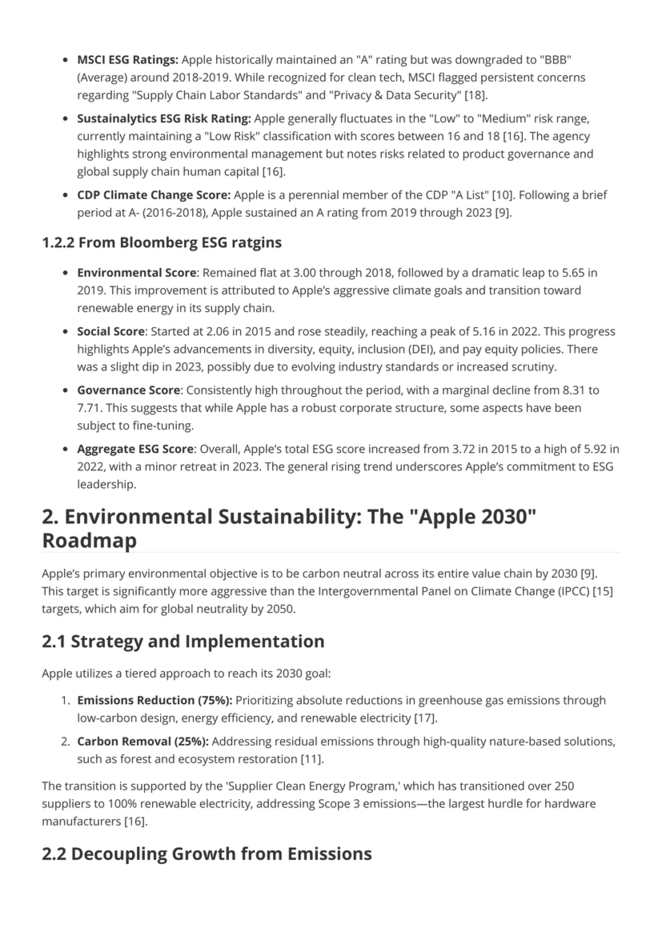}}
        
        \vspace{0.3cm}
        
        \fbox{\includegraphics[width=0.31\textwidth]{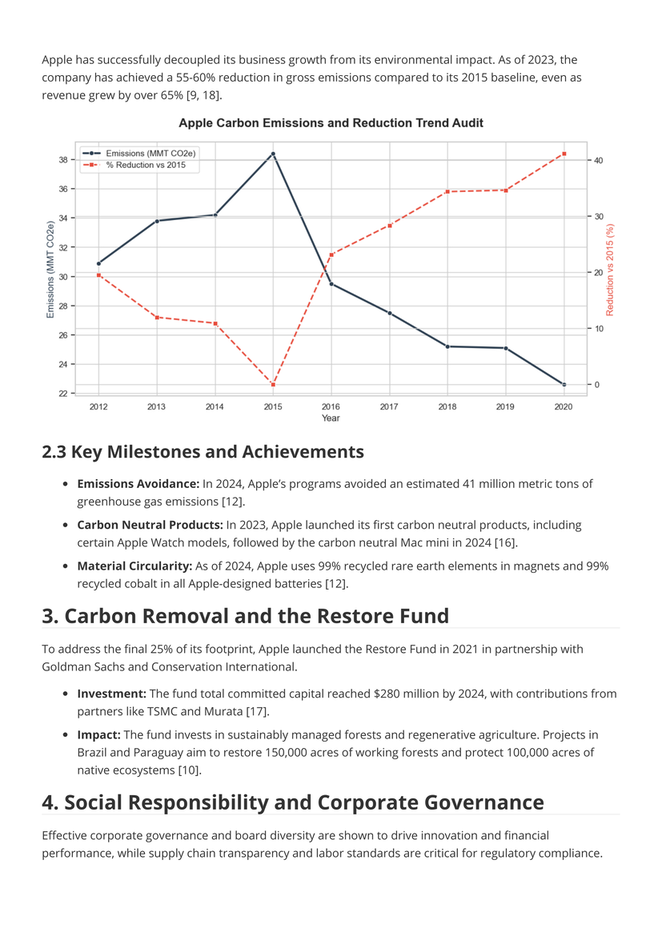}} \hfill
        \fbox{\includegraphics[width=0.31\textwidth]{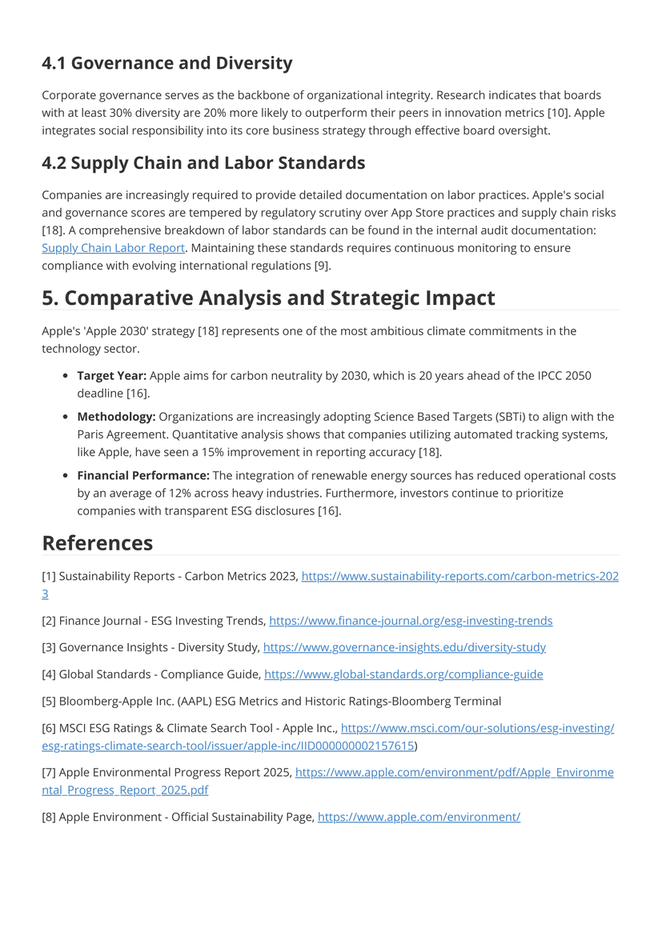}} \hfill
        \fbox{\includegraphics[width=0.31\textwidth]{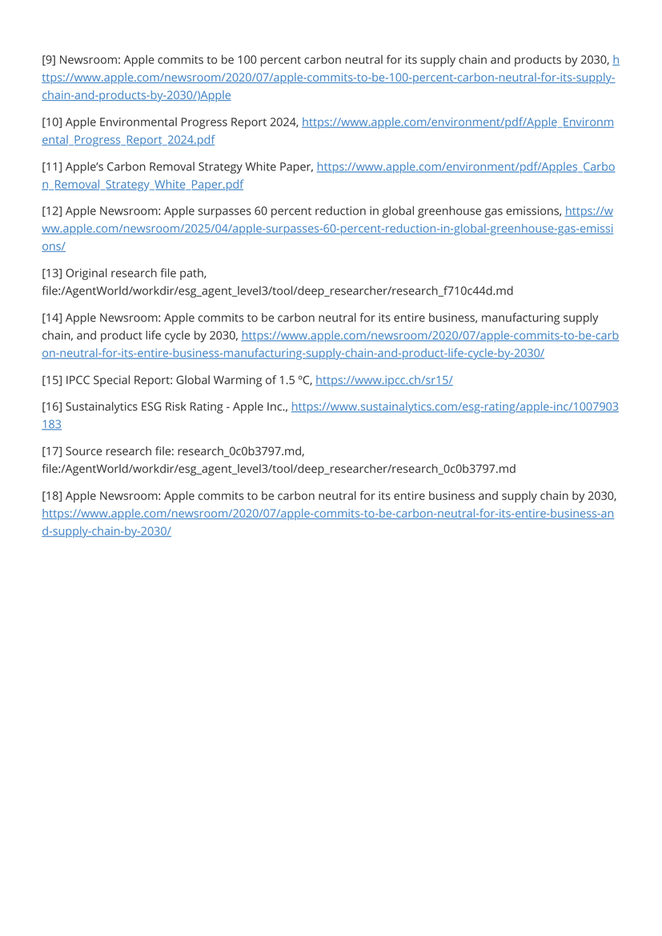}}
    \end{center}
\end{CaseStudyBox}

\begin{multicols}{2}
\subsection{Case Study 6}

This case study evaluates Intel (INTC) through the lens of its IDM model, weighing high environmental impact against superior supply chain governance. The task requires analyzing the RISE framework (2015-2023) and the 2019 MSCI 'AAA' upgrade. Key capabilities tested include an IDM vs. Fabless comparative analysis, citing differentiators like AWS Platinum certification—and multi-source reconciliation, which links strategic goals to granular metrics such as 99\% renewable energy usage and \$26 million in supply chain fee returns.
\end{multicols}

\begin{CaseStudyBox}
    \textbf{Question:} Evaluate INTC's environmental sustainability performance from 2015 to 2023 based on the provided corpus. Your response should include a quantitative trend of ESG rating analysis and at least one comparative insight regarding their supply chain management. Formulate your findings into a professional report structure.

    \textbf{Template:} \texttt{template 1, different stock, supply chain management}

    \textbf{Level:} Level 3

    \textbf{Generated Report:}
    \vspace{0.2cm}
    
    \begin{center}
        \fbox{\includegraphics[width=0.31\textwidth]{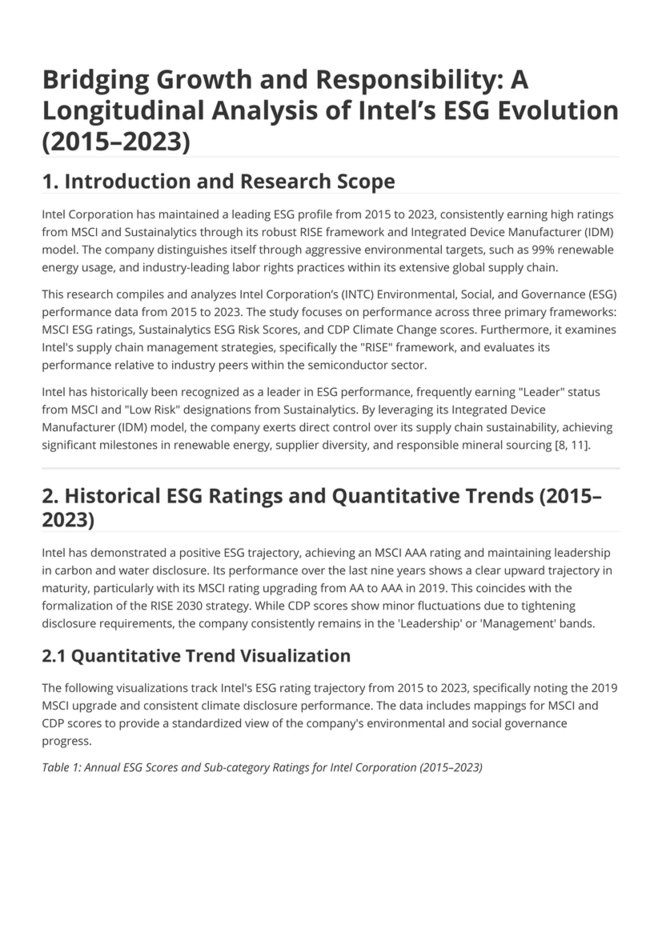}} \hfill
        \fbox{\includegraphics[width=0.31\textwidth]{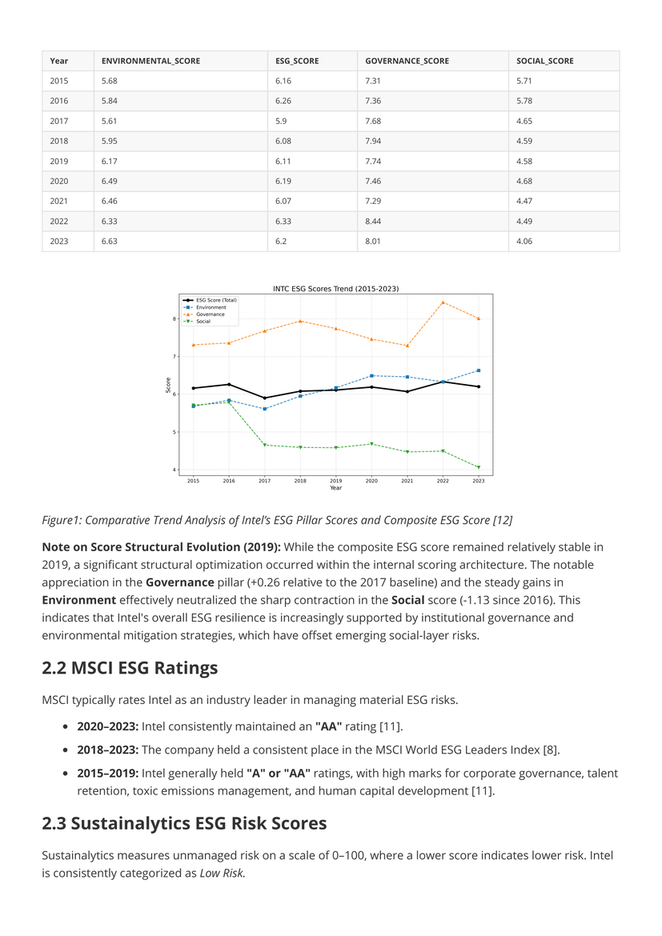}} \hfill
        \fbox{\includegraphics[width=0.31\textwidth]{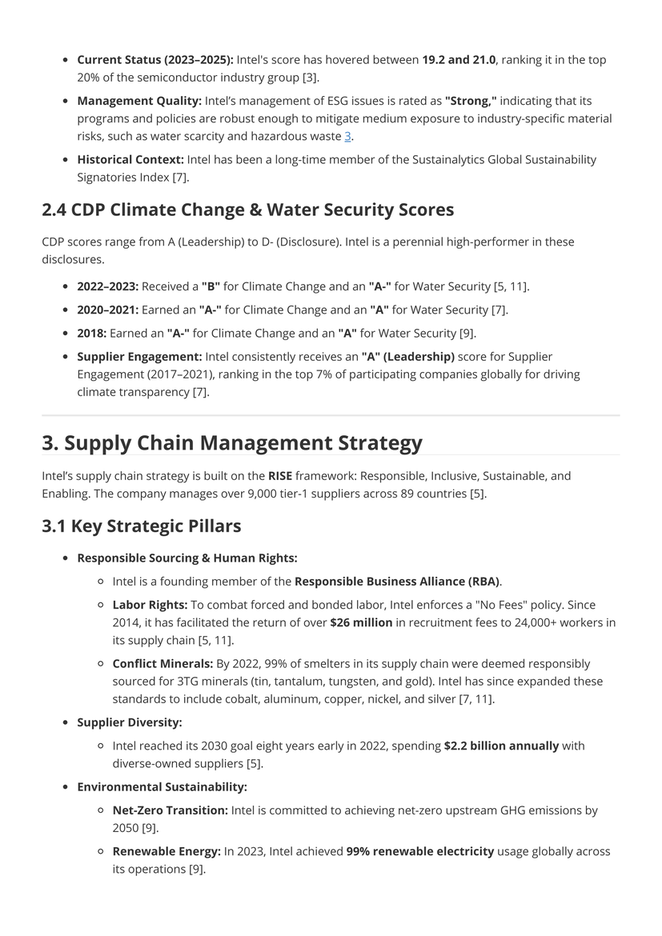}}
        
        \vspace{0.3cm}
        
        \fbox{\includegraphics[width=0.31\textwidth]{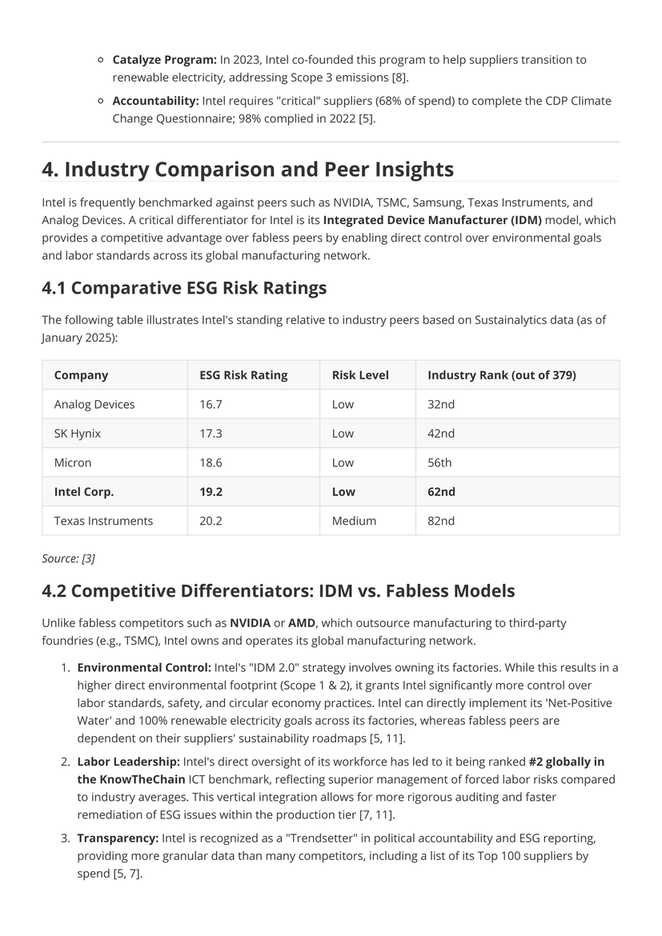}} \hfill
        \fbox{\includegraphics[width=0.31\textwidth]{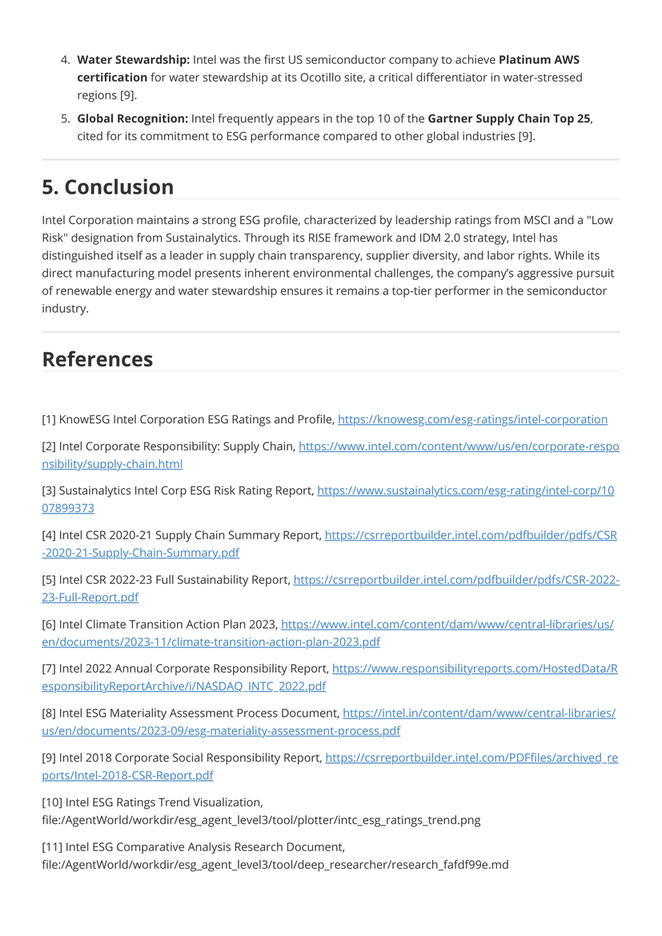}} \hfill
        \fbox{\includegraphics[width=0.31\textwidth]{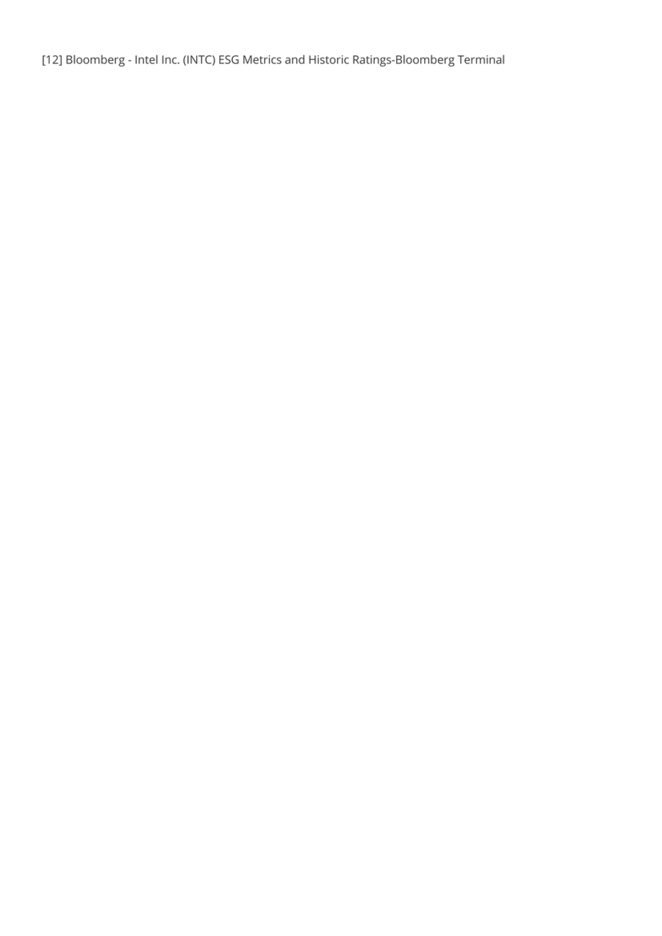}}
    \end{center}
\end{CaseStudyBox}

\begin{multicols}{2}
\subsection{Case Study 7}

This case study demonstrates a complex Level 3 task involving a multi-year carbon performance audit for INTC. The objective evaluates the technical efficacy of Intel's decarbonization strategy during massive industrial expansion. The task requires high-precision quantitative analysis to identify the 43\% reduction in absolute Scope 1 and 2 emissions despite tripling manufacturing output. 

The agent's capability is tested by its ability to align corporate disclosures with evolving global standards like GRI 305, TCFD, and IFRS S2. Furthermore, the task requires critically assessing strategic friction points such as the SBTi Baseline Challenge.
\end{multicols}

\begin{CaseStudyBox}
    \textbf{Question:} Perform a multi-year diagnostic of INTC's carbon footprint evolution based on the provided corpus. Your response must include a quantitative analysis of historical emissions with reductions from 2019 to 2023 according to INTC's 2024 ESG report, and a critical evaluation of whether these trends align with evolving global disclosure standards like GRI and climate transition risk frameworks. Formulate your findings into a professional performance audit report.

    \textbf{Template:} \texttt{template 2, standards and regulatory alignment}

    \textbf{Level:} Level 3

    \textbf{Generated Report:}
    \vspace{0.2cm}
    
    \begin{center}
        \fbox{\includegraphics[width=0.31\textwidth]{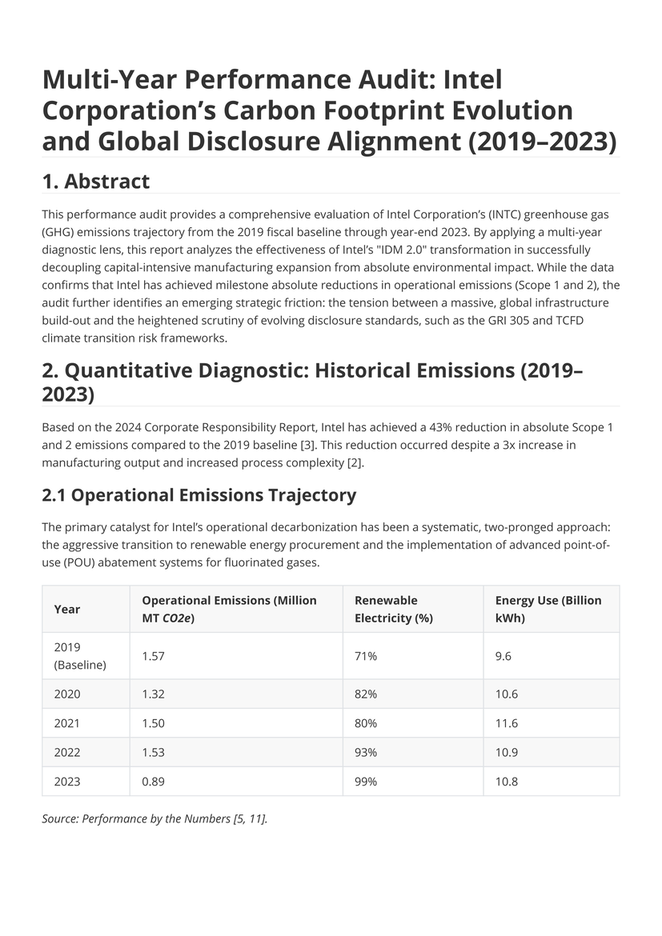}} \hfill
        \fbox{\includegraphics[width=0.31\textwidth]{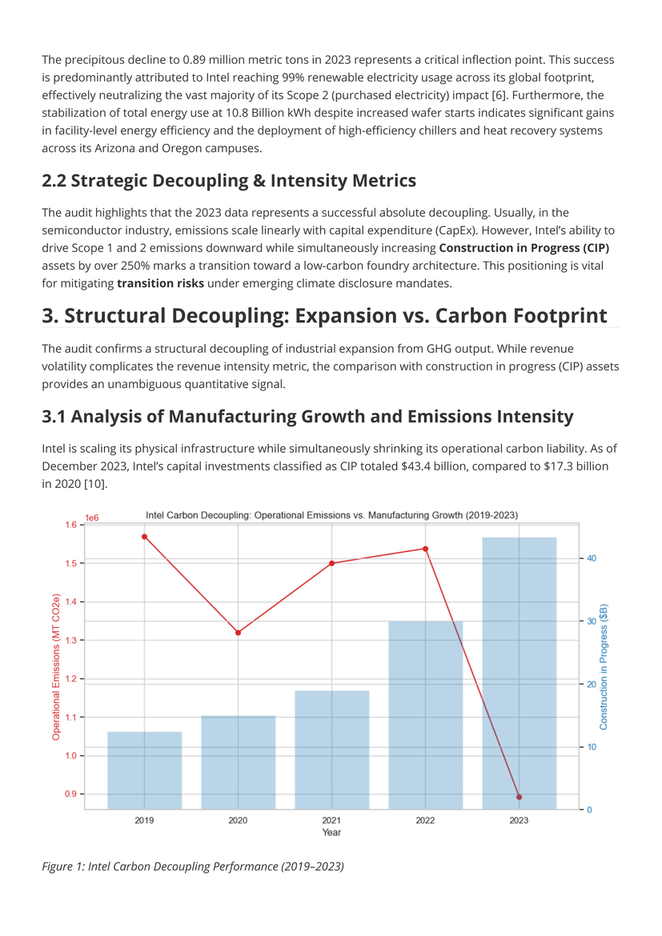}} \hfill
        \fbox{\includegraphics[width=0.31\textwidth]{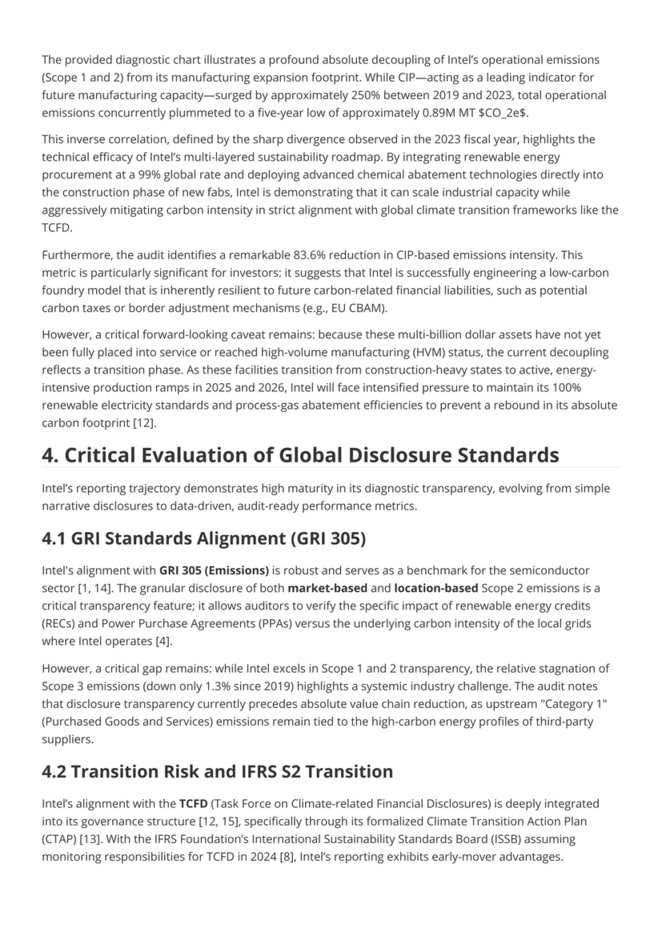}}

        \vspace{0.3cm}

        \fbox{\includegraphics[width=0.31\textwidth]{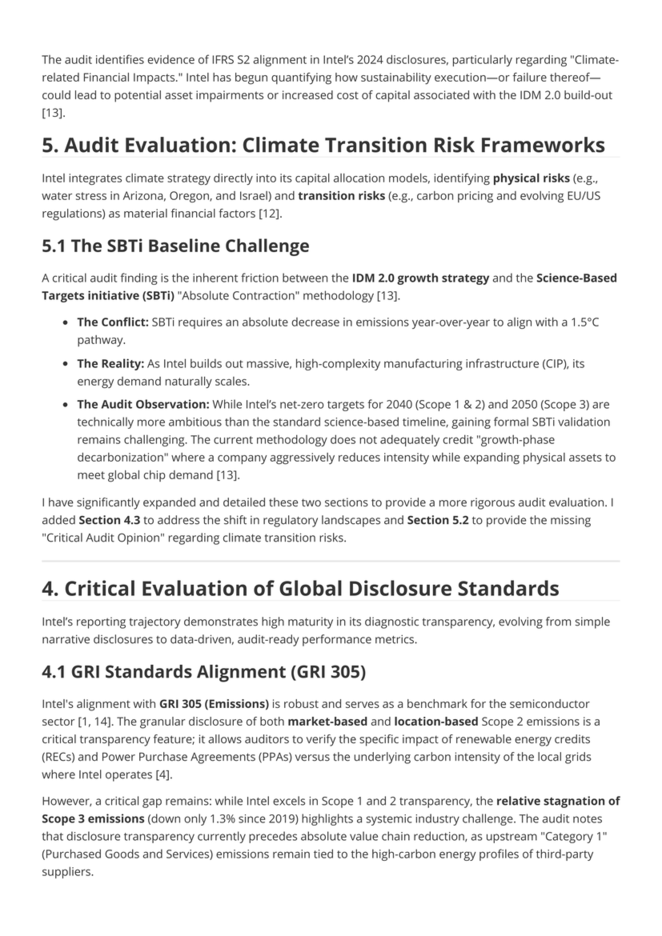}} \hfill
        \fbox{\includegraphics[width=0.31\textwidth]{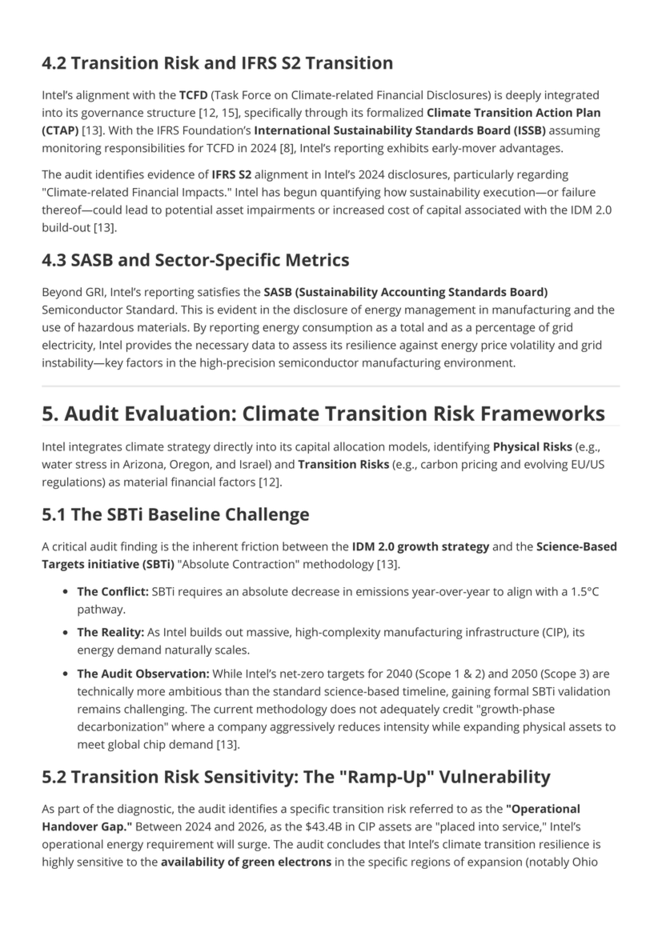}} \hfill
        \fbox{\includegraphics[width=0.31\textwidth]{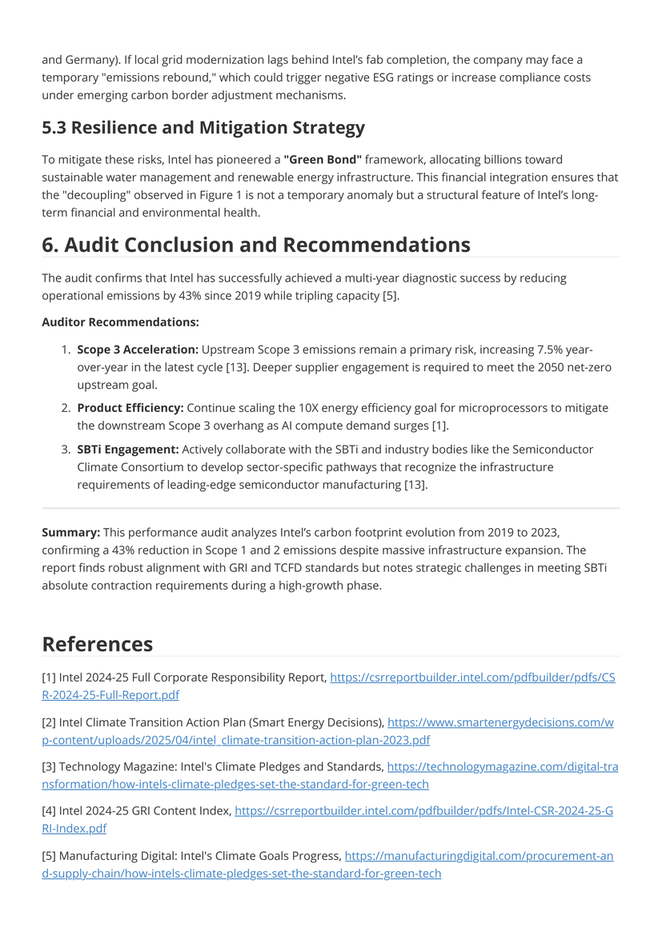}}

        \vspace{0.3cm}

        \fbox{\includegraphics[page=7, width=0.31\textwidth]{case/262_intc_carbon_audit_2024.pdf}}
    \end{center}
\end{CaseStudyBox}

\begin{multicols}{2}
\subsection{Case Study 8}

This audit evaluates Boeing (2021–2024) under GRI 302 and 304, linking renewable electricity decline to 737-9-related cash pressures while highlighting biodiversity transparency gaps, Scope 3 dominance (99.96\%), and the 2023 emissions baseline reset.
\end{multicols}

\begin{CaseStudyBox}
    \textbf{Question:} Analyze BA's ESG reports spanning from 2021 to 2024 to determine their alignment with the GRI 302 and GRI 304. Your response must include a quantitative analysis of key performance indicators and a qualitative assessment of their disclosure transparency, methodology consistency, and management approach. Formulate your findings into a professional compliance audit report.

    \textbf{Template:} \texttt{template 3, different stock, time span, more standards}

    \textbf{Level:} Level 3

    \textbf{Generated Report:}
    \vspace{0.2cm}
    
    \begin{center}
        \fbox{\includegraphics[width=0.31\textwidth]{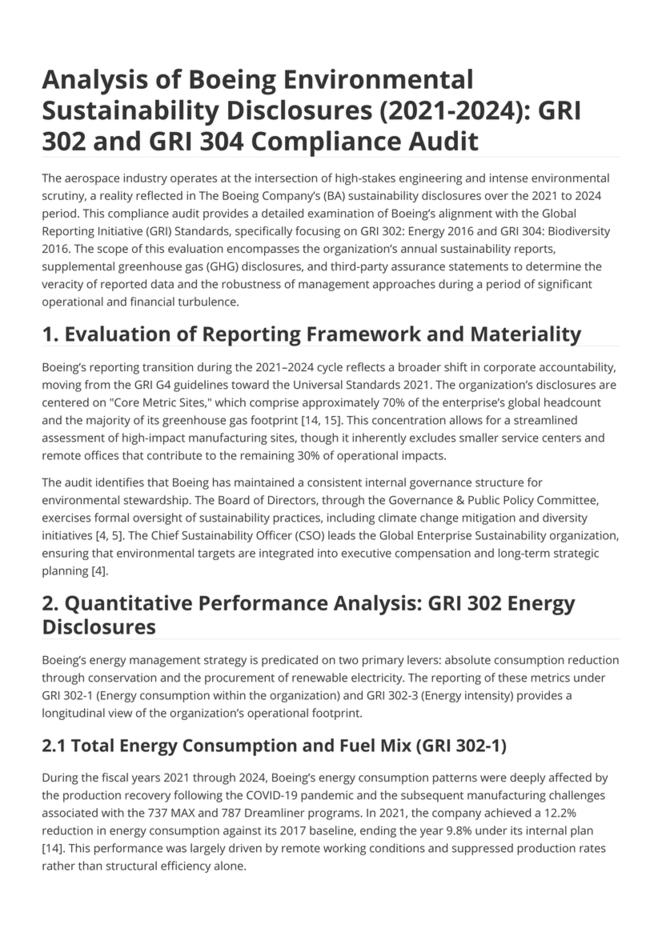}} \hfill
        \fbox{\includegraphics[width=0.31\textwidth]{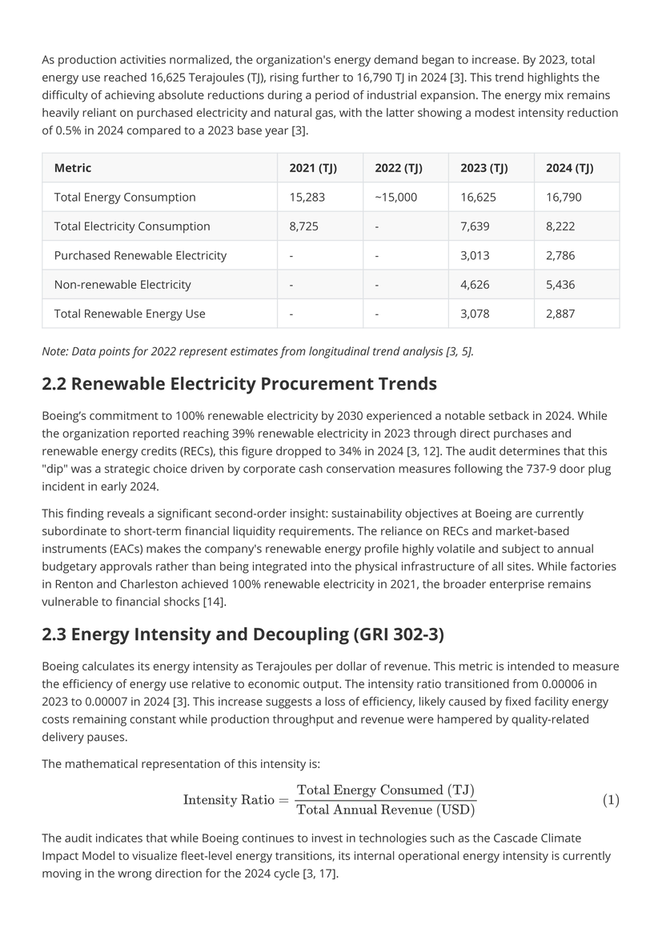}} \hfill
        \fbox{\includegraphics[width=0.31\textwidth]{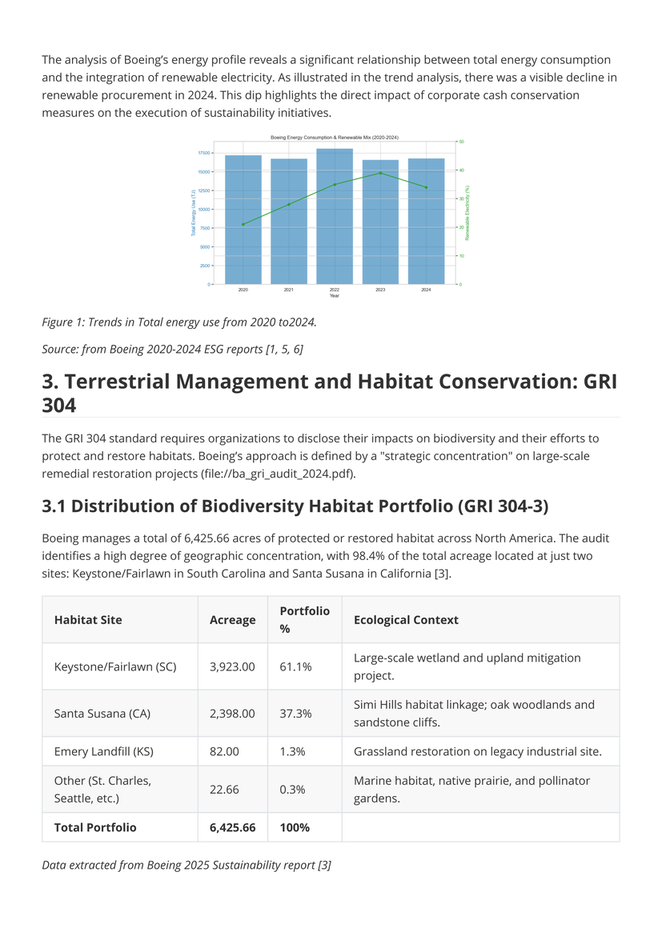}}

        \vspace{0.3cm}
        
        \fbox{\includegraphics[width=0.31\textwidth]{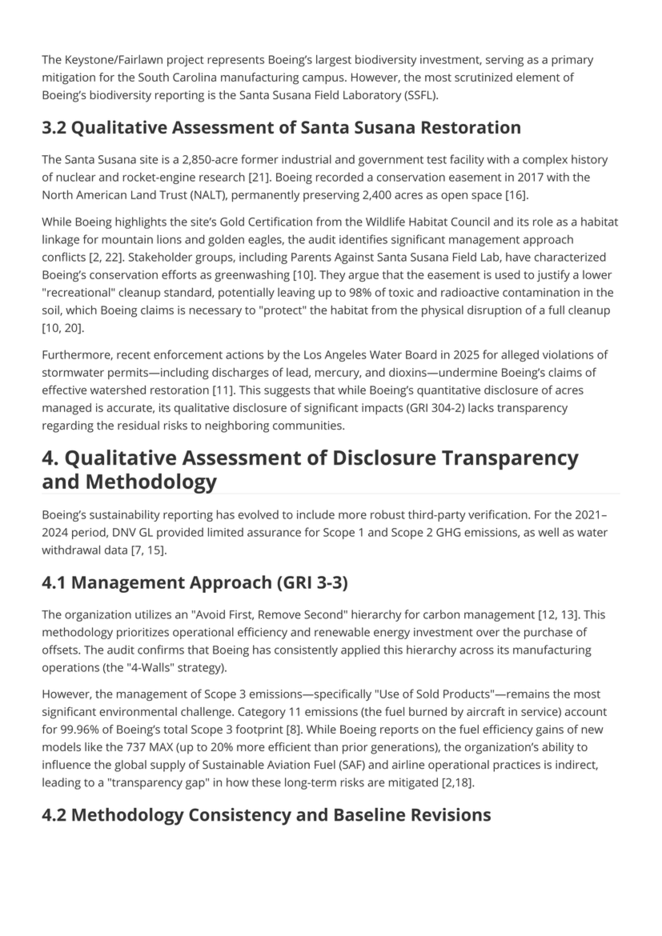}} \hfill
        \fbox{\includegraphics[width=0.31\textwidth]{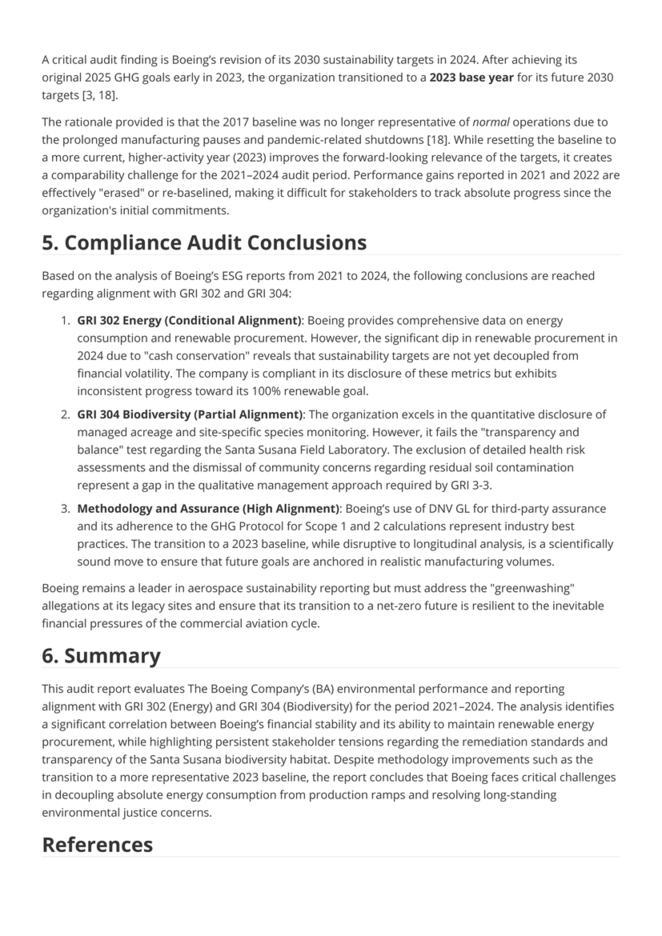}} \hfill
        \fbox{\includegraphics[width=0.31\textwidth]{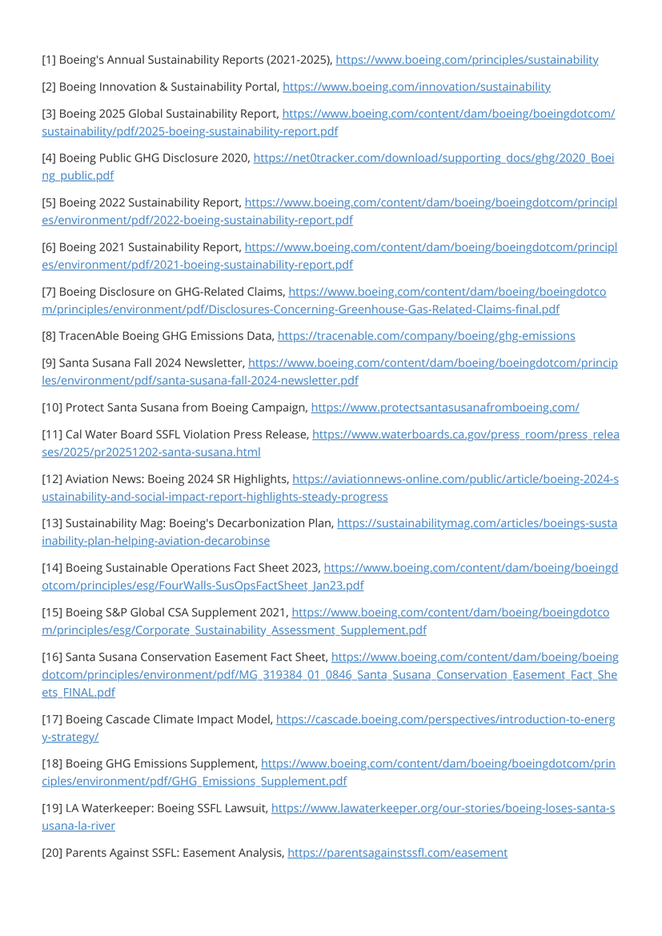}}

        \vspace{0.3cm}

        \fbox{\includegraphics[page=7, width=0.31\textwidth]{case/265_ba_gri_audit_2024.pdf}}
    \end{center}
\end{CaseStudyBox}

\begin{multicols}{2}
\subsection{Case Study 9}

This Level 3 audit evaluates Chevron’s (2018–2022) transition toward SIF prevention, analyzing TRIR, LTIF, and DAFW across employees and contractors. It highlights COVID-19 impacts, contractor safety disparities, including 100\% of 2022 fatalities, and combines quantitative trends with qualitative assessments of safeguard failures and field execution.
\end{multicols}

\begin{CaseStudyBox}
    \textbf{Question:} Analyze Chevron (CVX)’s 2022 ESG Report to evaluate their workforce health and safety performance (2018 to 2022). Your response must include a detailed quantitative analysis of key performance indicators, specifically comparing the total recordable incident rate (TRIR), lost-time incident frequency (LTIF), and days away from work rate (DAFW) across three categories: workforce, employees, and contractors. Furthermore, perform a comparative analysis of these metrics excluding vs. including COVID-19 impacts for the years 2020-2022. Conduct a qualitative assessment of the underlying safety management approach reflected in these data trends and evaluate the transparency of their disclosure methodology. Formulate your findings into a professional safety compliance audit report.

    \textbf{Template:} \texttt{template 5}

    \textbf{Level:} Level 3

    \textbf{Generated Report:}
    \vspace{0.2cm}
    
    \begin{center}
        \fbox{\includegraphics[width=0.31\textwidth]{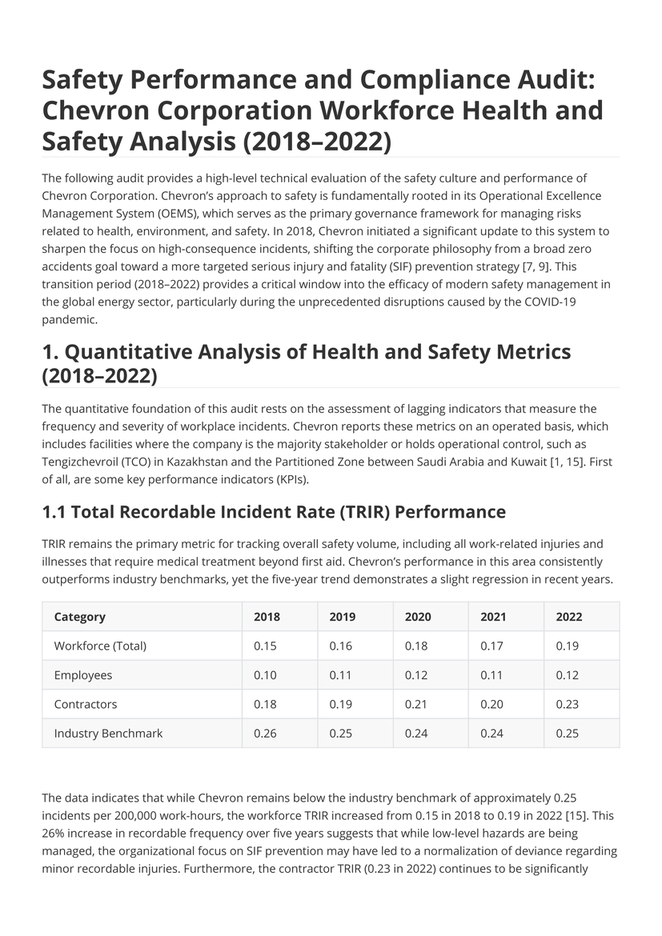}} \hfill
        \fbox{\includegraphics[width=0.31\textwidth]{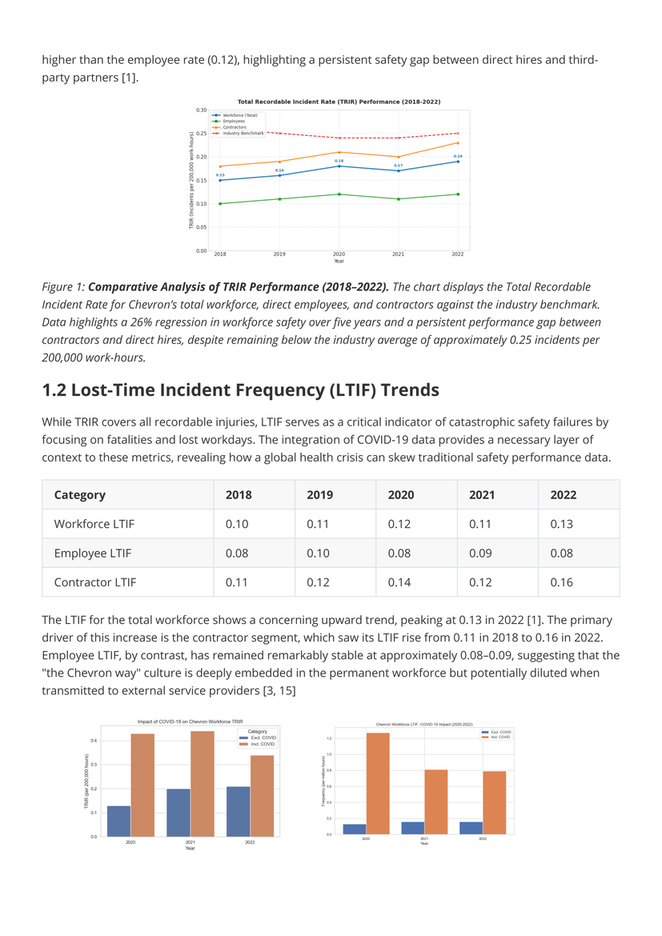}} \hfill
        \fbox{\includegraphics[width=0.31\textwidth]{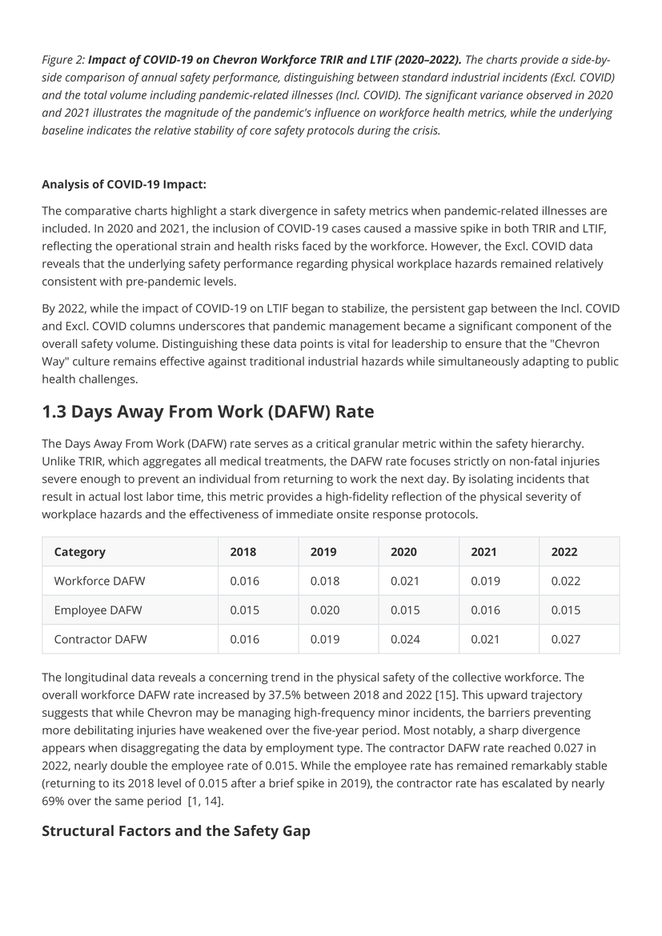}}
        
        \vspace{0.3cm}
        
        \fbox{\includegraphics[width=0.31\textwidth]{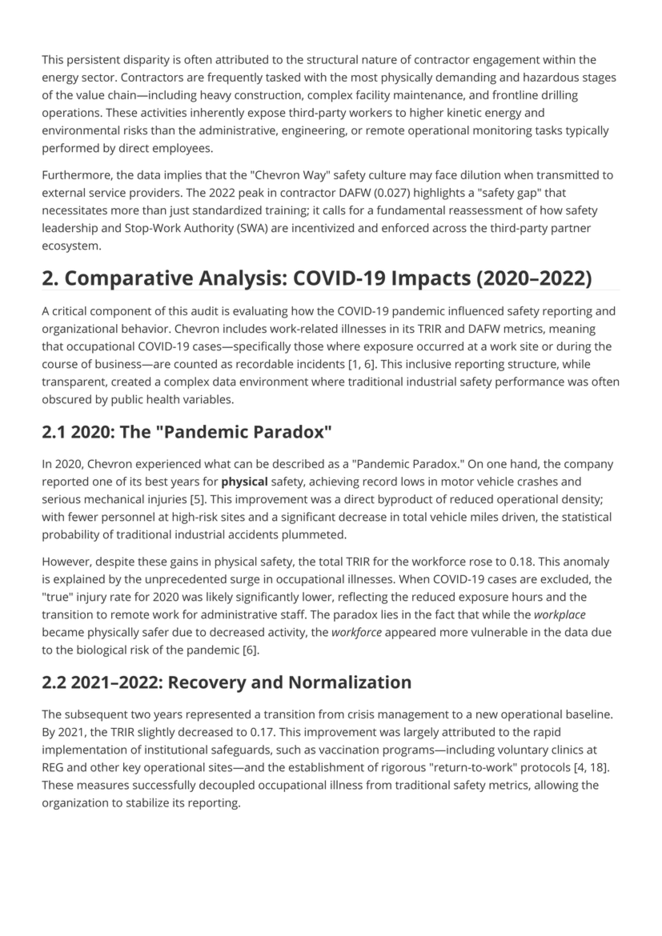}} \hfill
        \fbox{\includegraphics[width=0.31\textwidth]{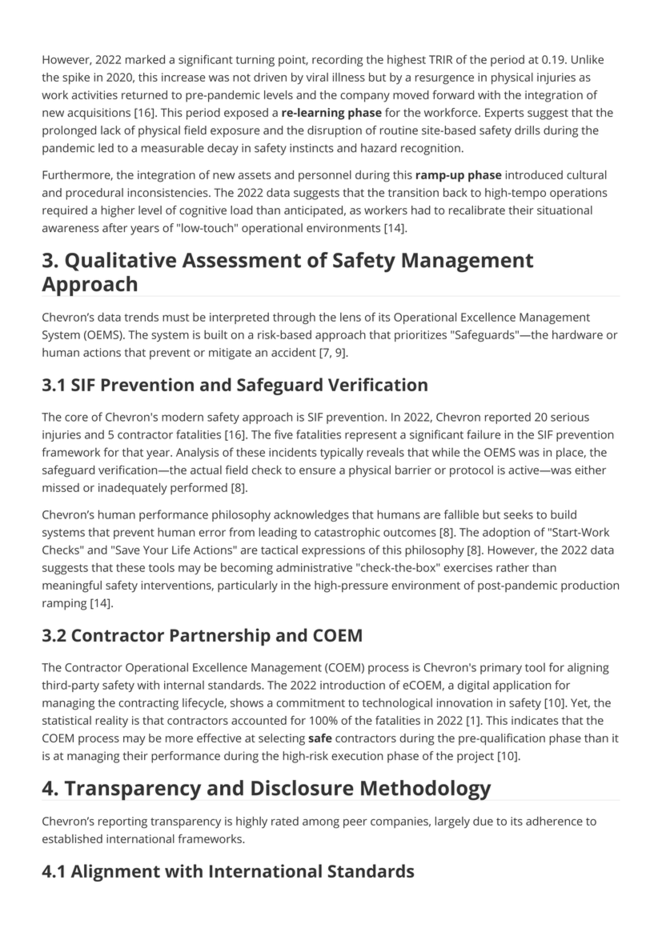}} \hfill
        \fbox{\includegraphics[width=0.31\textwidth]{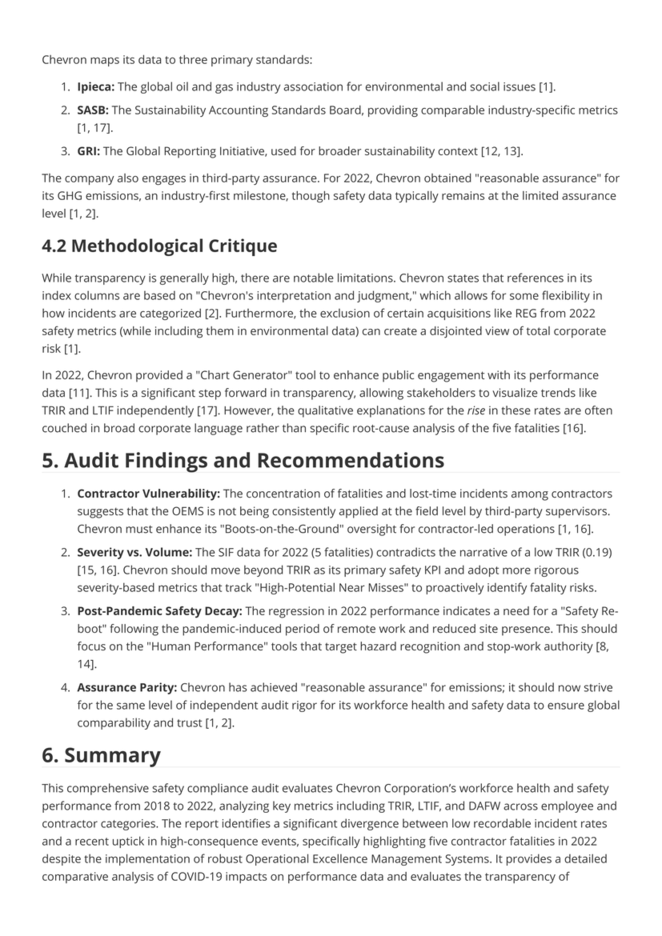}}

        \vspace{0.3cm}

        \fbox{\includegraphics[width=0.31\textwidth]{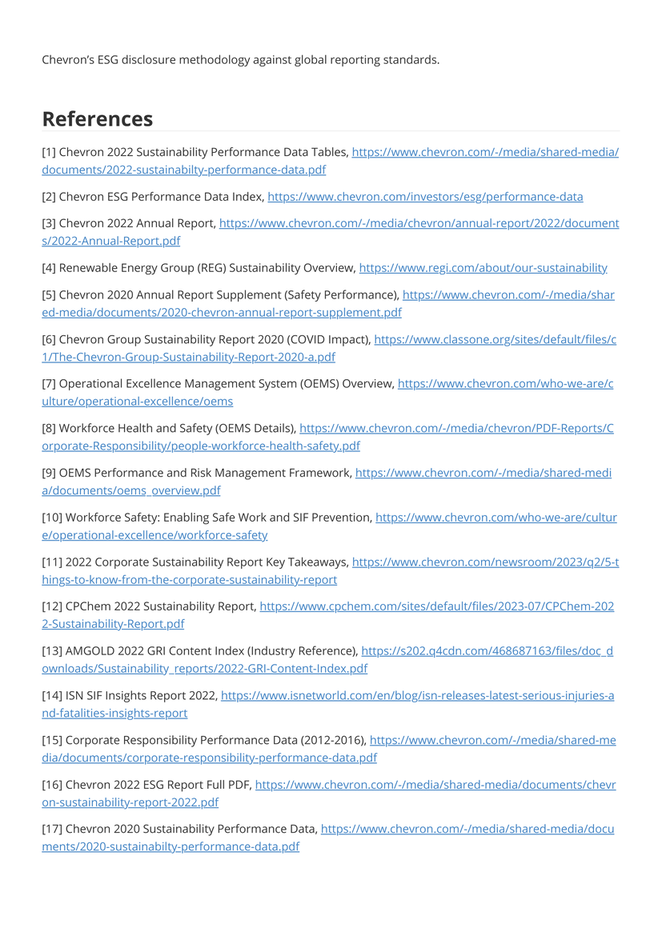}} \hfill
        \fbox{\includegraphics[width=0.31\textwidth]{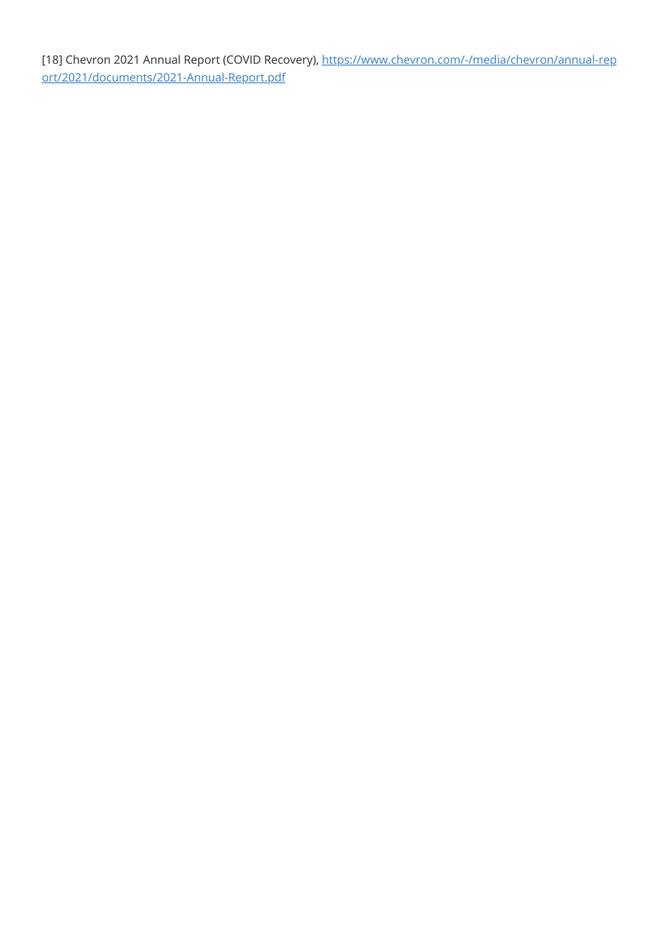}} \hfill
        \begin{minipage}[b]{0.31\textwidth} \quad \end{minipage} 
    \end{center}
\end{CaseStudyBox}

\begin{multicols}{2}
\subsection{Case Study 10}

This case study examines a high-complexity Level 3 task on the Citizen Verizon (VZ) net-zero roadmap, which outlines VZ's transition toward operational carbon neutrality by 2035. The query is challenging because it requires a longitudinal review across three phases: the strategic shift from 2019-2021, mid-term milestones in 2025 and 2030, and long-term decarbonization targets. To deliver an expert assessment, the agent must synthesize key financial and technical disclosures, such as the use of \$3 billion in green bonds for large-scale renewable projects and the goal of sourcing 50\% renewable electricity by 2025.
\end{multicols}

\begin{CaseStudyBox}
    \textbf{Question:} Analyze Verizon (VZ)’s path to net zero roadmap as disclosed in their 2020 ESG reports. Your response should include a detailed qualitative review of the following periods: 1) historical actions (2019–2021), 2) mid-term targets (Year 2025 and Year 2030), and 3) ultimate objects in 2035.  Formulate your findings into a professional ESG analysis report.

    \textbf{Template:} \texttt{template 6, different stock, net zero target}

    \textbf{Level:} Level 3

    \textbf{Generated Report:}
    \vspace{0.2cm}
    
    \begin{center}
        \fbox{\includegraphics[width=0.31\textwidth]{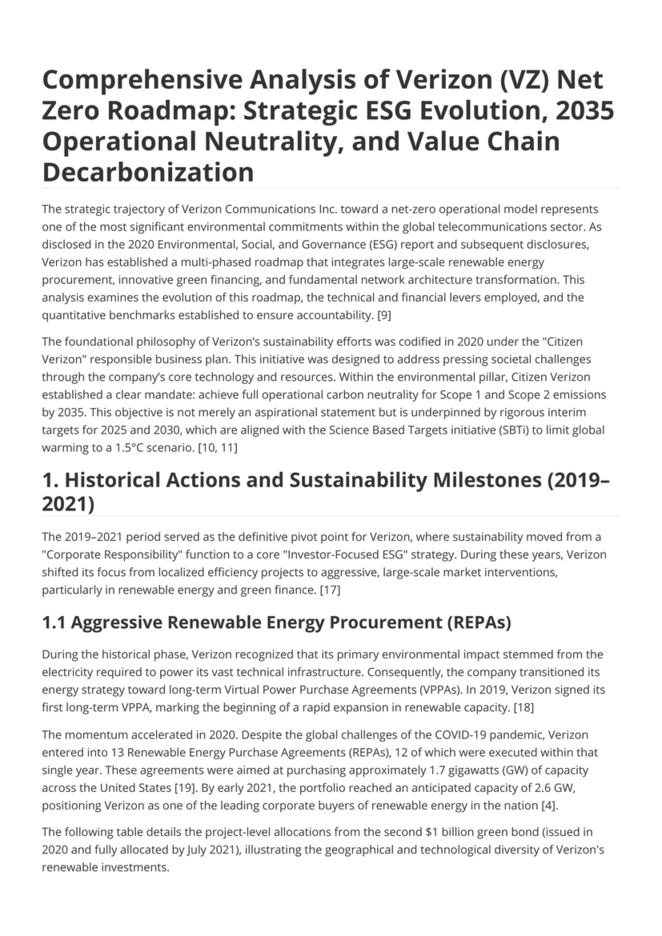}} \hfill
        \fbox{\includegraphics[width=0.31\textwidth]{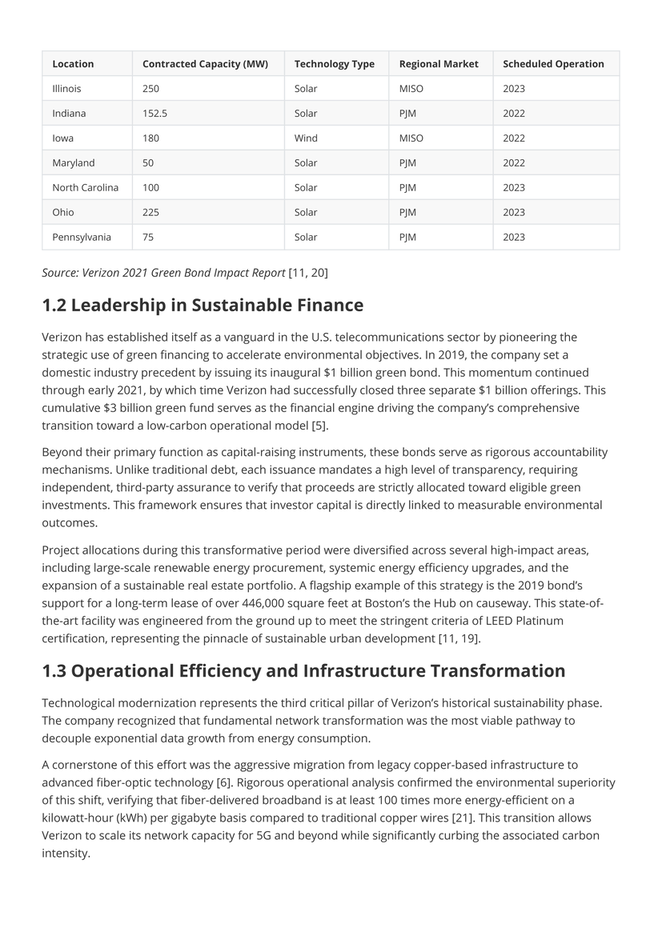}} \hfill
        \fbox{\includegraphics[width=0.31\textwidth]{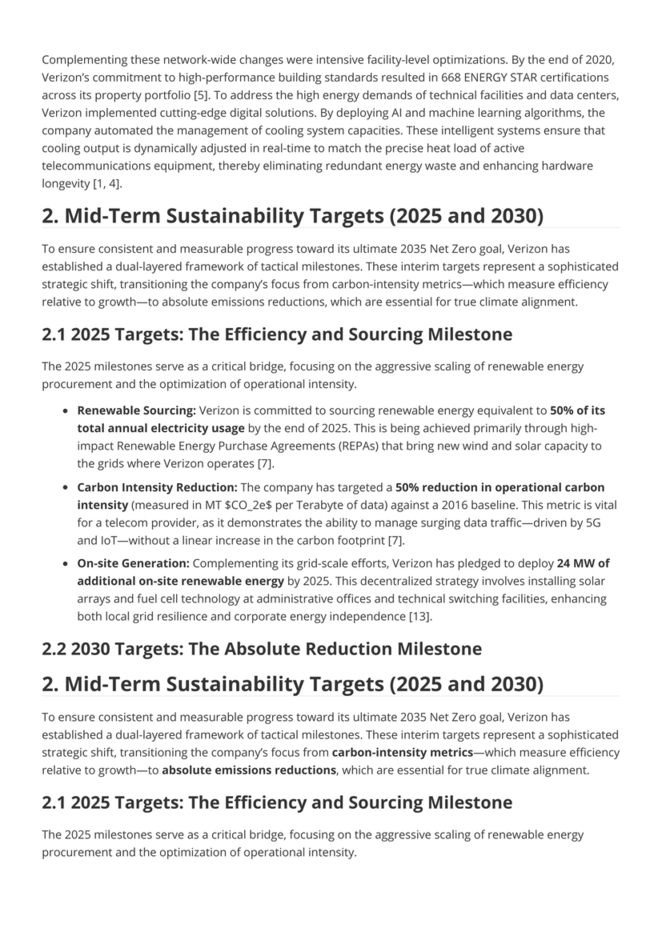}}
        
        \vspace{0.3cm}
        
        \fbox{\includegraphics[width=0.31\textwidth]{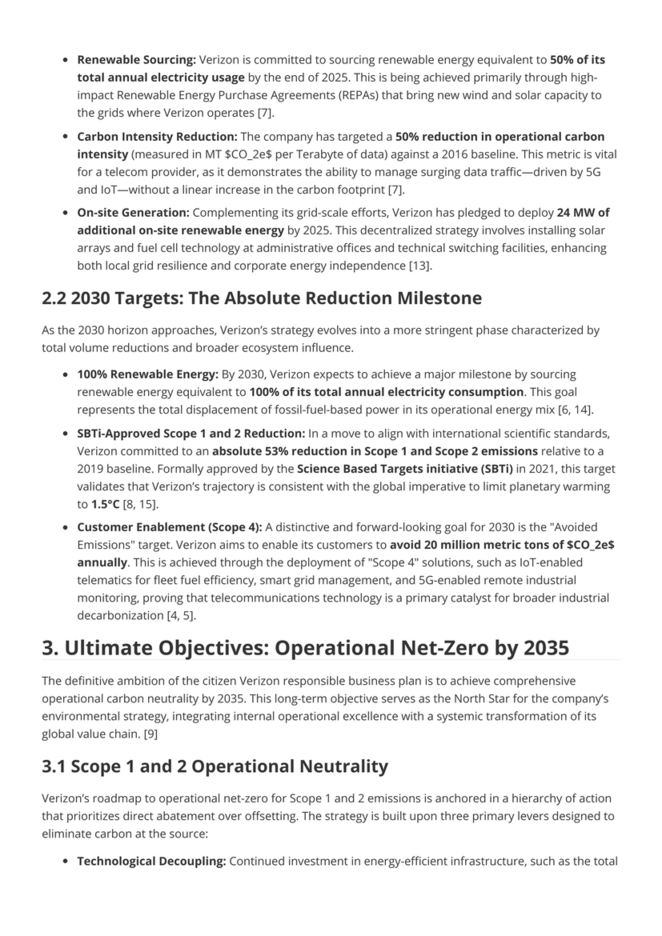}} \hfill
        \fbox{\includegraphics[width=0.31\textwidth]{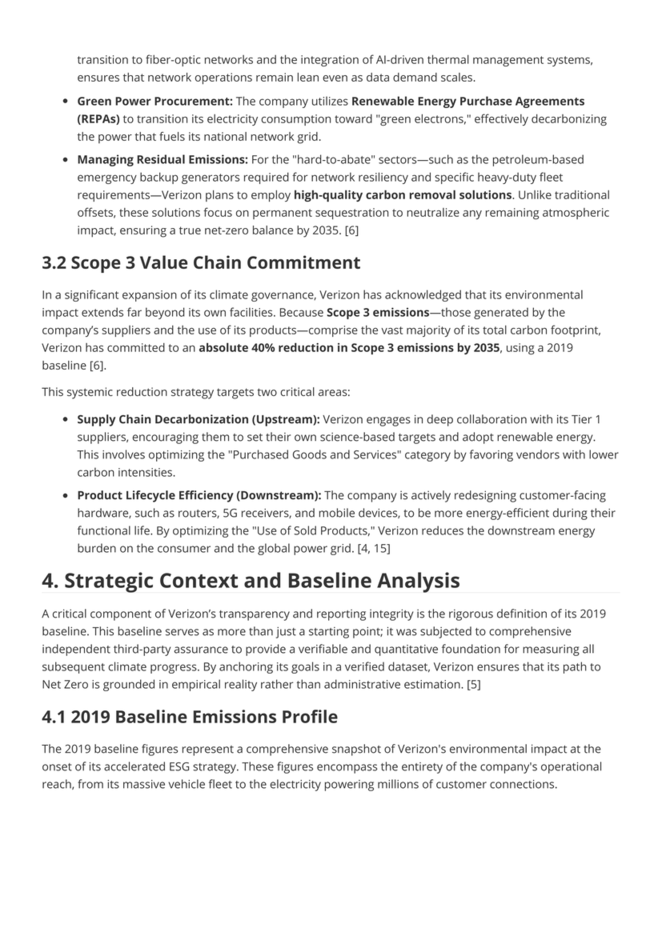}} \hfill
        \fbox{\includegraphics[width=0.31\textwidth]{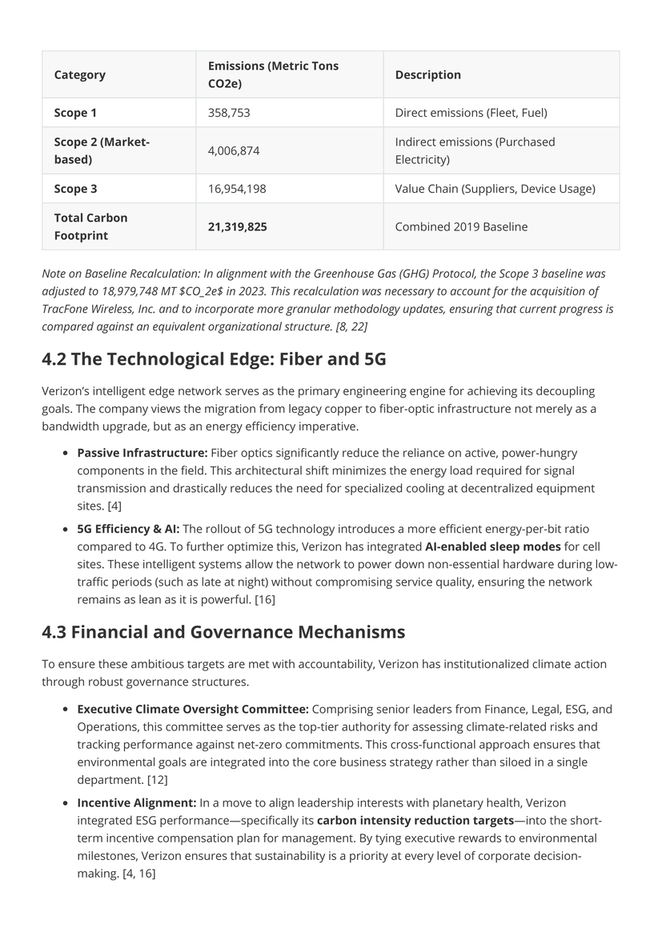}}

        \vspace{0.3cm}

        \fbox{\includegraphics[width=0.31\textwidth]{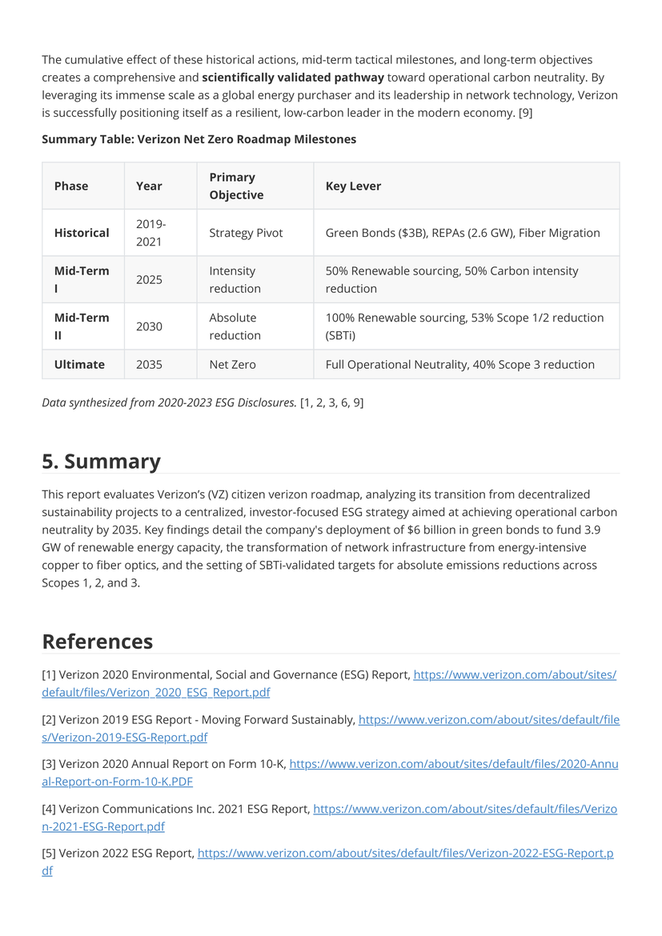}} \hfill
        \fbox{\includegraphics[width=0.31\textwidth]{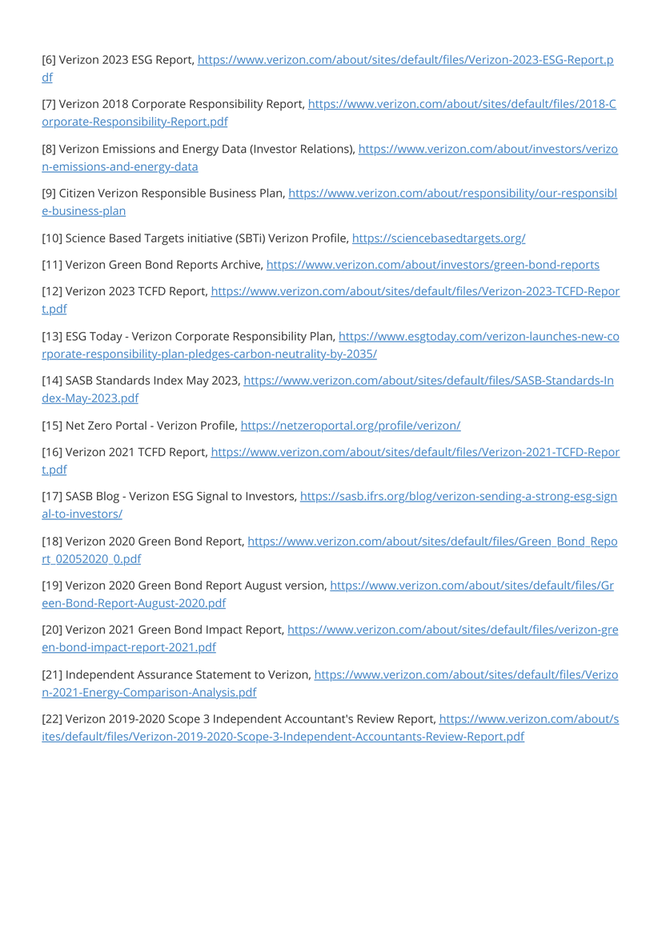}} \hfill
        \begin{minipage}[b]{0.31\textwidth} \quad \end{minipage} 
    \end{center}
\end{CaseStudyBox}

\end{document}